\documentclass[11pt]{article}

\usepackage{pareto}
\usepackage{amsfonts}      
\usepackage{nicefrac}
\usepackage{tikz}
\usetikzlibrary{arrows.meta,positioning,shapes.geometric}
\usepackage{fvextra}   
\usepackage[hyperpageref]{backref}

\newcommand{\doi}[1]{doi:\,\href{https://doi.org/#1}{#1}}

\newcommand{\yesrate}{\text{yes-rate}}

\crefname{tcb@cnt@theorem}{Def.}{Defs.}
\Crefname{tcb@cnt@theorem}{Def.}{Defs.}

\newcommand{\LineStretch}{1.145}
\newcommand{\ParaSkipPt}{0}
\newcommand{\PreConclusionEm}{0}
\hypersetup{
  pdftitle    = {LLMs Show No Signs Of Individuated Metacognition},
  pdfauthor   = {Anonymous},
  pdfkeywords = {metacognition, calibration, large language models,
                 confidence elicitation, factor analysis, AI evaluation},
}

\title{LLMs Show No Signs Of Individuated Metacognition}

\shorttitle{LLMs Show No Signs Of Individuated Metacognition}

\author{%
\setcounter{footnote}{1}%
\hfill\begin{tabular}{c@{\hspace{3em}}c}
M. Moran\thanks{Both authors corresponding.} & Mark Whiting\footnotemark[\value{footnote}] \\
Pareto AI & Pareto AI \\
\texttt{mmoran@smf-ai.com} & \texttt{mark@pareto.ai}
\end{tabular}\hfill\null
}

\date{}

\begin{document}

\maketitle

\begin{abstract}
Confidence-weighted routing, selective abstention, and ensemble weighting all assume that a model's stated confidence is informative about its capability on the question being asked. They presume functional metacognition, the capacity to assess one's own capabilities, without exercising them. Aggregate calibration is well studied, with mixed results, but the underlying structure of elicited confidence is less well understood. We decompose binary confidence judgements from 20 frontier Large Language Models (LLMs) across six benchmarks using tetrachoric factor analysis paired with pairwise calibration, asking whether two models that differ in confidence also differ in performance. On factual recall and information retrieval benchmarks the cross-model confidence matrix is approximately rank-one and a single dominant factor captures most of the latent variance. Models retrieving facts share an item-level difficulty axis and differ mainly in their decision thresholds along it. Across all benchmarks the relationship between confidence and performance collapses once items that all models agree on are removed. Inter-model pairwise calibration is small even where statistically significant, and what remains shrinks to nothing once base-rate differences along the shared factor are controlled for. Mathematical reasoning is the apparent exception, but this turns out to be a confound where reasoning models answer questions about their confidence by trying to solve them in their chain of thought, bypassing the sub-symbolic self-knowledge we seek to measure. We find no evidence for significant verbalised individuated metacognition in any tested domain. A trivial logistic regression on surface features of the confidence-judgement reasoning text beats the median confidence model's repeatedly sampled judgements on half the benchmarks with equivalent information, evidence that extracting uncertainty from the reasoning trace recovers signal the binary judgement does not.
\end{abstract}

\keywords{metacognition, calibration, large language models, confidence
          elicitation, factor analysis, pairwise calibration, AI evaluation}

\section{Introduction}

Confidence-weighted routing~\citep{chen2024frugalgpt,aggarwal2024automix}, selective abstention~\citep{geifman2017selective,kamath2020selective}, and ensemble weighting~\citep{wang2023selfconsistency,jiang2023llm} all assume that a model's stated confidence is informative about that model's capability on the item being asked. Surface-level calibration is well studied~\citep{kadavath2022language,lin2022teaching}, but the structure of these signals across models is less well understood. The extent to which the calibration signal is private to each model, and its alignment with performance relative to other models, remains an open question.

\begin{definition}{Individuated metacognition}{individuated}
A model's capacity to assess its own capabilities at the item level in a way specific to that model, without exercising those capabilities, beyond what is explained by a shared signal.
\end{definition}

We find that models possess knowledge of what language models as a class are capable of but have little understanding of their specific capabilities. Confidence variation across frontier models reduces largely to a smaller number of dominant shared difficulty estimation axes with model-specific thresholds, not to a high-dimensional basis of individuated self-knowledge. Additionally, each model's position on these axes is not well correlated with actual performance. Performance is also intrinsically low-dimensional, contributing further evidence to a g factor for LLMs~\citep{burnell2023revealing,ilic2024evidence}, but the correlation between the latent confidence and latent performance axis is low outside of universally agreed upon items. As we show explicitly with pairwise calibration measurements, one model being more confident than another has near-zero correlation with performance beyond what is explained by performance base-rate differences.

We evaluate 20 frontier LLMs across six benchmarks spanning factual recall, multi-domain knowledge, legal reasoning, mathematics, and scientific code generation. In addition to finding a low-dimensional structure and a lack of individuated calibration signal; a trivial logistic regression on features of a model's own confidence-judgement reasoning text beats its binary self-assessment on three of six benchmarks, indicating that elicited reasoning carries information the model's final answer does not use.

Our core findings:

\begin{itemize}
  \item \textbf{Confidence largely reduces to a shared difficulty estimation axis on knowledge tasks.} LLM binary confidence across factual, retrieval, and legal benchmarks decomposes into a single dominant tetrachoric factor accounting for base-rate artifacts. Models differ chiefly in their response thresholds along that shared axis, not in the underlying signal. Reasoning tasks show more complex structure, but this does not translate to higher individualised calibration (\S\ref{sec:decomposition}).

  \item \textbf{Confidence and performance are both low-rank but along different axes.} Both signals reduce to a single shared item-level axis, so a low-dimensional confidence heuristic is in principle the correct approach. However, the primary confidence axis aligns with the primary performance axis mostly on items where everyone agrees (trivial floor and impossible ceiling). On contentious items the two are largely independent. (\S\ref{sec:alignment}).

  \item \textbf{LLM confidence elicitation generates information the final response does not use.} A trivial classifier on hand-crafted surface features of the confidence-judgement reasoning text matches or beats the model's own binary self-assessment on three of six benchmarks. An analogous classifier reading the answer-attempt text wins on two further benchmarks. Both classifiers are weak lower bounds and suggest the model generates the required information to make such judgements but does not reliably make use of it (\S\ref{sec:residual}, \S\ref{app:si:external_classifier}).

  \item \textbf{Pairwise calibration is statistically insignificant without performance differences.} The distribution of Kendall-$\tau$ between model pairs is barely higher than the base-rate-aligned empirical null distribution across all benchmarks. The statistically significant but minor effect size vanishes when controlling for differences in performance base rate, where trivial gains are possible (\S\ref{sec:alignment}). Pairwise calibration can be achieved trivially with large differences in performance - so this fails a necessary condition for individuated metacognition despite the overall pairwise signal being small but statistically significant. In short, Claude's self-assessments are just as likely to describe Gemini's actual performance as its own.

  \item \textbf{Reinforcement learning solicits symbolic shortcuts.} The apparent MathBench exception to the pairwise-calibration result collapses on closer inspection. The signal is concentrated entirely in Reasoning--Non-Reasoning pairs, where reasoning-trained models attempt the problem inline when asked about confidence. This is a symbolic work-around rather than the sub-symbolic self-knowledge we seek to measure (\S\ref{sec:decomposition}).
\end{itemize}

\section{Related Work}

Surface-level LLM calibration is well studied with mixed results. Large models are reasonably well calibrated in aggregate~\citep{kadavath2022language} and verbalised numeric confidence on RLHF chat models can outperform logit baselines on per-model calibration~\citep{tian2023calibration}, yet across providers and prompts verbalised confidence shows systematic overconfidence and poor per-item accuracy~\citep{xiong2024llmuncertainty}. Sampling-based entropy~\citep{kuhn2023semantic} and external classifiers~\citep{ulmer2024apricot,pedapati2025confidence} improve on self-reports but do not resolve whether any model's confidence reflects individuated capability assessment. Recent work on reasoning-trained models reports that calibration accuracy scales with the length and content of pre-judgement reasoning~\citep{podolak2025read,yoon2025reasoning}, a finding we revisit in \S\ref{sec:decomposition}.

Metacognition, cognition about one's own cognition, has a long history in human and animal cognitive science. Foundational work distinguishes the monitoring and control of one's own mental processes~\citep{nelson1990,koriat1997}. The signal-detection framework~\citep{fleming2014} measures metacognitive sensitivity as how well a subject's confidence discriminates their own correct from incorrect responses, separating self-knowledge from underlying task ability, and is formalised within-subject as meta-d'~\citep{maniscalco2012metad}. The Dunning-Kruger effect~\citep{kruger1999unskilled} is one well-known failure mode, with low performers systematically overestimating their own competence. Recent LLM work asks whether large models exhibit analogous capacity, with fine-tuned models shown to outperform external observers at predicting their own behaviour~\citep{binder2025introspection}. Adopting the terminology of~\cite{nelson1990}, we restrict attention to metacognitive monitoring, the model's assessment of its own knowledge state.

Latent-factor analysis and psychometric methods have been increasingly applied to  LLMs in recent years.~\cite{kipnis2025metabench, burnell2023revealing,ilic2024evidence} apply factor analysis to LLM performance and find the correlation structure of LLM task performance to be low, or even one-dimensional depending on the domain. We extend this factor-analytic decomposition from performance to verbalized confidence, as a probe for user facing functional metacognition, and show that the individuation assumption underlying confidence-weighted routing fails and that any model's difficulty estimate is approximately as informative about another model's performance as about its own.

\section{Confidence Decomposition}
\label{sec:decomposition}

\label{sec:setup}
We evaluate 20 LLMs across six benchmarks with diverse cognitive demands (Table~\ref{tab:benchmarks}), measuring both performance and confidence.

\begin{definition}{Calibration (discriminatory sense)}{calibration}
\emph{Calibration} in the discriminatory sense measures whether a model's confidence rank-orders item-level outcomes. The literature also uses a frequency sense (expected calibration error, reliability diagrams~\citep{guo2017calibration}), which asks whether stated yes-rates match observed accuracy rates. We use the discriminatory sense throughout.
\end{definition}

The benchmark suite comprises SQuAD~\citep{rajpurkar2016}, MMLU-Pro~\citep{wang2024mmlupro}, LegalBench~\citep{guha2023legalbench}, the MATH dataset~\citep{hendrycks2021math} (referred to as MathBench throughout), Omni-MATH~\citep{gao2024omnimath}, and SciCode~\citep{tian2024scicode}.
Models include representatives from Anthropic~\citep{anthropic2024claude3} (Claude 3 Haiku through Sonnet 4.5), OpenAI~\citep{openai2023gpt4} (GPT-4o through GPT-5.2), Google~\citep{gemini2024gemini15} (Gemini 2.0 Flash through 3.1 Pro), DeepSeek~\citep{deepseekr1} (R1), Meta~\citep{dubey2024llama3} (Llama 3.1-70B, 3.3-70B), Mistral~\citep{jiang2023mistral} (Medium 3.1, Small 3.2-24B), and Qwen~\citep{yang2024qwen25,hui2024qwen25coder} (2.5-72B, 2.5-Coder-32B). GPT-5.2 refers to the OpenAI API endpoint by that identifier at the time of data collection. Provider citations are family-level references to the most recent technical reports available at time of writing; not every individual variant tested has a separate published model card.

We decompose every benchmark question into a minimal question specification and optional additional context. Per-benchmark decomposition procedures are described in \S\ref{app:decomposition}. SQuAD has been modified into a general-knowledge benchmark by removing the original Wikipedia passages and treating them as optional context, and SciCode has been transpiled by a frontier LLM to a new schema due to significant ambiguities in the original question set. Both changes preserve the original problems and are documented in \S\ref{app:decomposition}.

For each benchmark we elicit binary (yes/no) confidence judgments in one or both of two modes, prospective (\cref{def:prospective}) and counterfactual (\cref{def:counterfactual}). LegalBench is included only with prospective probes because its tasks lack the question-plus-removable-context structure required for the counterfactual probe (see \S\ref{app:decomposition}).

\begin{definition}{Prospective judgements}{prospective}
A \emph{prospective} judgement assesses capability before the model attempts the problem (``Can you answer this without the context paragraph?''). The model cannot see the extra context but is made aware it exists.
\end{definition}

\begin{definition}{Counterfactual judgements}{counterfactual}
A \emph{counterfactual} judgement assesses information necessity after the fact, filtered on the model succeeding at the task with full information (``Given you answered correctly, could you have answered correctly without the context paragraph?'').
\end{definition}

\begin{table}[!ht]
\centering
\caption{Six benchmarks spanning factual recall, legal reasoning, mathematics, and scientific code generation. Models that could not consistently conform to the output schema for a given benchmark were excluded from that benchmark's analysis.}
\label{tab:benchmarks}
\small
\begin{tabular}{llcl}
\toprule
\rowcolor{PBoxPrimary}
Benchmark & Trials & Probe Types \\
\midrule
SQuAD (Modified) & 1000 & Prospective, Counterfactual \\
MMLU-Pro& 1000 & Prospective, Counterfactual \\
LegalBench& 1000 & Prospective \\
MathBench& 500 & Prospective, Counterfactual \\
SciCode (Modified) & 421 & Prospective, Counterfactual \\
Omnimath & 1000 & Prospective, Counterfactual \\
\bottomrule
\end{tabular}
\end{table}

We use single-shot binary rather than sampling or graded confidence elicitation for three reasons. First, full logits are not available from all providers, and reasoning strings predetermine the logits of the final answer in an opaque way that would require multiple samples to recover. Second, although graded verbalised confidence on RLHF chat models can outperform logit baselines on per-model calibration~\citep{tian2023calibration}, our estimand is cross-model covariance, where graded scales introduce verbalisation drift across providers and remain unevenly conditioned even within model~\citep{yang2024verbalized,xiong2024llmuncertainty}. Third, the linear components of phrasing biases cancel in inter-model comparisons so we are less concerned about effects from word order than if we were measuring absolute calibrations. We use direct phrasing throughout, avoiding prompt engineering techniques that risk inducing divergent responses across architectures. Verbatim prompt templates for all six benchmarks are in \S\ref{app:si:prompts}. We use a temperature of zero across all benchmarks.

\begin{figure}[t]
\centering
\newcommand{\calpanel}[3]{%
  \begin{subfigure}[t]{0.235\textwidth}%
    \includegraphics[width=\linewidth]{#1}%
    \caption{#2}\label{fig:calibration_all:#3}%
  \end{subfigure}%
}
\calpanel{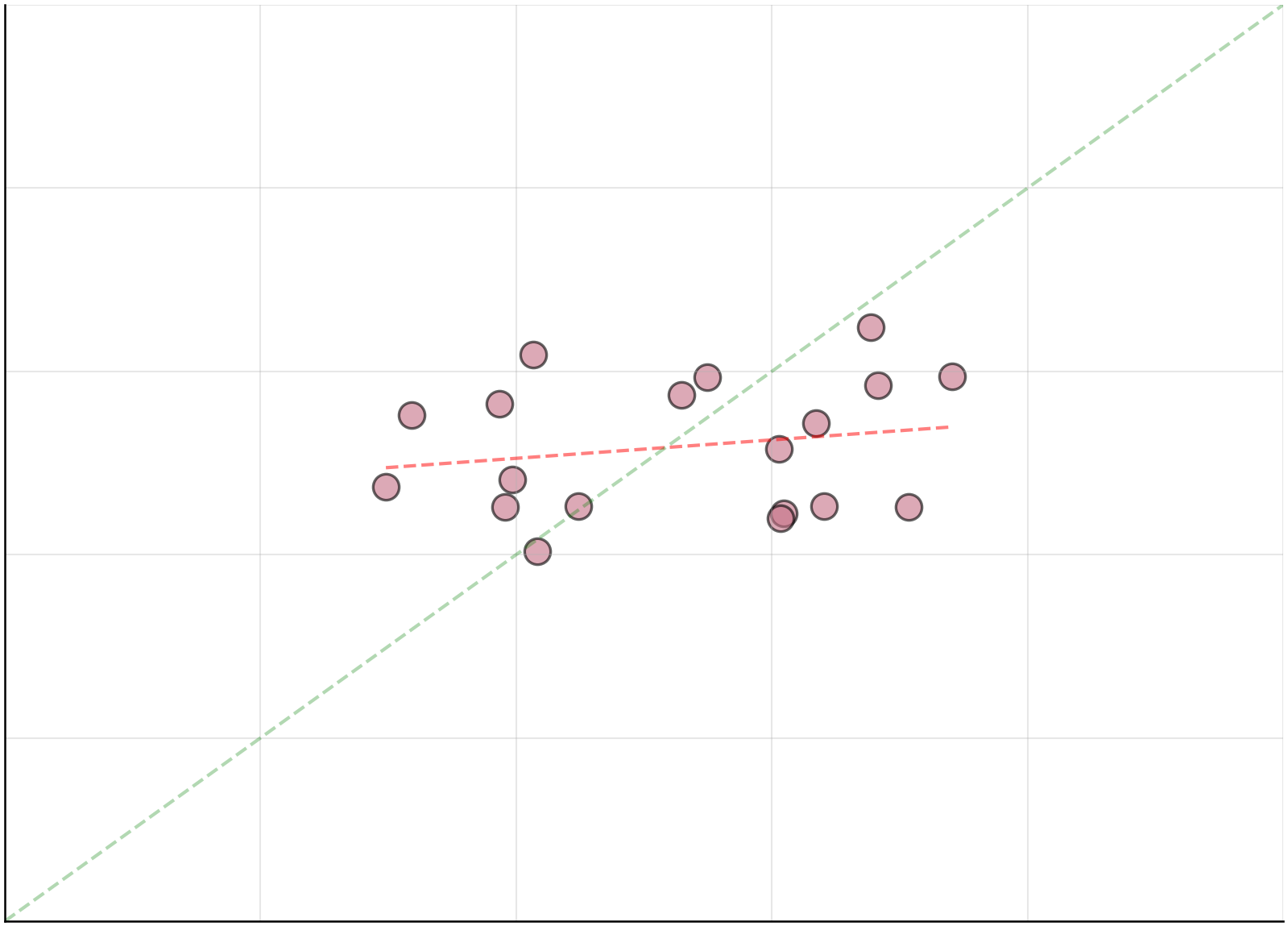}{SQuAD (prospective)}{a}\hfill
\calpanel{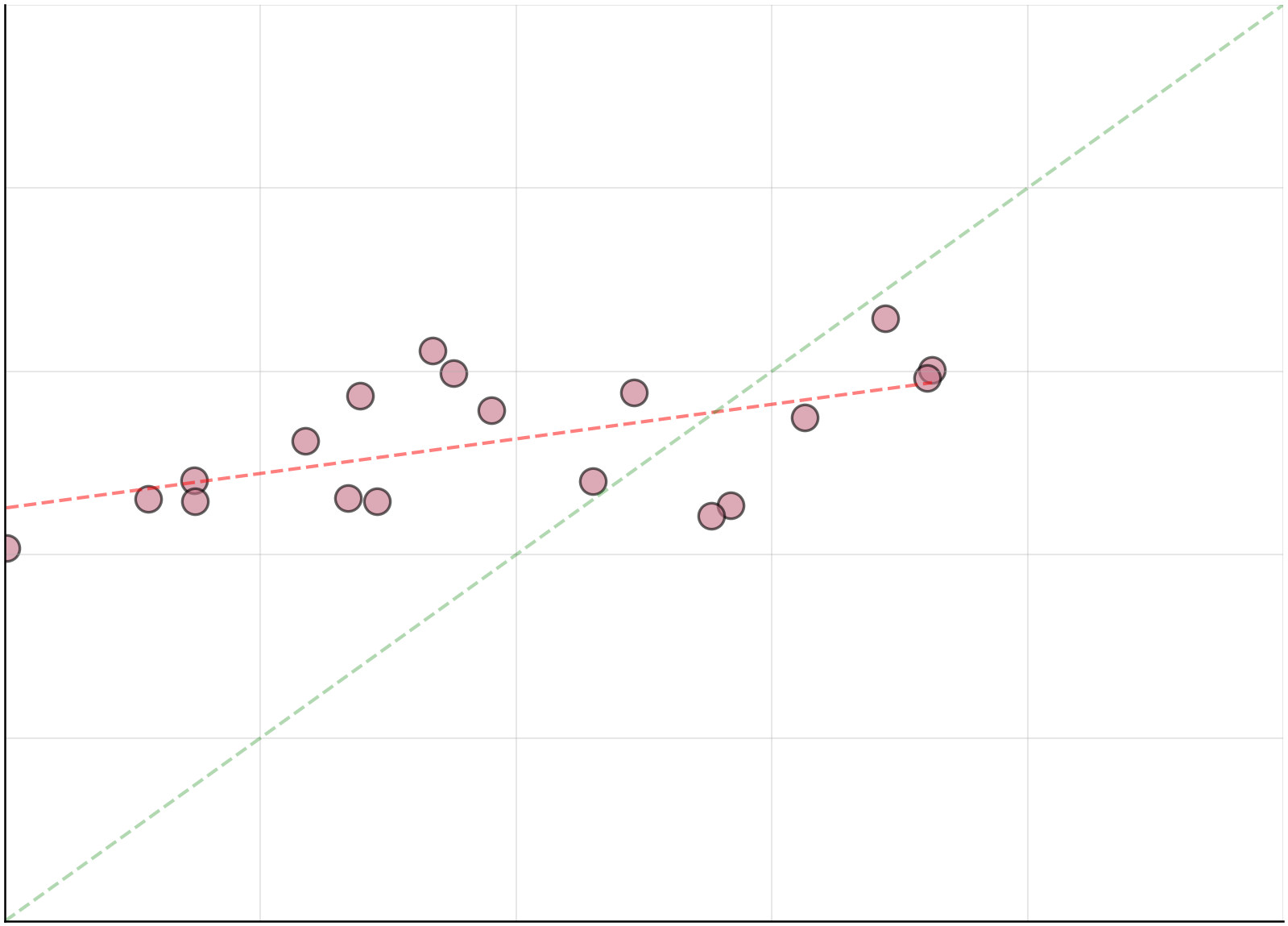}{SQuAD (counterfactual)}{b}\hfill
\calpanel{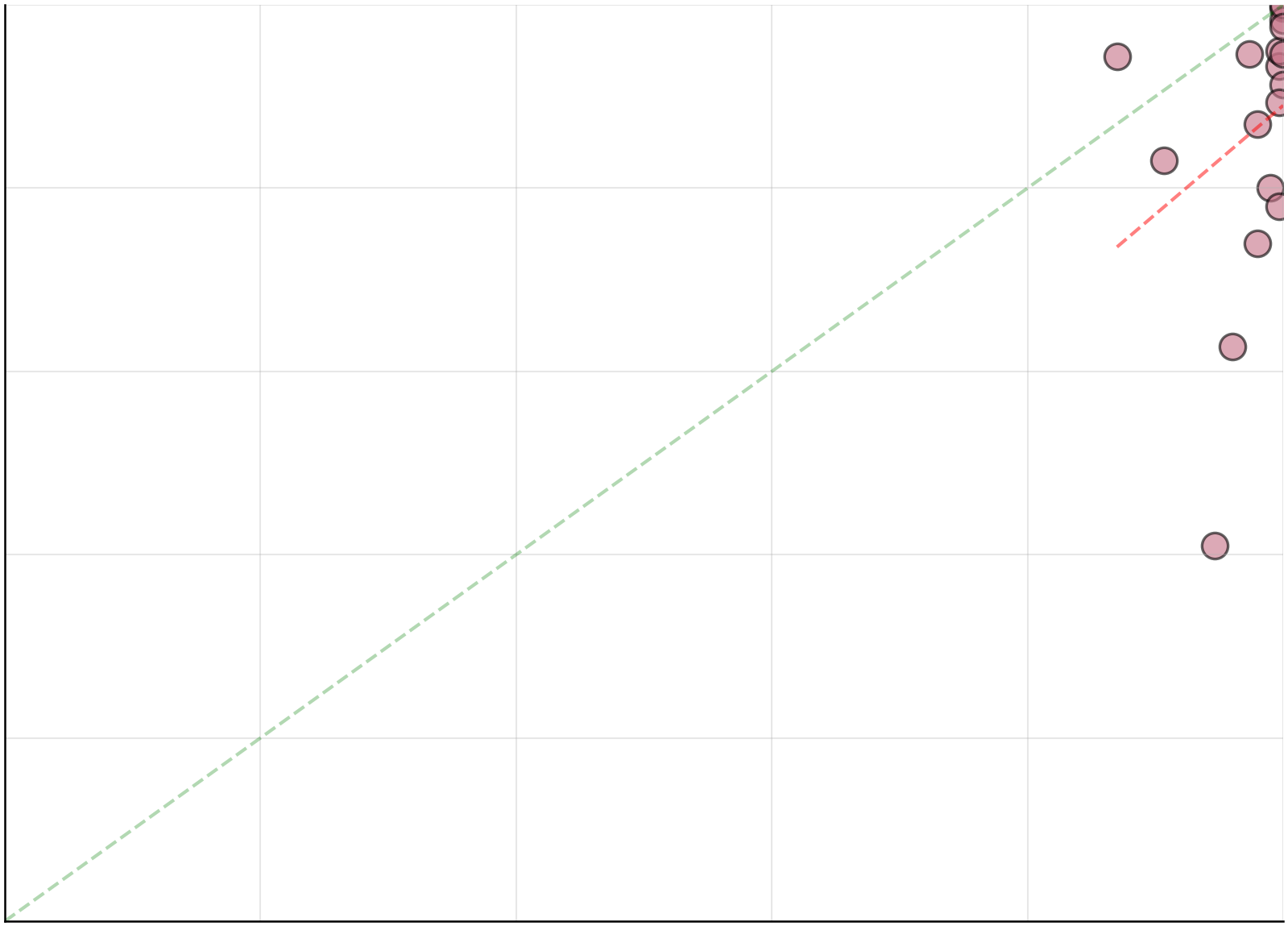}{MathBench (prospective)}{c}\hfill
\calpanel{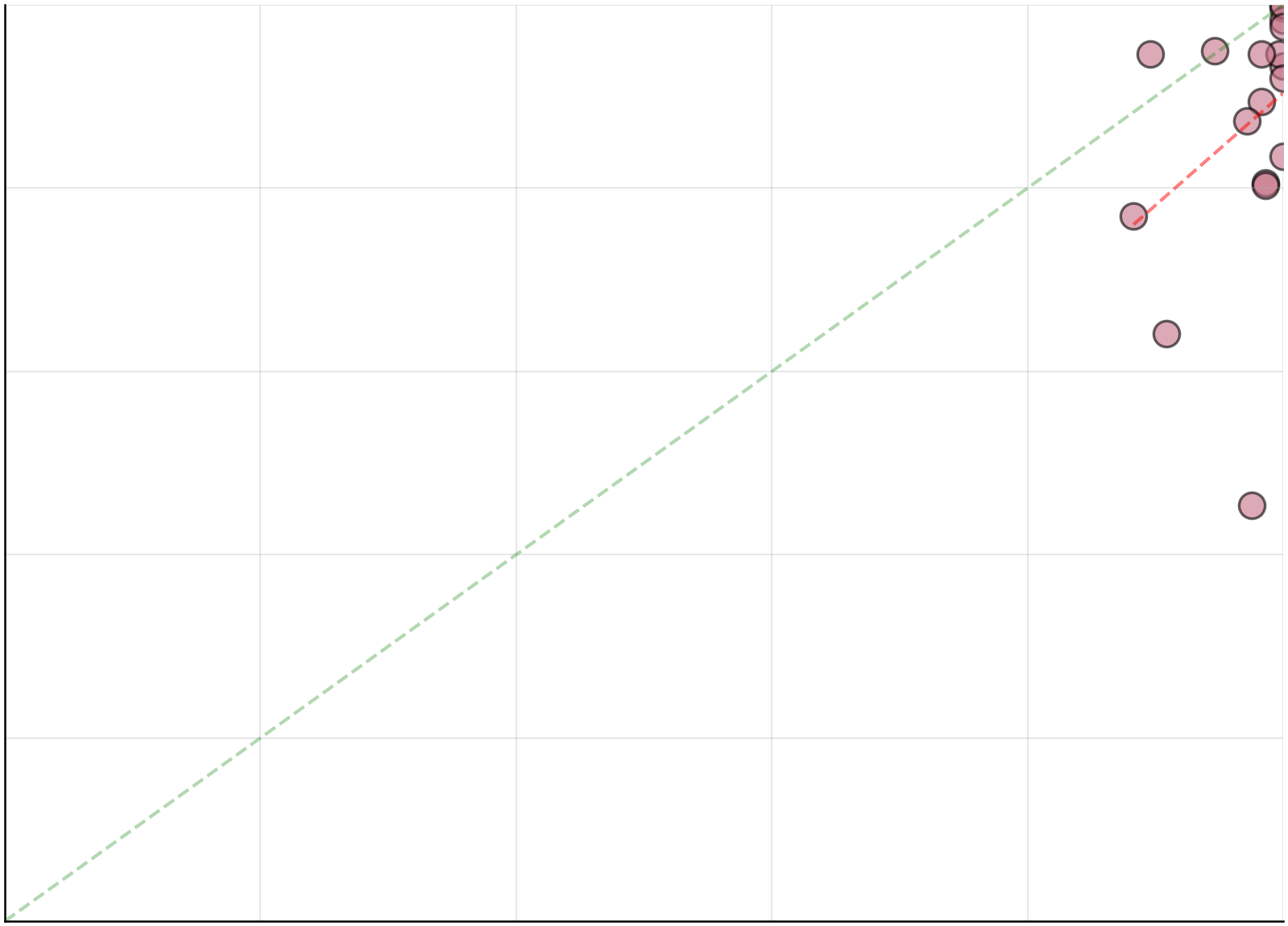}{MathBench (counterfactual)}{d}

\vspace{0.5em}
\calpanel{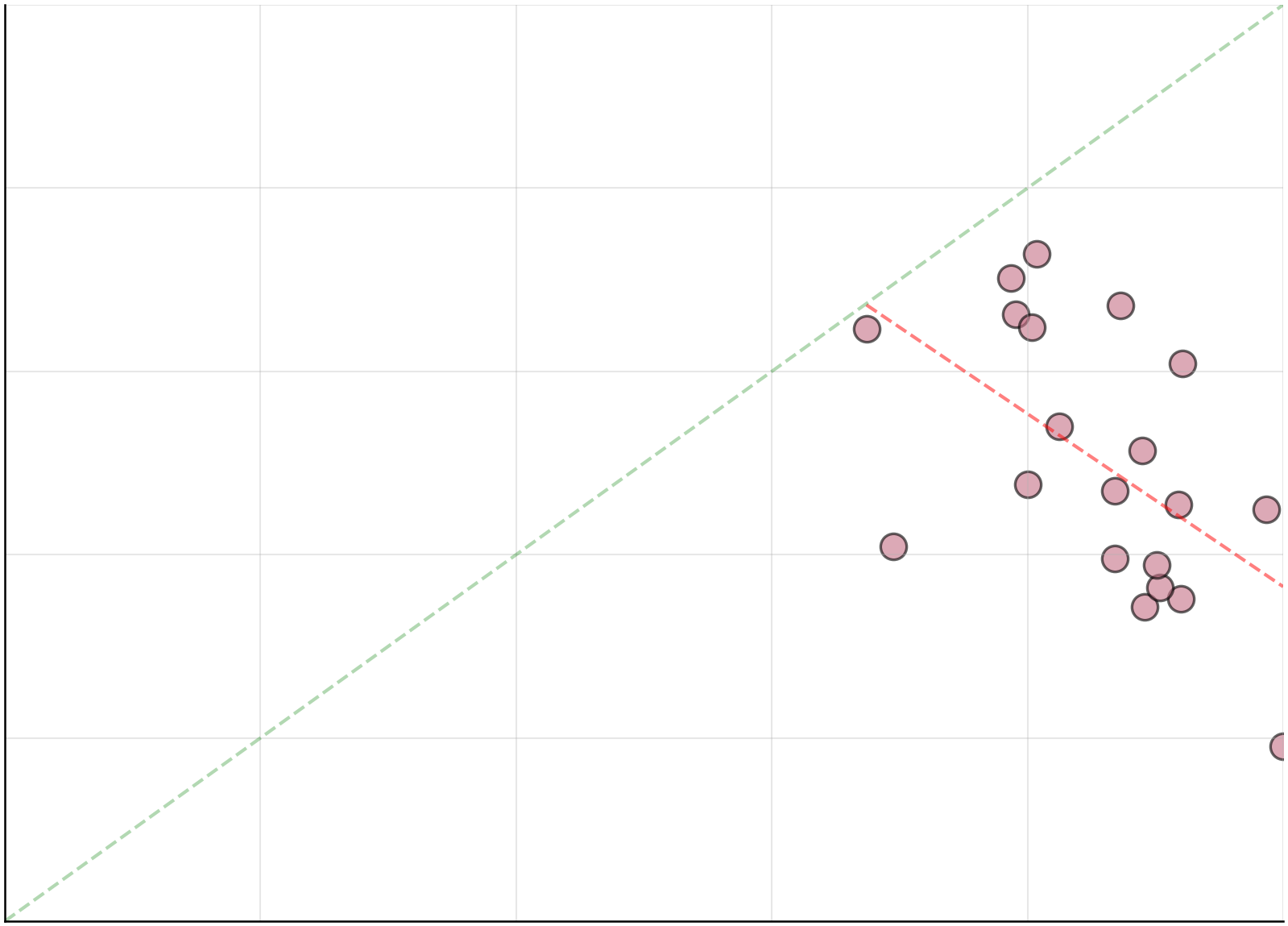}{MMLU-Pro (prospective)}{e}\hfill
\calpanel{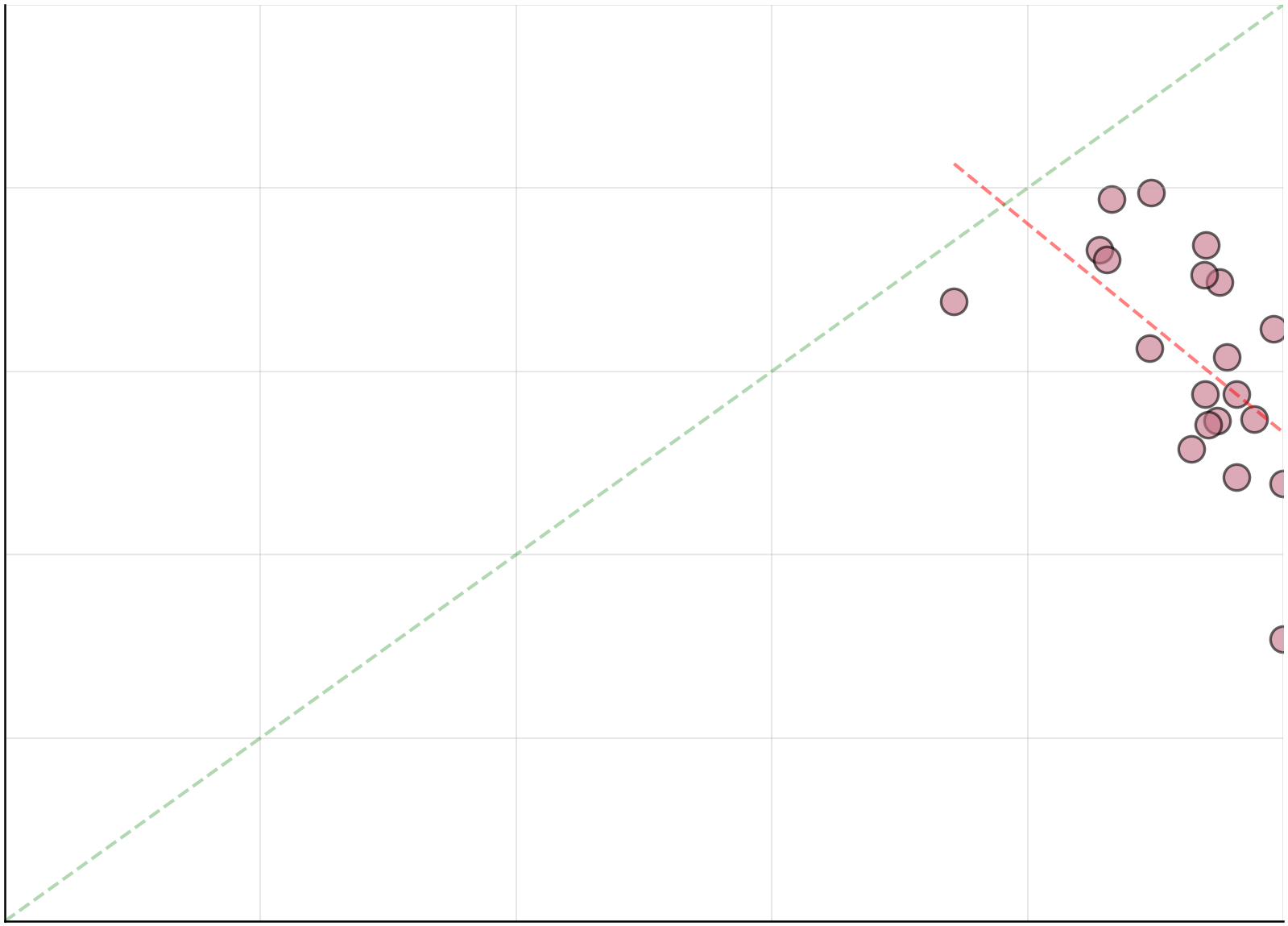}{MMLU-Pro (counterfactual)}{f}\hfill
\calpanel{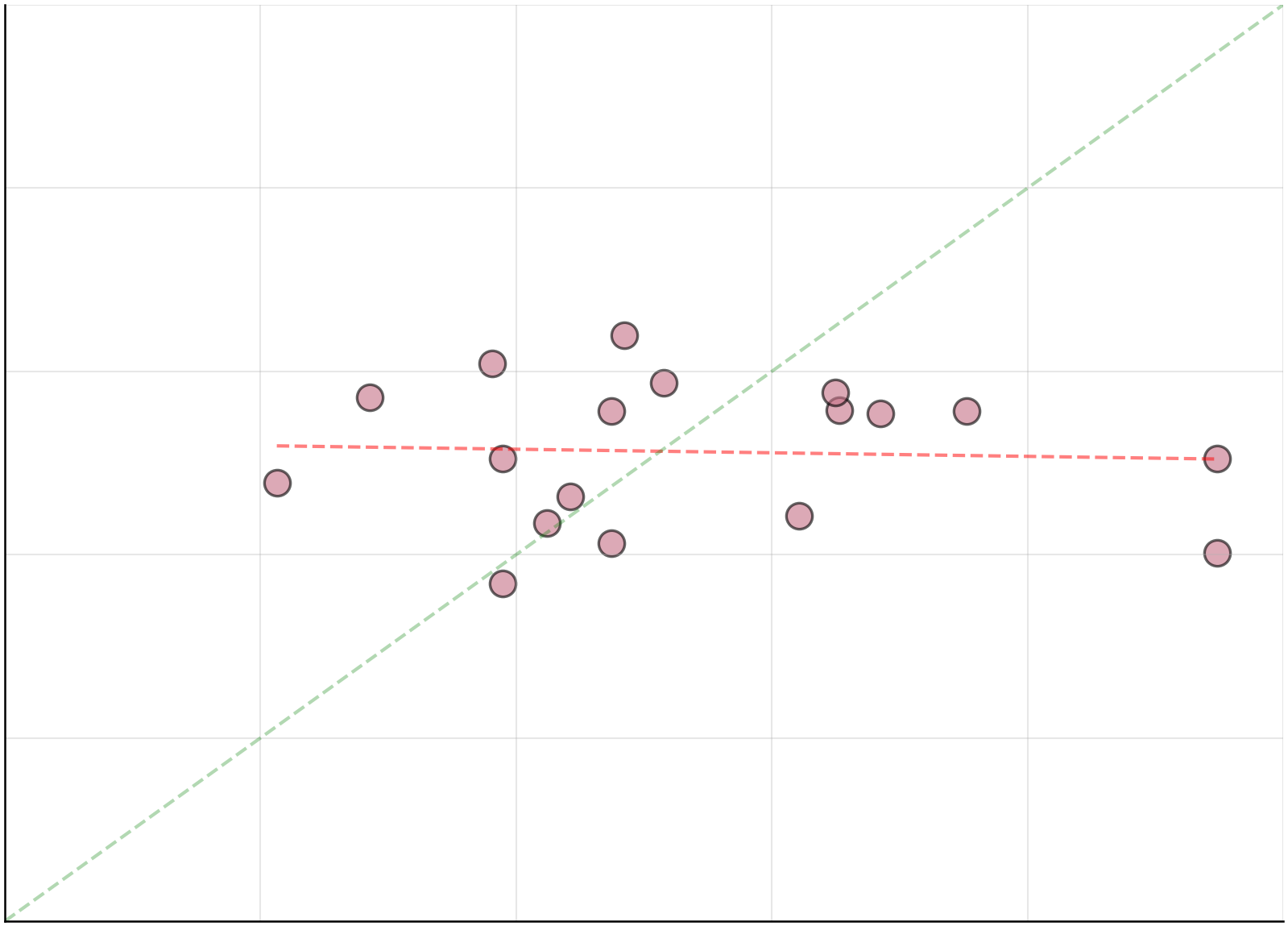}{SciCode (prospective)}{g}\hfill
\calpanel{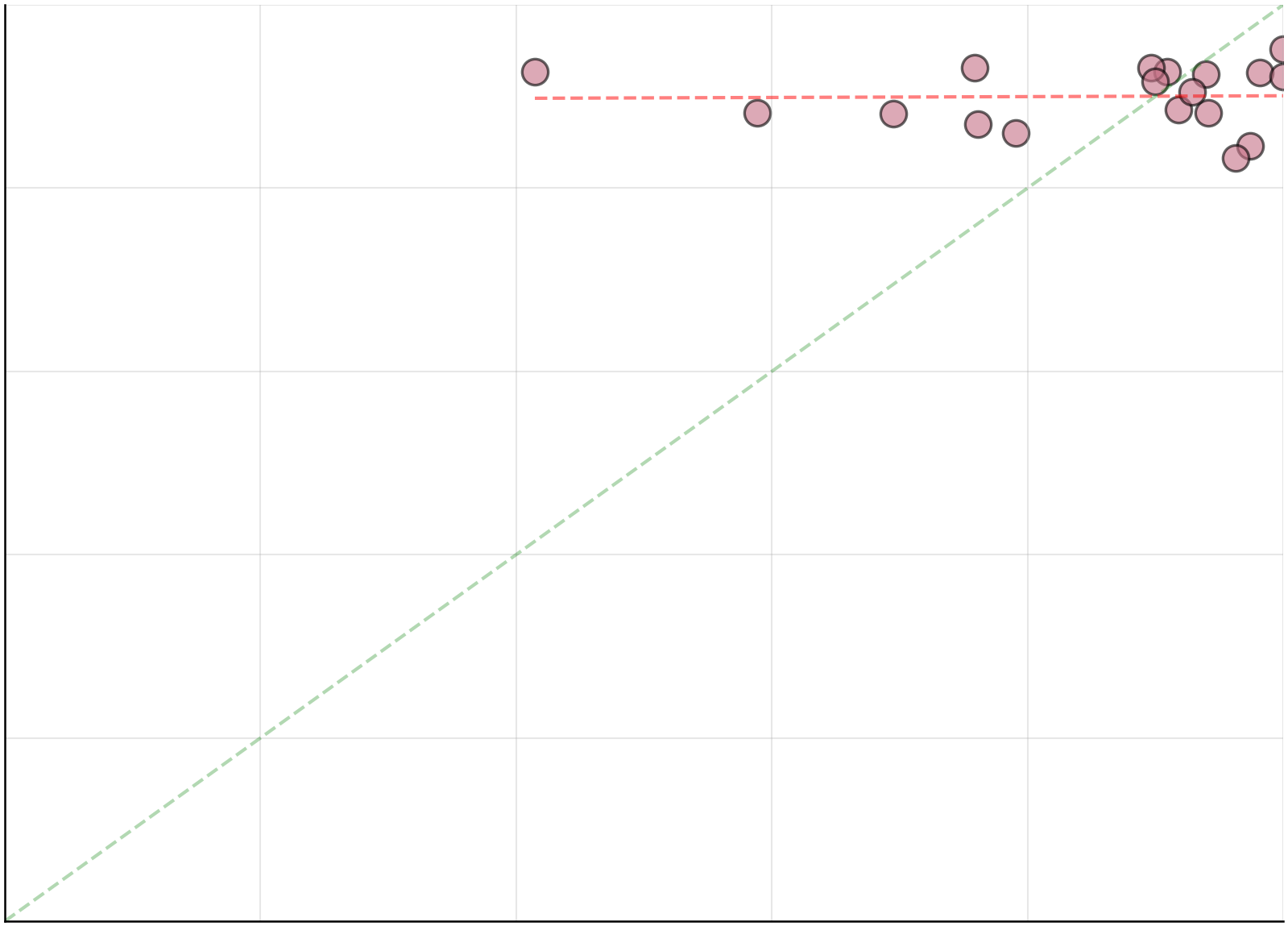}{SciCode (counterfactual)}{h}

\vspace{0.5em}
\calpanel{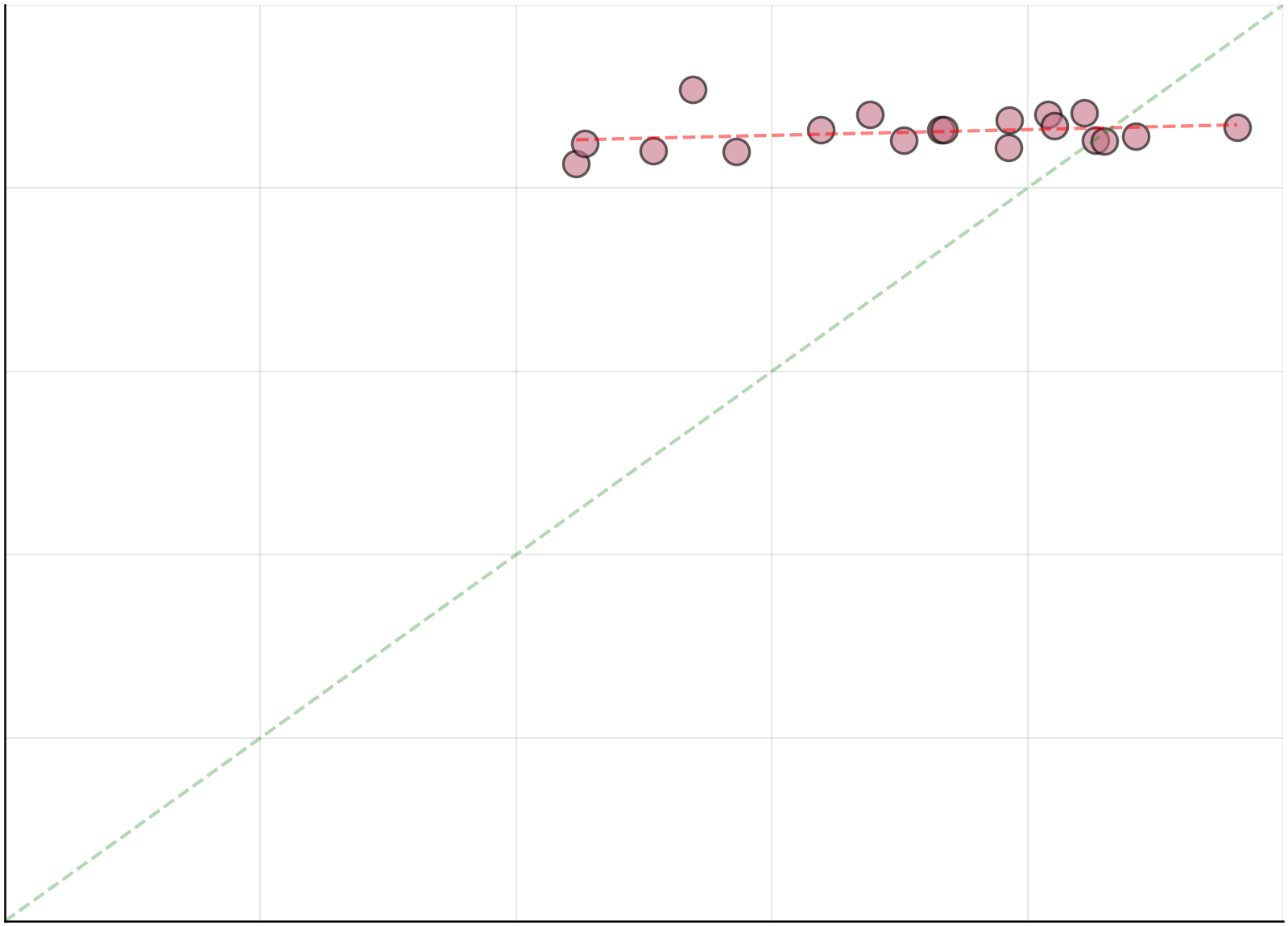}{LegalBench (prospective)}{i}\hfill
\begin{subfigure}[t]{0.235\textwidth}\centering\vspace{2.5em}\small (no counterfactual probe)\end{subfigure}\hfill
\calpanel{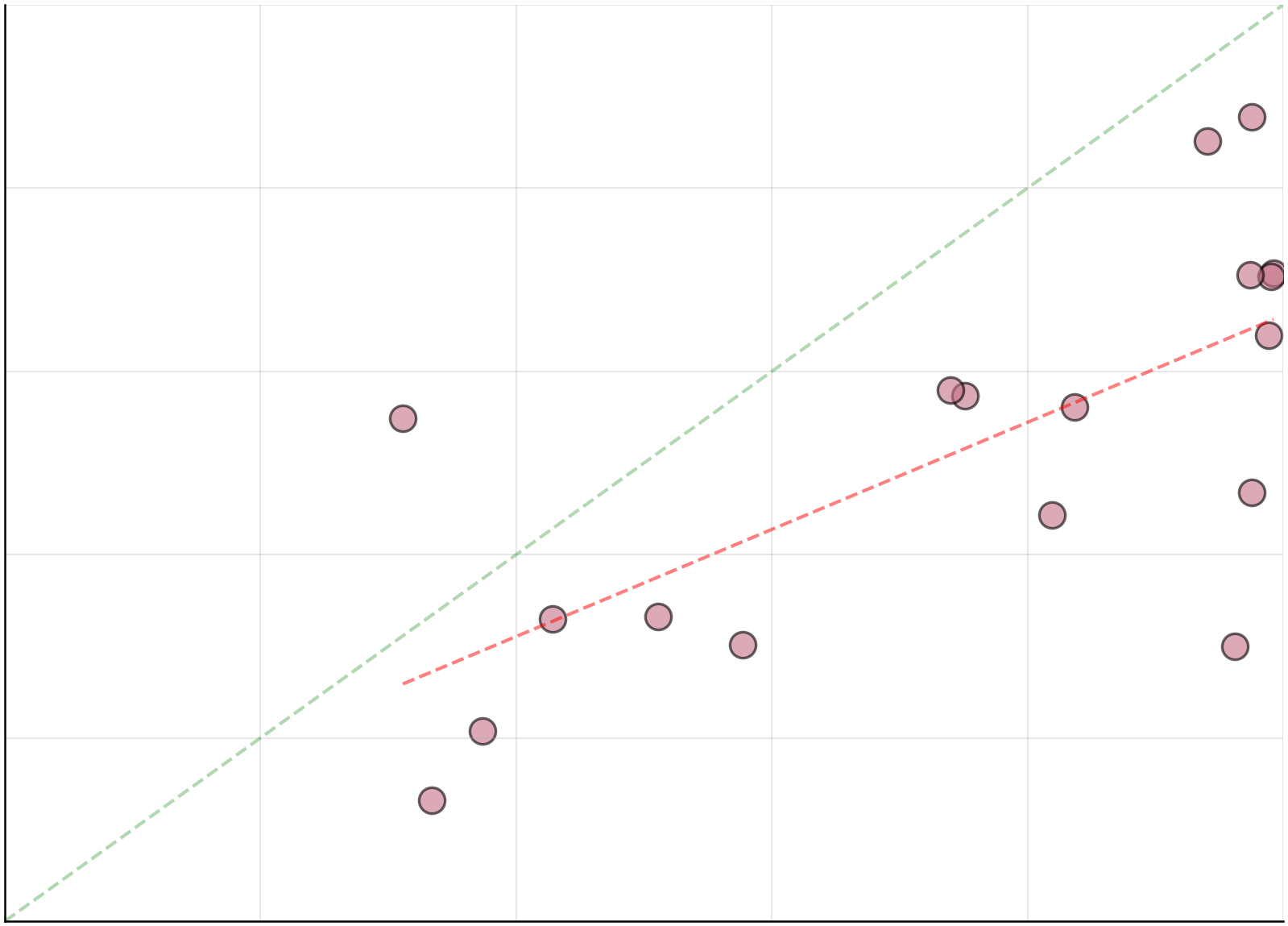}{OmniMath (prospective)}{j}\hfill
\calpanel{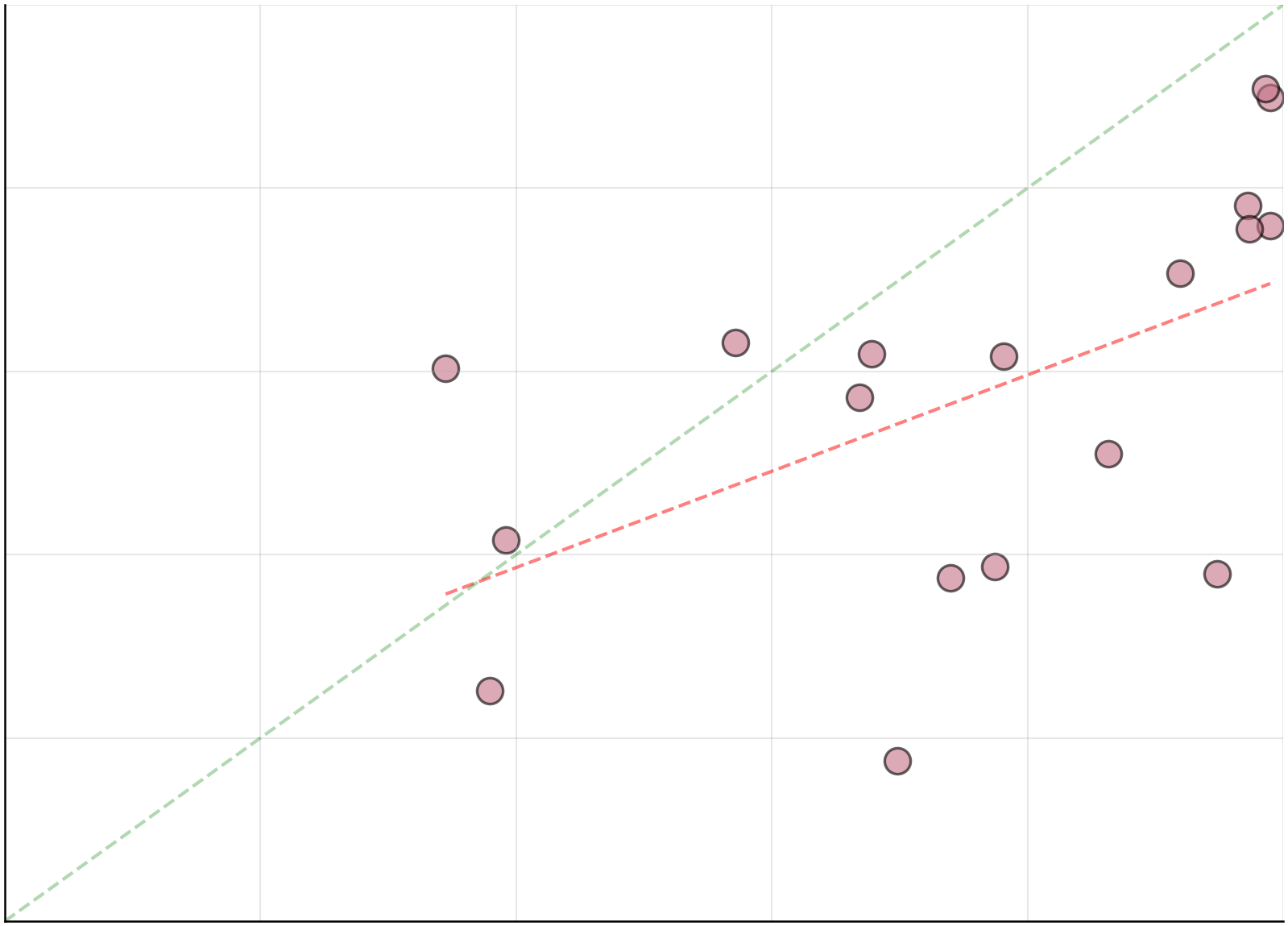}{OmniMath (counterfactual)}{k}

\caption{Model-level aggregate confidence against accuracy for all six benchmarks. Each point is one model. In every panel the horizontal axis is the model's mean confidence (fraction of items rated yes) and the vertical axis is its accuracy (fraction of items answered correctly); both run from 0 to 1 and tick labels are suppressed for legibility at this grid size. The left block (columns 1--2) shows retrieval and factual benchmarks (SQuAD, MMLU-Pro, LegalBench), the right block (columns 3--4) shows reasoning benchmarks (MathBench, SciCode, OmniMath). Within each block, the first column is the prospective probe and the second column is the counterfactual probe. The LegalBench counterfactual probe is not in the protocol and is omitted. Per-point model identities are in the labeled variant in \S\ref{app:si:calibration_all}.}
\label{fig:calibration_all}
\end{figure}

\begin{figure}[t]
\centering
\begin{subfigure}[b]{0.85\textwidth}
\includegraphics[width=\linewidth]{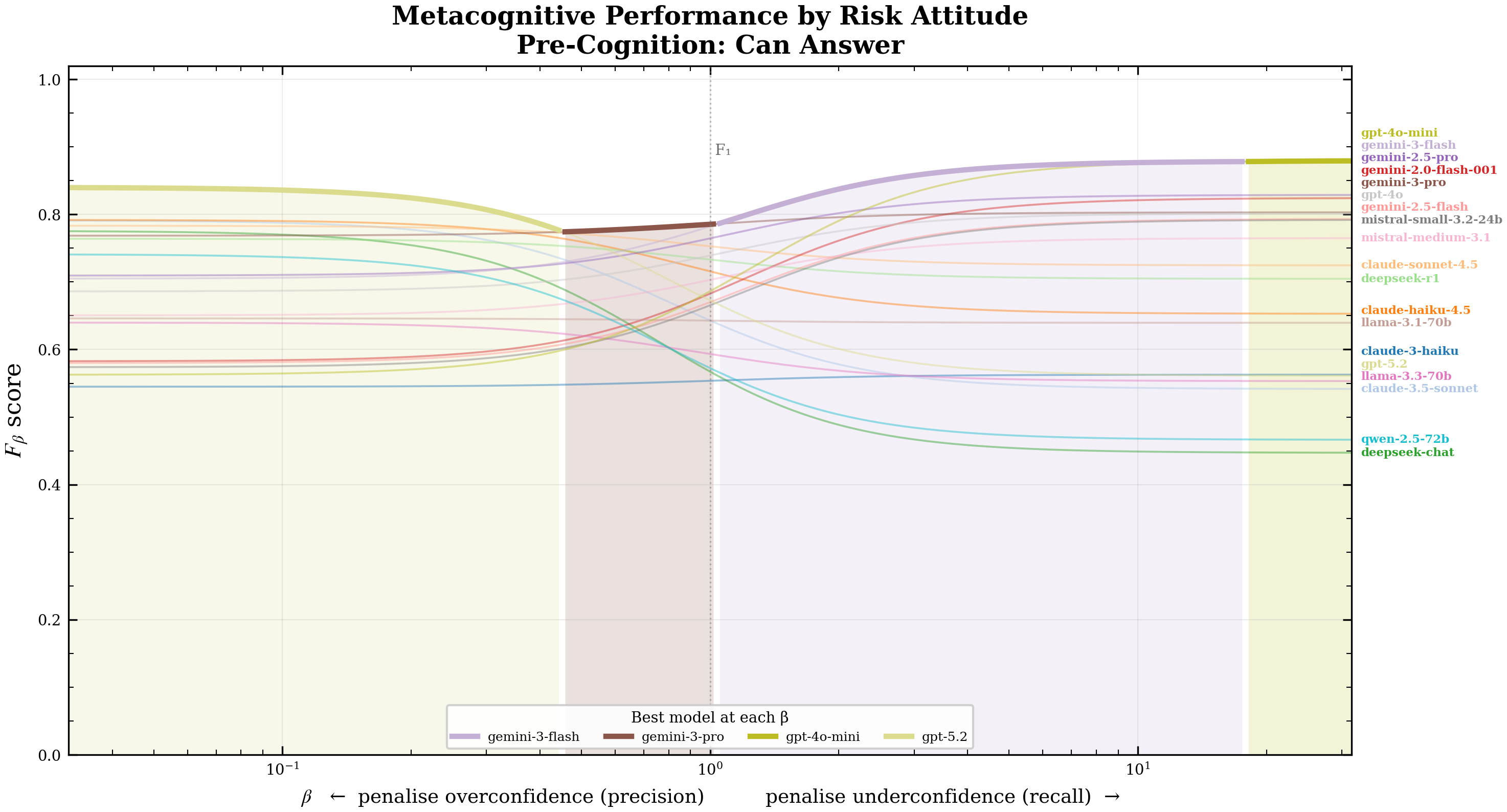}
\caption{SQuAD (prospective)}
\end{subfigure}

\vspace{0.6em}

\begin{subfigure}[b]{0.85\textwidth}
\includegraphics[width=\linewidth]{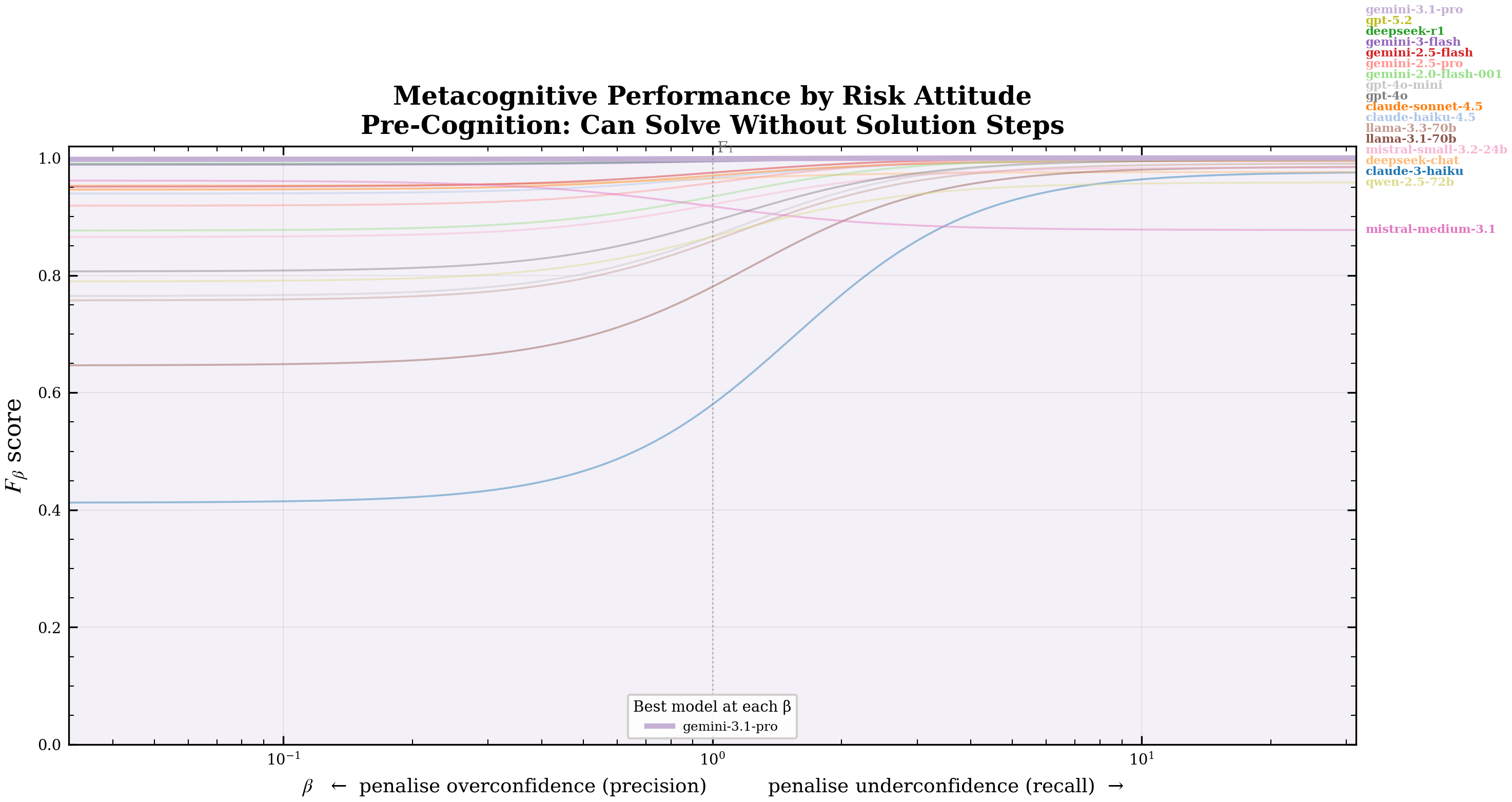}
\caption{MathBench (prospective)}
\end{subfigure}
\caption{Metacognitive $F_\beta$ score as a function of risk attitude $\beta$ (log scale). Low $\beta$ weights precision and penalises overconfidence; high $\beta$ weights recall and penalises underconfidence. The vertical reference at $\beta = 1$ is the standard $F_1$. On SQuAD no single model dominates and most scores fall in $[0.5, 0.85]$ across $\beta$. On MathBench Gemini 3.1 Pro dominates at every $\beta$, a consequence of uniform overconfidence across the rest of the model set.}
\label{fig:fbeta}
\end{figure}

\begin{figure}[t]
\centering
\includegraphics[width=\textwidth]{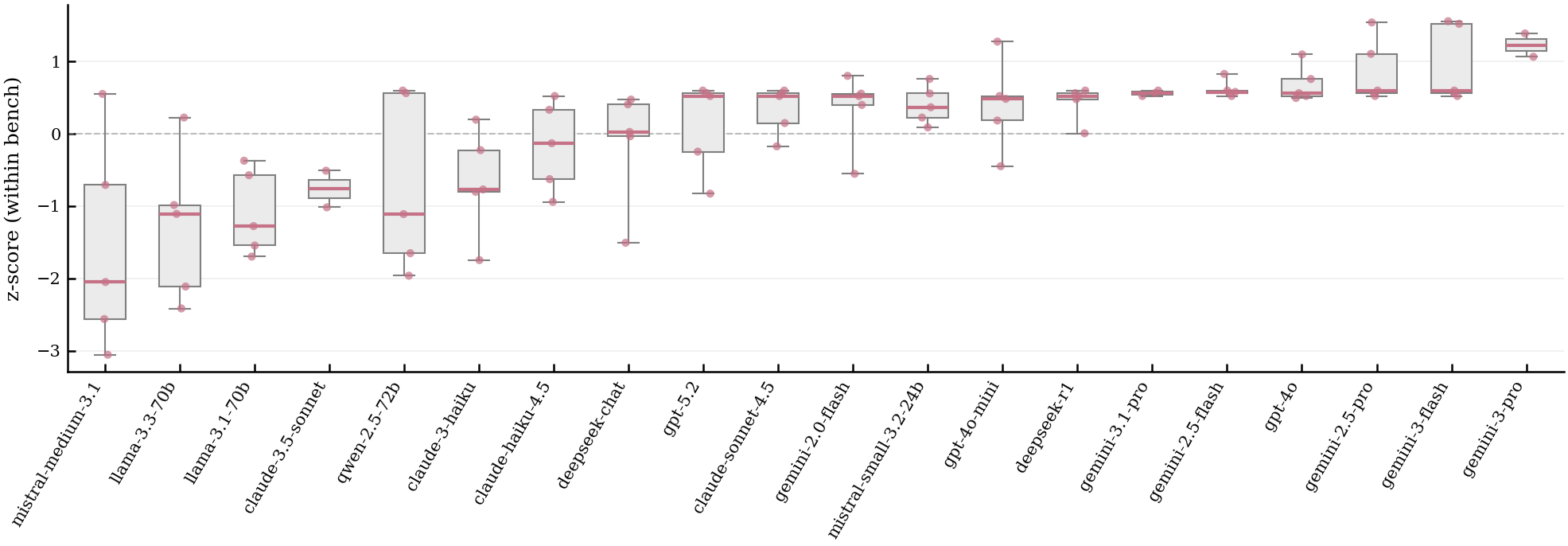}
\caption{Per-model confidence base rate (mean yes-rate, recovered as $\Phi(\text{threshold})$ from the tetrachoric factor analysis), z-scored within each benchmark and probe column. Five columns are shown (MathBench confidence-before, confidence-after, confidence-after-with-answer; SQuAD needs-context, context-necessary). Each box is one model's z-score distribution across columns; models are ordered left-to-right by mean z-score. Tight boxes indicate stable cross-benchmark confidence, wide boxes indicate task-dependent confidence. Median pairwise Pearson $r \approx 0.5$ between columns of yes-rates. Per-model confidence biases therefore travel across benchmarks, and the pairwise calibration analysis conditions on each model's marginal yes-rate to prevent these biases from inflating apparent metacognitive concordance. Provider trends are visible at family level, with the Gemini family consistently the most overconfident.}
\label{fig:confidence_consistency}
\end{figure}

First, aggregate confidence correlates poorly with aggregate performance (Figure~\ref{fig:calibration_all}, with the $F_\beta$ analysis in Figure~\ref{fig:fbeta}), motivating a structural rather than marginal analysis. Models are also consistent in their confidence relative to other models across benchmarks (Figure~\ref{fig:confidence_consistency}), foreshadowing the shared-mechanism finding.

\subsection{Tetrachoric Factor Analysis}

Our first aim is to recover the latent correlation structure of binary confidence judgements across models. A natural choice would be principal component analysis (PCA), but the binary nature of the responses makes this unsuitable. Two models that agree perfectly on which items are hard but commit to ``yes'' at different base rates can show arbitrarily low correlations when treated as continuous variables. We therefore turn to the tetrachoric correlation matrix.

\begin{definition}{Tetrachoric correlation}{tetrachoric}
Tetrachoric correlation~\citep{pearson1900} models each binary judgement as a thresholded continuous signal $Z_i \sim \mathcal{N}(0,1)$ with model-specific threshold $\tau_i$; $\rho_{\text{tet}}$ is the latent Pearson correlation between $Z_i$ and $Z_j$, estimated pairwise from each $2 \times 2$ contingency table (full estimator in \S\ref{app:si:tetrachoric}).
\end{definition}

Geometrically, the tetrachoric correlation is akin to the cosine similarity between the latent signals each model is thresholding, and is high when disagreements between two models are one-sided. When two models disagree on an item, it is almost always the higher-base-rate model saying ``yes'' while the lower says ``no''.

Our analysis of the tetrachoric correlation matrix proceeds in two steps. We first eigendecompose the correlation matrix, which yields the same variance-explaining decomposition as PCA on continuous variables. Eigenvalues are reported in normalized form $\lambda_i / n$ where $n$ is the number of models, summing to one across the spectrum. From the eigenspectrum we can understand the intrinsic dimensionality of the data per benchmark. A single dominant eigenvalue indicates a shared difficulty scale on which models read the same prompt cues and differ only in their thresholds. The single-factor signature is a dominant first eigenvalue paired with inverse-quadratic loading-vs-base rate dependence, since models with extreme yes-rates have anomalously low first-component loadings because near-floor and near-ceiling responders contribute little discriminating information per pair relative to noise. A flat spectrum is consistent with either individuated structure or noise, which we distinguish via the pairwise calibration analysis in \S\ref{sec:decomposition}. Per-benchmark factor-overview plots, the full mechanistic explanation, and quadratic-fit $R^2$ values per condition are in \S\ref{app:si:factor_overview} ( Figs.~\ref{fig:si:latent_fa-factor_overview:prospective} and~\ref{fig:si:latent_fa-factor_overview:counterfactual}).


\begin{figure}[t]
\centering
\begin{subfigure}[b]{0.32\textwidth}
\includegraphics[width=\textwidth]{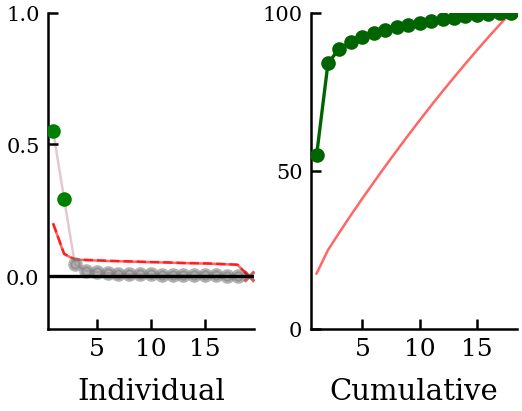}
\caption{SQuAD (prospective)}
\end{subfigure}
\hfill
\begin{subfigure}[b]{0.32\textwidth}
\includegraphics[width=\textwidth]{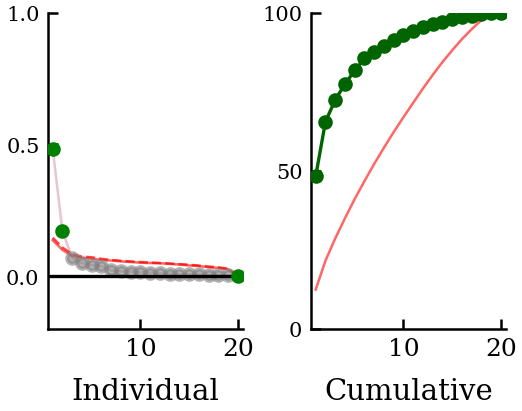}
\caption{MMLU-Pro (prospective)}
\end{subfigure}
\hfill
\begin{subfigure}[b]{0.32\textwidth}
\includegraphics[width=\textwidth]{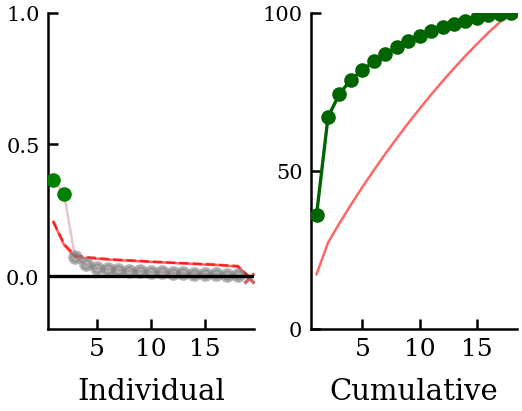}
\caption{LegalBench (prospective)}
\end{subfigure}

\vspace{0.5em}

\begin{subfigure}[b]{0.32\textwidth}
\includegraphics[width=\textwidth]{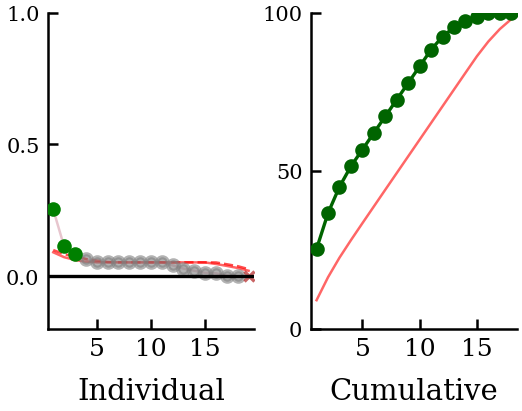}
\caption{MathBench (prospective)}
\end{subfigure}
\hfill
\begin{subfigure}[b]{0.32\textwidth}
\includegraphics[width=\textwidth]{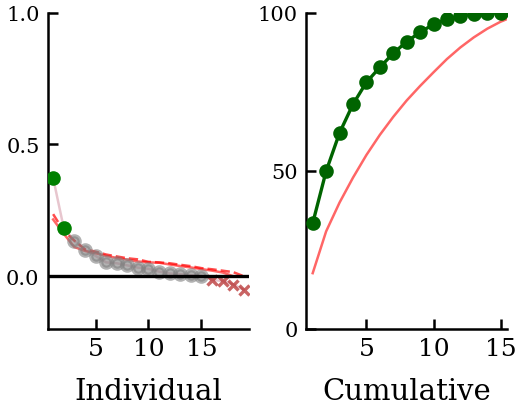}
\caption{SciCode (prospective)}
\end{subfigure}
\hfill
\begin{subfigure}[b]{0.32\textwidth}
\includegraphics[width=\textwidth]{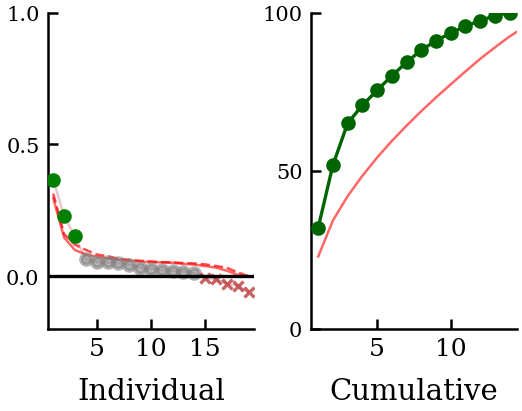}
\caption{OmniMath (prospective)}
\end{subfigure}
\caption{Eigenspectra of metacognitive tetrachoric correlation matrices (prospective conditions). Each panel shows the normalised per-rank eigenvalue (left axis, markers) with the per-rank 95th percentile of an empirical, base-rate-matched null overlaid as a red dashed curve, and the cumulative explained variance for the observed spectrum (green) and the null (pink) on the right. For each null sample, each model's column of binary judgments is independently shuffled to preserve its marginal yes-rate, the tetrachoric correlation matrix is re-estimated, and eigenvalues are taken; this is repeated $B=100$ times. Green markers are eigenvalues above the null (signal), grey markers are below (noise), and red crosses mark non-PSD negative eigenvalues where present. SQuAD and MMLU-Pro show a single eigenvalue well above the null with a sharp drop to the second eigenvalue; LegalBench has two leading eigenvalues that both clear the null. The reasoning benchmarks (MathBench, SciCode, OmniMath) have a long tail of eigenvalues sitting near or just above the null. SciCode and OmniMath additionally exhibit significant negative eigenvalues, a signature that the tetrachoric model's latent-normality assumption is violated.}
\label{fig:eigenspectra}

\end{figure}

All benchmarks (\cref{fig:eigenspectra}) exhibit a strong first eigenvalue, with SQuAD~\citep{rajpurkar2016} the strongest.
MMLU-Pro and LegalBench show moderate dominance with 3--4 effective factors.
On these factual tasks, models broadly agree about which items they will fail and the spectrum rolls off rapidly, with most eigenvalues below the empirical base-rate-shuffled null. MathBench and OmniMath do not fit this pattern. The spectrum tracks the empirical null closely rather than dropping sharply below it; there is no single shared difficulty axis, and different models show heterogeneous structure in which problems they expect to solve. The non-Gramian signature on these benchmarks indicates the latent-normality assumption is violated, so factor-analytic results for MathBench, SciCode, and OmniMath are treated as purely exploratory. Per-condition eigenspectra for both prospective (\cref{def:prospective}) and counterfactual (\cref{def:counterfactual}) probes are in Figs.~\ref{fig:si:latent_fa-eigenspectrum:prospective} and~\ref{fig:si:latent_fa-eigenspectrum:counterfactual}, and we found little difference between the two in overall structure. On retrieval and factual benchmarks, factor loading is parabolic in response threshold $\Phi^{-1}(\yesrate)$ with quadratic $R^2 \in [0.63, 0.95]$ (full per-condition fits in \S\ref{app:si:factor_overview}; the base-rate dependence is discussed in \S\ref{app:si:yesrate}), which is a sign of an inflated dimensionality due to low signal on extreme base rate models. This suggests the first and second eigenvalues are entangled and the true intrinsic dimensionality is lower.

\subsection{Pairwise Calibration}
A low-rank tetrachoric eigenbasis explains most of the variance, but some variance remains. Two questions follow. Does this shared difficulty factor track performance, and are the leftover eigenvalues noise or individuated structure? Analyzing the pairwise calibration between models allows us to answer both of these questions.

\begin{figure}[t]
\centering
\begin{subfigure}[b]{0.48\textwidth}
\includegraphics[width=\textwidth]{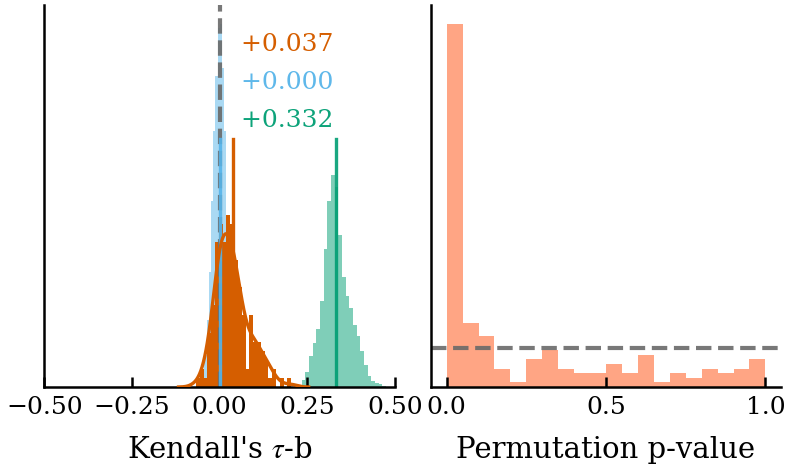}
\caption{SQuAD (prospective)}
\end{subfigure}
\hfill
\begin{subfigure}[b]{0.48\textwidth}
\includegraphics[width=\textwidth]{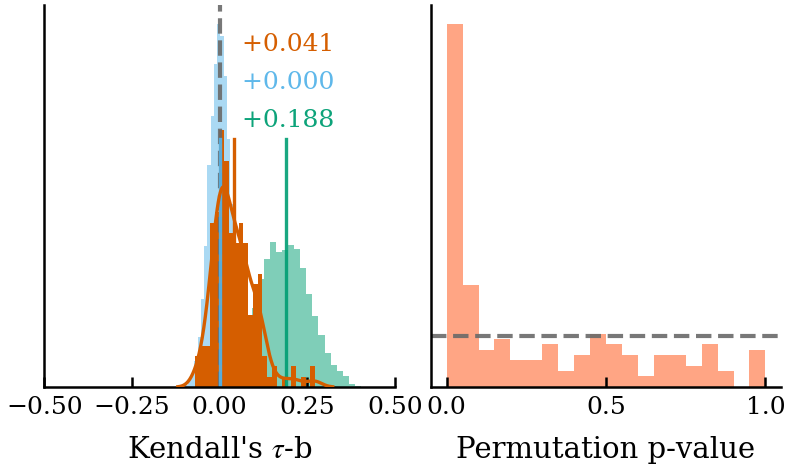}
\caption{MMLU-Pro (prospective)}
\end{subfigure}

\vspace{0.5em}

\begin{subfigure}[b]{0.48\textwidth}
\includegraphics[width=\textwidth]{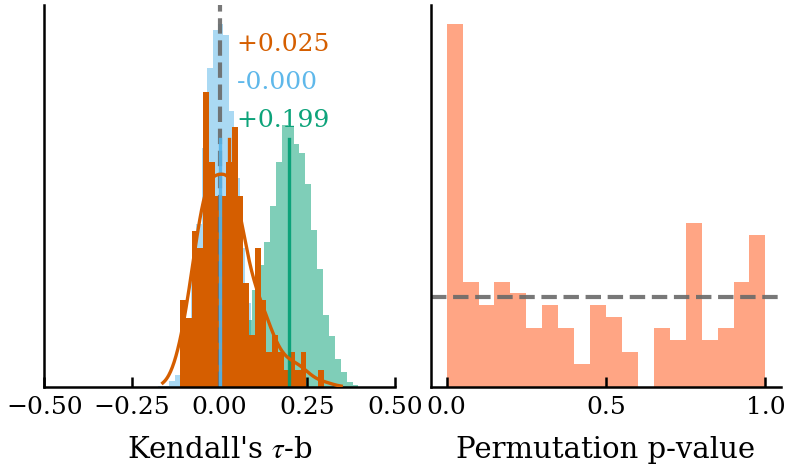}
\caption{OmniMath (prospective)}
\end{subfigure}
\hfill
\begin{subfigure}[b]{0.48\textwidth}
\includegraphics[width=\textwidth]{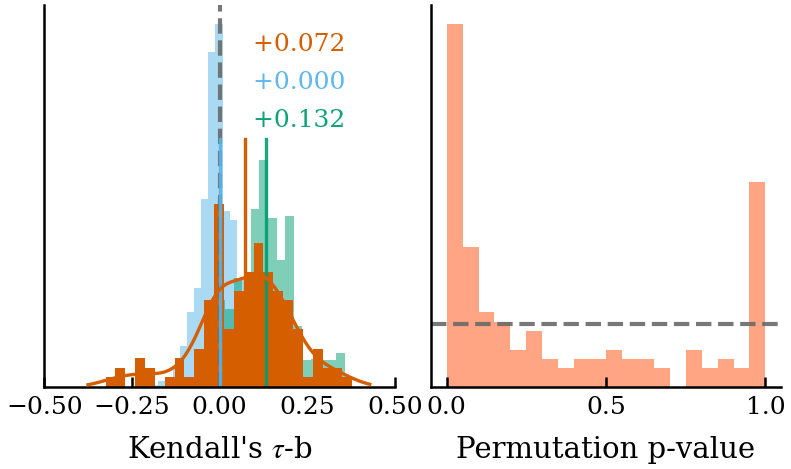}
\caption{MathBench (prospective)}
\end{subfigure}
\caption{Pairwise Kendall's $\tau$-b distributions across all model pairs, representative subset (full set in \cref{fig:si:tau_histogram:prospective,fig:si:tau_histogram:counterfactual}). Left panels: orange is observed $\tau$-b. Blue is a base-rate-matched null with each of $\text{perf}_A$, $\text{perf}_B$, $\text{conf}_A$, $\text{conf}_B$ shuffled independently. Green is a calibration-preserving null with each model's $(\text{perf}, \text{conf})$ pairs permuted together, holding each model's calibration profile fixed but randomising item identity. Solid verticals mark the three means; dashed grey marks $\tau=0$. Right panels: per-pair permutation $p$-value distribution against the dashed grey uniform-null reference. Effect size is small but statistically significant with more highly significant pairs than expected if sampling from the null distribution. With the exception of MathBench, the observed mean sits significantly below the calibration-preserving null, implying that almost all calibration is shared rather than individuated.}
\label{fig:tau_hist}

\end{figure}

\begin{definition}{Pairwise calibration}{pairwise}
The pairwise calibration between two models is the Kendall's $\tau$-b~\citep{kendall1945ties} between their confidence-difference and performance-difference vectors across items, using the tie-corrected $\tau$-b variant so that pairs the models agree on do not attenuate the correlation.
\end{definition}

This is pairwise calibration in the discriminatory sense (\cref{def:calibration}). For each model pair and each item, we ask whether model A actually outperforms model B when A says ``I can solve this'' and B says ``I cannot'' and vice versa. A \emph{concordant} pair is one where the more-confident model performs better; a \emph{discordant} pair is the reverse. Positive $\tau$ therefore means a model's confidence carries information about its own capability relative to the other model. Aggregating $\tau$ across all pairs forms a pairwise calibration distribution, and comparing it to null distributions from base-rate-matched item-permuted models (\S\ref{app:si:nulls}) allows us to compare the distributions in tau space and create a distribution of hypothesis tests for every pair relative to the empirical null.

The distribution of pairwise $\tau$  (\cref{fig:tau_hist}) is shown across a representative sample. On SQuAD ($\tau = 0.037$) and OmniMath ($\tau = 0.025$) the distribution is compact and modestly above the null, a real but small signal. MathBench stands apart, with a wider, right-shifted distribution ($\tau = 0.072$) in which several pairs exceed $\tau = 0.3$. Significance is assessed via probability of item level structure given the marginals through permutation tests, analytically resolved with Fisher's exact test~\citep{5d8fa06c-0d2a-391a-b97f-85223f1f3b0f} on the 3x3 contingency table. Per-benchmark distributions for all prospective and counterfactual probes are in Figs.~\ref{fig:si:tau_histogram:prospective} and~\ref{fig:si:tau_histogram:counterfactual}.

We also include a partially shuffled null distribution where individual calibrations are preserved but question IDs are permuted between models (the calibration-preserving null in \S\ref{app:si:nulls}), providing us with what the pairwise null would be if there was no shared calibration signal. This null distribution shows significantly higher pairwise calibration than the real signal suggesting that almost all calibration is shared between models. This is most obvious for SQuAD where the two distributions are cleanly separated.

\begin{figure}[t]
\centering
\begin{subfigure}[b]{0.32\textwidth}
\includegraphics[width=\textwidth]{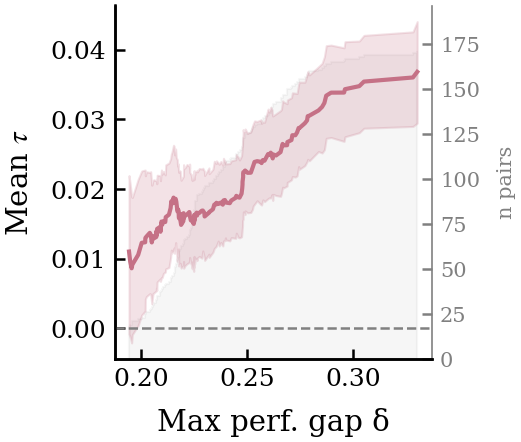}
\caption{SQuAD (prospective)}
\end{subfigure}
\hfill
\begin{subfigure}[b]{0.32\textwidth}
\includegraphics[width=\textwidth]{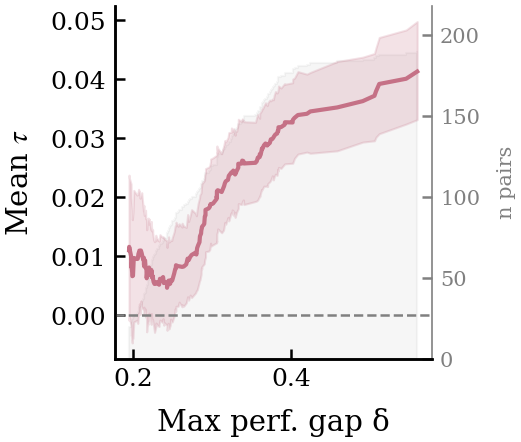}
\caption{MMLU-Pro (prospective)}
\end{subfigure}
\hfill
\begin{subfigure}[b]{0.32\textwidth}
\includegraphics[width=\textwidth]{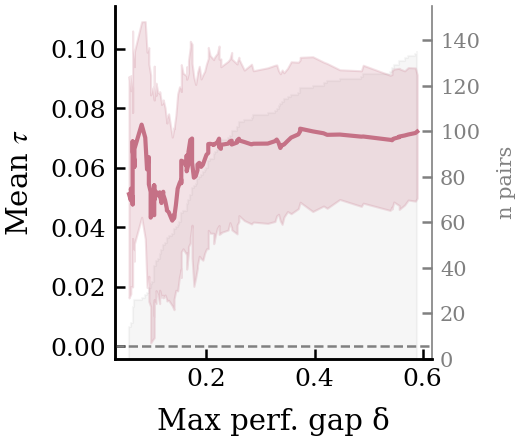}
\caption{MathBench (prospective)}
\end{subfigure}
\\[1ex]
\begin{subfigure}[b]{0.48\textwidth}
\includegraphics[width=\textwidth]{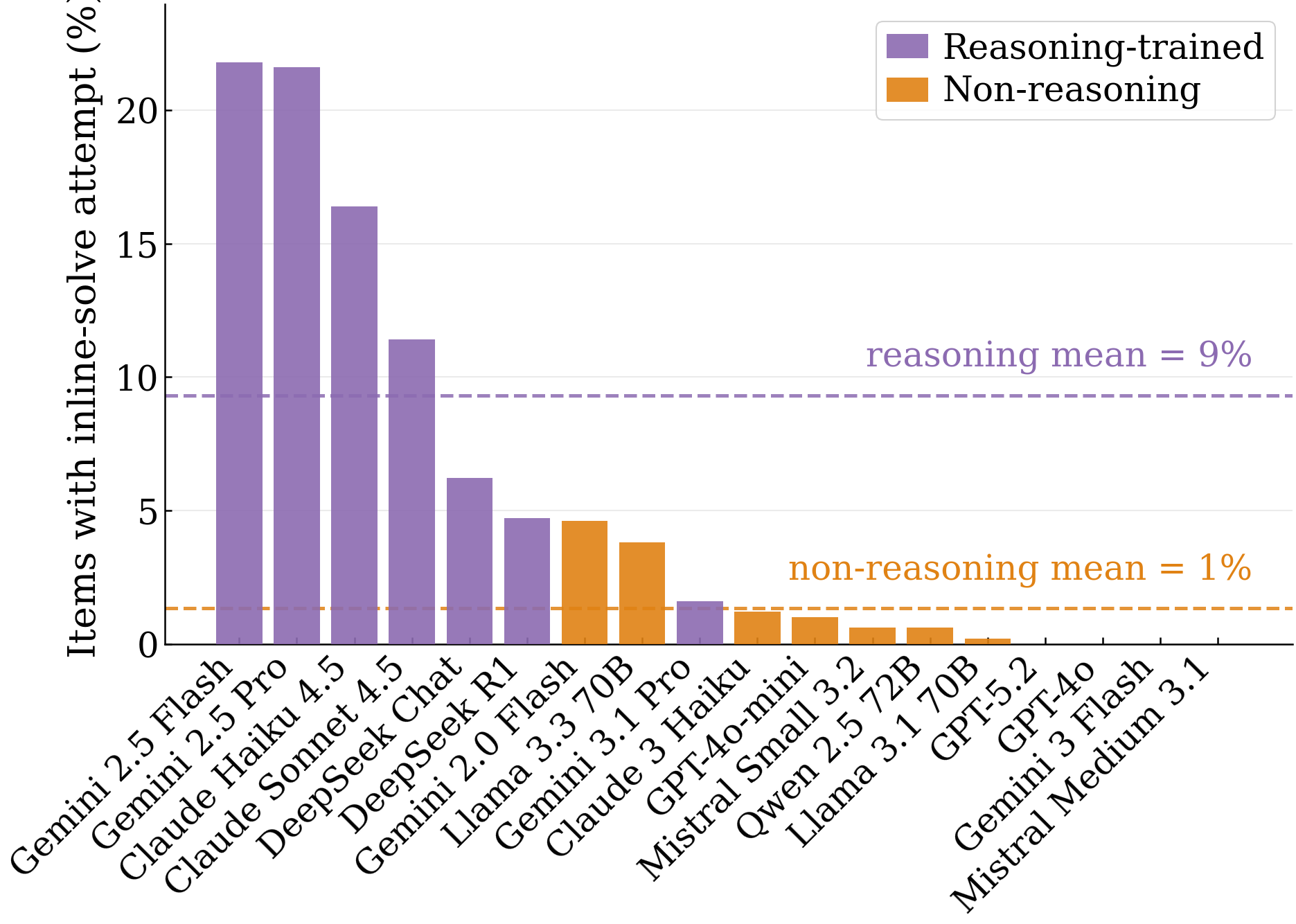}
\caption{MathBench inline solve rate by model (prospective)}
\end{subfigure}
\hfill
\begin{subfigure}[b]{0.48\textwidth}
\includegraphics[width=\textwidth]{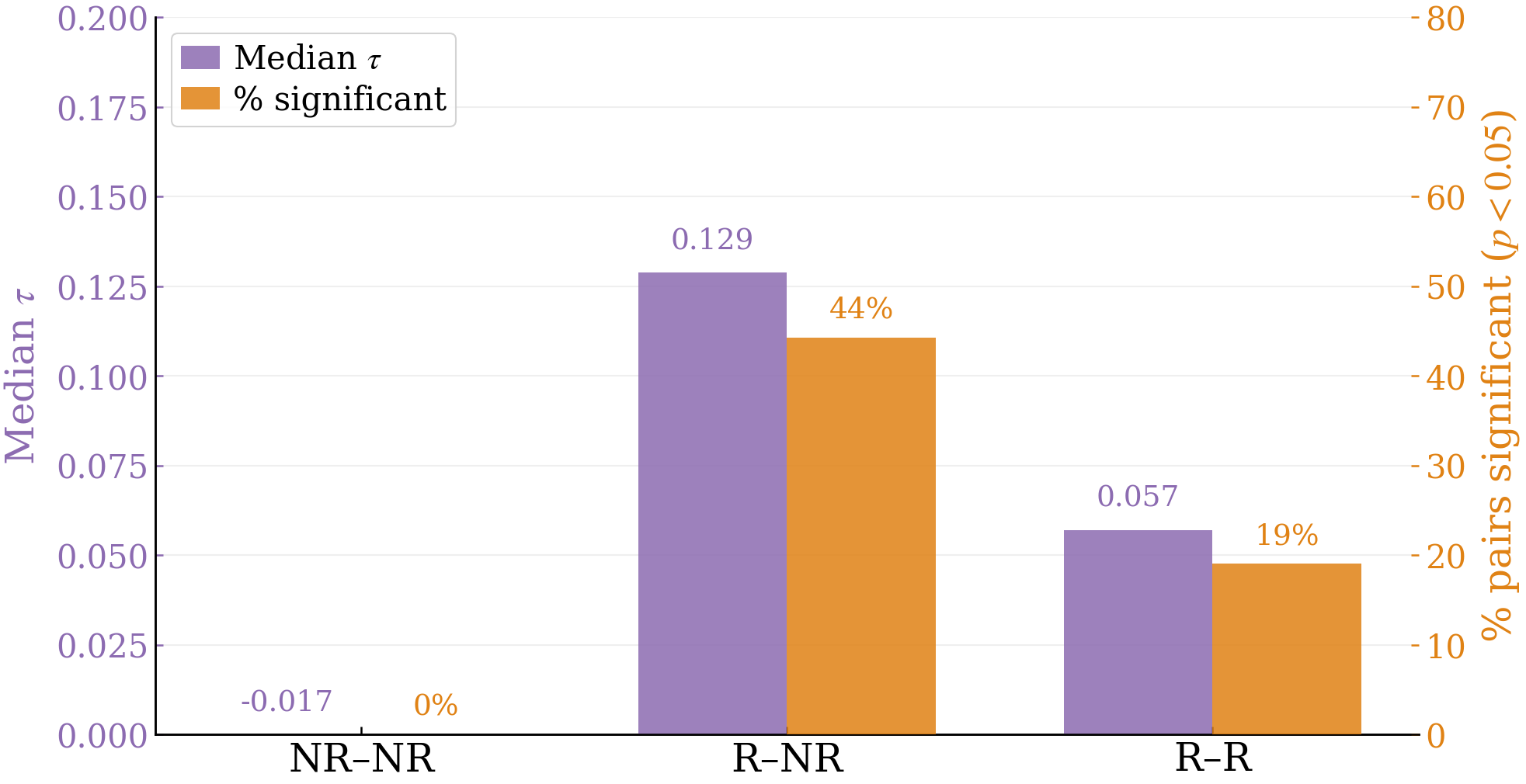}
\caption{MathBench concordance by training regime (prospective)}
\end{subfigure}

\caption{Mean pairwise Kendall's $\tau$ as a function of the maximum allowed performance base-rate difference $\delta$. \textbf{(a--b)} SQuAD and MMLU-Pro. Pink curve is mean $\tau$ on the left axis, grey shaded region is the number of admitted pairs on the right axis. Restricting to similar-capability pairs makes calibration vanish, consistent with performance base-rate differences projected onto the shared difficulty axis. \textbf{(c)} MathBench. $\tau$ remains flat across the full $\delta$ range, so even among matched-accuracy pairs, confidence differences predict which model handles specific items better. \textbf{(d)} Inline solve rate per model during prospective confidence elicitation, with reasoning-trained models in purple and non-reasoning models in orange. \textbf{(e)} Individuated metacognition on MathBench is concentrated in Reasoning--Non-Reasoning pairs. Purple bars on the left axis are median $\tau$ within each pair type; orange bars on the right axis are the share of pairs reaching $p < 0.05$. Median $\tau$ and the significant-pair share are both substantially smaller in R--R and NR--NR pairs than in R--NR pairs.}
\label{fig:climax}
\end{figure}

The effect size is tiny, and minuscule compared to the individual calibration preserving null, but there is a statistically significant signal that requires further investigation. Models with different base-rate performance align naturally - for the obvious edge case consider a model that refuses to answer questions and always says it cannot do them - such a model would have arbitrarily high pairwise calibration. If this is the only mechanism, it is a falsifiable claim that inter-model calibration stops being statistically significant for models of equal performance. This is what we find (\cref{fig:climax}) for SQuAD and every other benchmark, with the exception of MathBench, but this is a confound.

For MathBench we stratify models by post-training regime, based on publicly available documentation. \emph{Reasoning-trained} models have explicit chain-of-thought or verifiable-reward post-training (DeepSeek-R1, Gemini 2.5/3.x, Claude 4.x, GPT-5.2, DeepSeek Chat); \emph{non-reasoning} models have no dedicated reasoning post-training. Stratifying pairs by training regime, individuated metacognition is concentrated in Reasoning--Non-Reasoning pairs (median $\tau = 0.129$), while Reasoning--Reasoning and Non-Reasoning--Non-Reasoning pairs are not statistically significant. Investigating further we discover that for a strong plurality of items with high pairwise calibrations, the reasoning models were simply solving them inline and using the results of their attempt to determine their abilities.

\begin{definition}{Inline-solve detection}{inlinesolve}
A rule-based detector flags a response as an \emph{inline solve} when it scores above threshold on a five-feature rubric covering solution-convergence phrases, mathematical notation, computation markers, numbered steps, and length (full rubric in \S\ref{app:si:inline_solve_detector}).
\end{definition}

Reasoning models that solve inline carry item-level signal because their confidence reflects an attempt, but inline solving is the symbolic counterpart of the sub-symbolic noetic capacity we seek to measure (\cref{def:subsymbolic}).~\cite{podolak2025read} and~\cite{yoon2025reasoning} document the same dependence of confidence accuracy on in-CoT computation within single reasoning models, with~\cite{yoon2025reasoning} reading it as a calibration capability of reasoning models. Our pairwise stratification in Figure~\ref{fig:climax}(e) adjudicates between these readings. When both models in a pair can solve inline, the confidence difference carries no statistically significant signal, as expected if both are computing rather than introspecting.

\begin{definition}{Sub-symbolic (noetic) metacognition}{subsymbolic}
\emph{Sub-symbolic metacognition} is noetic awareness in the sense of~\cite{tulving1985noetic}, an assessment of one's own capabilities without exercising them. Inline solving is its symbolic counterpart, where the model runs the same forward reasoning it would use to answer the question and reports the outcome as confidence.
\end{definition}

\subsection{Comparing Performance and Confidence Factors}
\label{sec:alignment}

Models largely agree about what they are likely to succeed at but we find performance is itself low-rank, as has previously been reported~\citep{kipnis2025metabench} (Figs.~\ref{fig:si:eigenanalysis_perf-eigenspectrum:prospective} and~\ref{fig:si:eigenanalysis_perf-eigenspectrum:counterfactual}). As such, the low-rank confidence results may be the appropriate structure, which requires interrogation.

For each item we compute its position on the dominant performance axis and the dominant confidence axis, thus achieving the equivalent outcome to projecting items onto the PCA basis (per-item Factor 1 scores; method in \S\ref{app:si:filtered_alignment}) and correlate the two across items using ordinary least squares regression.

\begin{definition}{Contentious-item subset}{contentious}
The items whose cross-model performance rate and cross-model confidence rate both fall within $[0.1, 0.9]$, isolating items with substantial inter-model disagreement on both axes.
\end{definition}

\begin{figure}[t]
\centering
\begin{subfigure}[b]{0.48\textwidth}
\includegraphics[width=\textwidth]{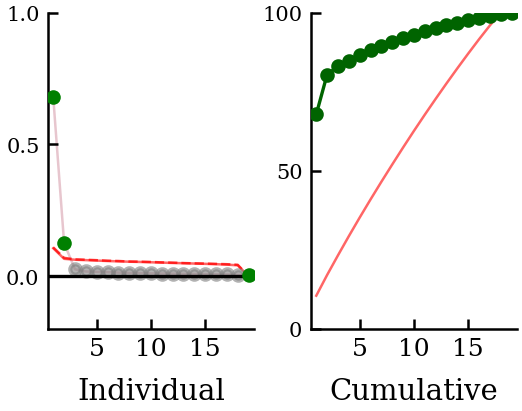}
\caption{SQuAD performance eigenspectrum (prospective)}
\end{subfigure}
\hfill
\begin{subfigure}[b]{0.48\textwidth}
\includegraphics[width=\textwidth]{figures/squad/needs_context/latent_fa/eigenspectrum_clean.png}
\caption{SQuAD confidence eigenspectrum (prospective)}
\end{subfigure}
\\[1ex]
\begin{subfigure}[b]{0.48\textwidth}
\includegraphics[width=\textwidth]{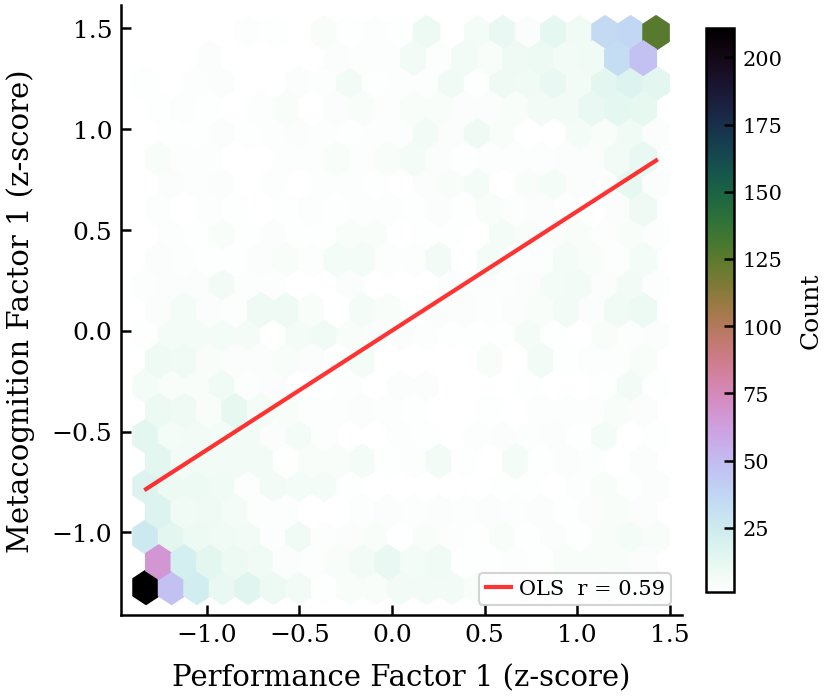}
\caption{SQuAD F1 alignment, unfiltered (prospective)}
\end{subfigure}
\hfill
\begin{subfigure}[b]{0.48\textwidth}
\includegraphics[width=\textwidth]{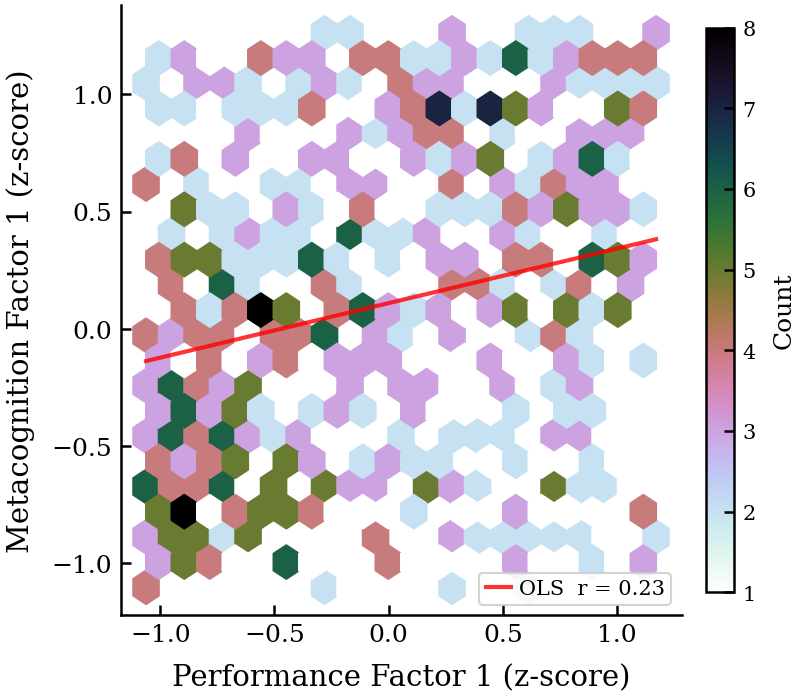}
\caption{SQuAD F1 alignment, filtered (prospective)}
\end{subfigure}
\caption{SQuAD performance and confidence are each approximately rank-one along distinct axes. \textbf{(a, b)} Performance and confidence tetrachoric eigenspectra. PC1 captures ${\sim}68\%$ of the performance variance and ${\sim}55\%$ of the confidence variance, with sharp drops to PC2 in both cases. Panel (b) reproduces a panel of Fig.~\ref{fig:eigenspectra}. \textbf{(c)} Per-item Factor 1 scores from the two matrices correlate at $r = 0.591$ across all items. \textbf{(d)} On the items in the contentious-item subset (\cref{def:contentious}), alignment falls to $r = 0.230$. The unfiltered correlation is concentrated on items every model gets right or wrong. On items with inter-model disagreement, the confidence axis tracks a surface-cue proxy for difficulty (length, vocabulary, perceived complexity) rather than the latent difficulty axis that performance lives on.}
\label{fig:filtered}
\end{figure}

We report two values per benchmark: the unfiltered Pearson $r$ across all items, and the filtered $r$ on the contentious-item subset (full procedure in \S\ref{app:si:filtered_alignment}). SQuAD prospective shows $r = 0.591$ unfiltered, reducing to $r = 0.230$ on the contentious-item subset (Fig.~\ref{fig:filtered}). Across the benchmarks the filtered $r$ shrinks toward zero or flips sign (Table~\ref{tab:alignment_full}). The agreement between the performance factor and the confidence factor is concentrated on items where most models agree, either at the floor or ceiling. On contentious items the axes do not coincide. This combines with the pairwise analysis to suggest that what signal exists is shared, and even then that signal is not particularly strong on a per-item level.

\begin{table}[t]
\centering
\caption{Pearson $r$ and $R^2 = r^2$ between per-item Factor~1 scores on the performance and metacognitive matrices, unfiltered and on the contentious-item subset (\cref{def:contentious}). SQuAD prospective is the row instantiated as Fig.~\ref{fig:filtered} panels (c, d).}
\label{tab:alignment_full}
\small
\begin{tabular}{llcccc}
\toprule
\rowcolor{PBoxPrimary}
Benchmark & Probe & Unfilt.\ $r$ & Unfilt.\ $R^2$ & Filt.\ $r$ & Filt.\ $R^2$ \\
\midrule
LegalBench & Prospective & $+0.038$ & $0.001$ & $-0.028$ & $0.001$ \\
MathBench & Prospective & $+0.209$ & $0.044$ & $-0.000$ & $0.000$ \\
MathBench & Counterfactual & $+0.360$ & $0.130$ & $+0.344$ & $0.118$ \\
MMLU & Prospective & $+0.376$ & $0.141$ & $+0.189$ & $0.036$ \\
MMLU & Counterfactual & $-0.312$ & $0.097$ & $-0.148$ & $0.022$ \\
OmniMath & Prospective & $+0.431$ & $0.186$ & $+0.075$ & $0.006$ \\
OmniMath & Counterfactual & $+0.495$ & $0.245$ & $+0.140$ & $0.020$ \\
SciCode & Prospective & $+0.277$ & $0.077$ & $+0.149$ & $0.022$ \\
SciCode & Counterfactual & $+0.085$ & $0.007$ & $+0.056$ & $0.003$ \\
SQuAD & Prospective & $+0.591$ & $0.349$ & $+0.230$ & $0.053$ \\
SQuAD & Counterfactual & $+0.290$ & $0.084$ & $+0.032$ & $0.001$ \\
\bottomrule
\end{tabular}
\end{table}

\subsection{Probing Unused Information}
\label{sec:residual}

To overcome potential noise from single shot binary elicitation, we move from the ``wide'' cross-model decomposition of 20 models with one response each to a ``tall'' analysis. For a representative model on each benchmark, selected to have performance and confidence base rates close to the benchmark median to avoid artefacts of floor or ceiling effects (Table~\ref{tab:calibration_roc}), we run the same confidence probe 20 times at high temperature and measure how well the empirical yes-rate predicts actual correctness.

We compare three predictors of correctness for each representative model. \textbf{Internal} is the model's empirical yes-rate over 20 samples at temperature 1. \textbf{Population} is the cross-model consensus, the average of every model's single-shot binary confidence judgement. The \textbf{external classifier} is a deliberately trivial ridge logistic regression on hand-crafted surface features of the confidence-judgement reasoning trace, counting hedging adverbs (``maybe''), causal connectives (``because''), insight markers (``think''), self-correction phrases (``wait''), a 25-word lexicon of incorrectness-predictive chain-of-thought tokens (``stuck'', ``guess'') taken from prior work, and basic length, sentence-count, and digit-density features. The full inventory and per-benchmark coefficients are in \S\ref{app:si:external_classifier}.

In all but one case where the performance is near identical, the population aggregate outperforms the individual model at estimating its own performance. The external classifier is competitive with or beats the model's own sampled confidence score in area under the receiver operating characteristic curve (AUC) on three of six benchmarks (LegalBench, MathBench, MMLU; Table~\ref{tab:calibration_roc}). An analogous classifier reading the answer-attempt text instead wins on two further benchmarks (SciCode, OmniMath). Both are deliberately weak surface-feature baselines, so the gap lower-bounds the correctness-predictive signal that the binary tokenisation discards from text the model itself produced. SQuAD/GPT-4o is the only configuration where no external classifier exceeds Internal. Representative ROC curves are in Fig.~\ref{fig:si:roc_comparison}, and a cross-benchmark transfer matrix in Fig.~\ref{fig:si:transfer}.

\begin{table}[t]
\centering
\caption{ROC AUC for predicting correctness on the no-help condition. Internal is the model's binary self-assessment. Pop is the cross-model consensus probability. The external-classifier feature subsets read the question text (Q), the model's confidence-judgement reasoning trace (M), and the model's answer-attempt text (P). QM is the like-for-like comparison to Internal because both have access to the same pre-judgement context. QP and QMP read post-hoc answer text and bound from above. Bold marks the best predictor per row.}
\label{tab:calibration_roc}
\small
\setlength{\tabcolsep}{4pt}
\begin{tabular}{llccccccc}
\toprule
\rowcolor{PBoxPrimary}
Benchmark & Model & Internal & Pop & Q & QM & QP & QMP \\
\midrule
LegalBench & Gemini Flash      & 0.474 & 0.532          & 0.662 & 0.788 & 0.766 & \textbf{0.861} \\
MathBench  & GPT-4o            & 0.630 & 0.742          & 0.634 & 0.703 & 0.756 & \textbf{0.773} \\
MMLU       & GPT-4o            & 0.632 & 0.630          & 0.635 & 0.706 & 0.657 & \textbf{0.717} \\
SciCode    & Llama 3.1-70B     & 0.590 & 0.657          & 0.543 & 0.539 & \textbf{0.858} & 0.731 \\
OmniMath   & Mistral Medium 3.1 & 0.664 & 0.728         & 0.564 & 0.619 & \textbf{0.754} & 0.718 \\
SQuAD      & Claude 3 Haiku    & 0.622 & \textbf{0.750} & 0.536 & 0.657 & 0.582 & 0.664 \\
SQuAD      & GPT-4o            & 0.731 & \textbf{0.751} & 0.502 & 0.683 & 0.579 & 0.691 \\
\bottomrule
\end{tabular}
\end{table}

\section{Discussion}
\label{sec:discussion}

On retrieval and factual tasks, difficulty is surface-readable and shared. Models agree on what is hard, the population signal dominates, and no individuated structure survives. Mathematical reasoning superficially looks different, with pairwise calibration surviving at matched accuracy and a high-dimensional factor space, but the mechanism is computation rather than introspection. Reasoning-trained models solve problems inline during confidence elicitation (Figure~\ref{fig:climax}d), and the signal among Reasoning--Reasoning pairs where both models can do so is not statistically significant.

That reasoning models apparent inter-model calibration was a symbolic confound was a disappointing discovery. We hypothesized that reasoning models would show emergent sub-symbolic metacognition, on the grounds that knowing which techniques one can and cannot execute correctly is itself useful for problem-solving. If such emergence is possible, reinforcement learning from verifiable rewards (RLVR) appears to suppress it in favor of active self interrogation. Hedging guarantees zero reward, since uncertainty is not a correct answer, and being confidently wrong returns the same zero, so overconfidence has strictly positive expected reward under any signal that does not separately price uncertainty. Variants that fold listener-aware, calibrated-confidence, or token-level self-certainty terms into post-training rewards~\citep{stengeleskin2024lacie,damani2025beyond,zhao2026intuitor} push against this. The cross-benchmark consistency of confidence base rates we observe (median pairwise Pearson $r \approx 0.5$; Figure~\ref{fig:confidence_consistency}) is consistent with confidence expression being a model-level policy reshapable by such interventions. Whether any such shaping produces sub-symbolic metacognition distinct from the inline-computation pathway documented here remains open. We do not find it in our analysis.

\section{Limitations}
\label{sec:limitations}
\begin{itemize}
\item We elicit one binary yes/no judgement per (model, item) trial at temperature 0. On uncertain items this masks the variance in the model's responses at higher temperature. We take the zero-temperature response as the model's position, which may be a marginal preference that becomes high variance with increased temperature.

\item The reasoning/non-reasoning stratification (Section~\ref{sec:decomposition}) relies on public documentation of training procedures, so borderline models may be miscategorised and this uncertainty propagates into the regime-stratified $\tau$ estimates. The inline-solve detection rubric is similarly heuristic but intentionally so as the category is fuzzy.
\item The tetrachoric latent-normality assumption is presumed rather than tested. The bivariate-normal latent model is almost certainly wrong on the Gaussian assumption, and base-rate noise in the marginals isn't modeled so inflates dimensionality for extreme base rates.
\item It is a well known problem that many benchmarks contain ambiguous and unanswerable questions. This is likely amplified by the experimental choice to remove optional context that may be de facto load bearing. Arguably the ability to infer the authors intent accurately is a skill as any other, but it is not what benchmarks test and it would be undesirable for this to be a significant part of any signal.
\end{itemize}

\section{Compute and access}
Experiments are API-driven. The wide decomposition consumes on the order of $2 \times 10^{5}$ calls and the tall analysis adds one representative model per benchmark sampled 20 times at high temperature, totalling approximately \$4{,}000 in spend. Full per-provider details, model identifiers, and caching protocol are in \S\ref{app:si:compute}.

\vspace{\PreConclusionEm em}
\section{Conclusion}

Across six benchmarks and 20 models, LLM confidence largely reduces to a shared difficulty heuristic on knowledge tasks, with reasoning tasks showing more complex structure that nonetheless does not produce individuated metacognition. Models agree on which items are hard but fail to predict their own relative advantage. Performance is itself approximately rank-one, so a rank-one confidence signal is naively optimal, but the two rank-one axes coincide only on items everyone agrees on and diverge precisely where individuated metacognition would matter. The MathBench exception resolves to a confound, since pairwise calibration survives at matched performance only because reasoning-trained models solve inline during elicitation, and the signal is statistically significant only between reasoning and non-reasoning pairs. A trivial classifier on the confidence-judgment reasoning text matches or beats Internal on three of six benchmarks (\S\ref{app:si:external_classifier}), so information is present that the models are not using. We find no evidence for significant individuated metacognition in any domain tested under binary probes.

\clearpage

\section*{Author contributions}

Author contributions follow the CRediT (Contributor Roles Taxonomy)
system. M.M.\ and M.W.\ contributed to \emph{conceptualization},
\emph{project administration}, and \emph{writing -- review and editing}.
M.M.\ contributed to \emph{methodology}, \emph{software},
\emph{validation}, \emph{formal analysis}, \emph{investigation},
\emph{data curation}, \emph{writing -- original draft}, and
\emph{visualization}. M.W.\ contributed to \emph{resources} and
\emph{funding acquisition}.

\bibliography{bibliography}

\appendix

\renewcommand{\thetable}{\thesection.\arabic{table}}
\renewcommand{\thefigure}{\thesection.\arabic{figure}}
\makeatletter
\@addtoreset{table}{section}
\@addtoreset{figure}{section}
\makeatother
\setcounter{table}{0}
\setcounter{figure}{0}

\clearpage

\section{Additional analyses}
\label{app:si}

\subsection{External classifier methodology}
\label{app:si:external_classifier}

The external classifier in \S\ref{sec:residual} is a ridge logistic regression (sklearn default $C{=}1.0$, $5$-fold \texttt{GroupKFold} on item ids) on hand-crafted surface-lexicon features, fit separately for each (benchmark, model) cell. Four feature subsets are evaluated, named by their text sources. \textbf{Q} is the question, \textbf{M} is the model's confidence-judgement reasoning trace, \textbf{P} is the model's answer-attempt text.

\begin{itemize}
\item \textbf{Q}, 19 features of the question text (length, readability, type indicators). Baseline.
\item \textbf{QM}, Q plus 21 features of the confidence-judgement reasoning trace. The reasoning trace is the text the model produced before emitting its yes/no judgement, so QM has access to the same context the binary commit had at decision time.
\item \textbf{QP}, Q plus 21 features of the answer-attempt text. Answer-attempt text is post-hoc relative to the prospective binary, so QP is an upper bound on what surface features of the answer attempt could convey rather than a like-for-like baseline.
\item \textbf{QMP}, all three feature groups combined.
\end{itemize}

The 21 response-side features cover hedging and modal/uncertainty adverbs, a 25-word ``harmful''-language lexicon, causation/tentativeness/insight categories, and structural features (sentence count, hedging-sentence rate, distinct-trigram ratio, digit count, math-operator count). The classifiers are weak by construction. A positive QM-versus-Internal gap therefore lower-bounds the correctness-predictive signal the binary tokenisation discards from reasoning context the model itself had access to. A richer (e.g.\ embedding-based) classifier would only widen the gap.

\noindent\textbf{Q feature list (19 features of the question text).}
\begin{itemize}
\item \textbf{Length / structure (4):} character count, word count, sentence count, mean word length.
\item \textbf{Lexical (1):} type-token ratio.
\item \textbf{Content flags (4):} indicator for any negation; indicator for any mathematical notation; indicator for any code-like markers; count of constraint markers (``must'', ``cannot'', ``at least'', etc.).
\item \textbf{Readability (1):} Flesch-Kincaid grade level.
\item \textbf{Multiple choice (1):} number of distinct A/B/C/D-style choice markers in the rendered prompt.
\item \textbf{Question-type one-hot (8):} indicators for ``what'', ``how'', ``why'', ``which'', ``compute'', ``find'', ``explain'', and ``other''.
\end{itemize}

\noindent\textbf{M and P feature list (21 features each, same definitions).}
M features are extracted from the model's confidence-judgment reasoning trace; P features are extracted from the model's answer-attempt text. The two feature sets are structurally identical but compute on different text sources. Items below give one feature name per bullet.
\begin{itemize}
\item \textbf{Length / structure (4):} character count, word count, sentence count, mean sentence length.
\item \textbf{Lexical diversity (2):} distinct-bigram ratio, distinct-trigram ratio.
\item \textbf{Hedging and certainty (4):} count of modal/uncertainty hedging words; hedging-sentence rate (share of sentences containing at least one hedge); count of certainty words; count of causal connectives.
\item \textbf{Content lexicons (5):} count of self-correction markers; count of a 25-word ``harmful'' chain-of-thought lexicon; count of insight markers (``think'', ``realise'', ``know''); count of difficulty markers; count of ``unknown''-style phrases (``not sure'', ``cannot tell'').
\item \textbf{Structural / numeric (6):} count of negations; count of first-person pronouns; count of step markers (``first'', ``next'', numbered steps); count of question marks; count of digit characters; count of arithmetic operators.
\end{itemize}
QM therefore has $19 + 21 = 40$ features, QP has $19 + 21 = 40$, and QMP has $19 + 21 + 21 = 61$.

\begin{figure}[H]
\centering
\includegraphics[width=\textwidth]{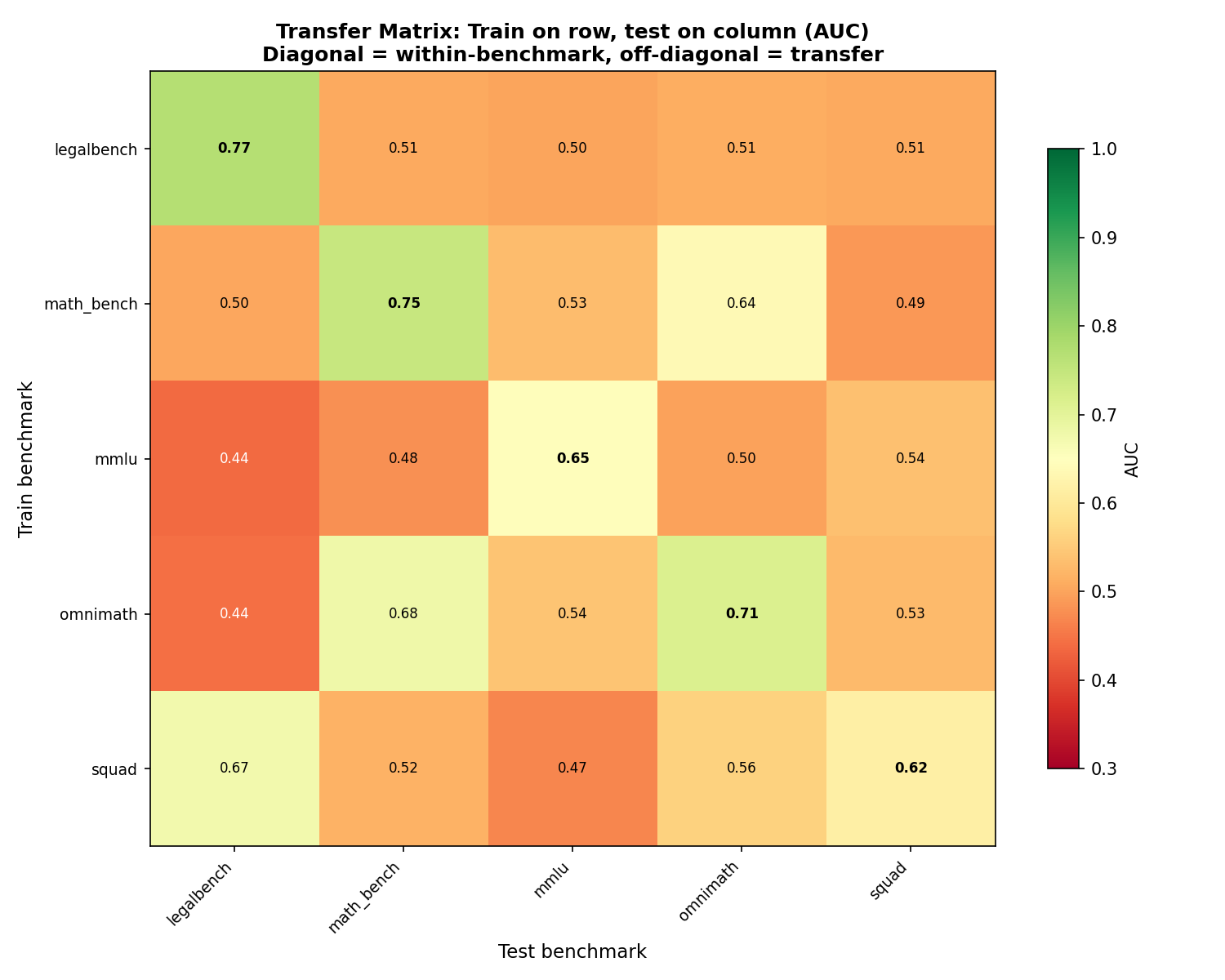}
\caption{Transfer matrix for the external classifier on confidence-judgement reasoning features, trained on the row benchmark and tested on the column benchmark. Five of the six paper benchmarks are shown. SciCode is omitted because the pooled feature matrix could not be assembled with current model coverage. Diagonal AUCs span 0.62--0.77. The MathBench~$\leftrightarrow$~OmniMath transfers are symmetric (AUC 0.64 and 0.68, mean 0.66), so the computation pair behaves like a same-domain cluster. The factual-recall pair is asymmetric: SQuAD$\to$LegalBench reaches AUC 0.67 but LegalBench$\to$SQuAD only 0.51. All other off-diagonals sit near 0.50. The reasoning-trace signature carries surface cues that travel between the two computation benchmarks, but transfer across the computation--recall divide is at chance.}
\label{fig:si:transfer}
\end{figure}

\noindent\textbf{Top features.} The largest single QMP coefficient is on confidence-judgement reasoning length. It is negative on every benchmark, so longer self-assessment reasoning predicts incorrect outcomes. Answer-side negation count, digit count, and math-operator count contribute next on math and code benchmarks. The ``harmful word'' lexicon contributes consistently to reasoning-trace coefficients. Reasoning-length features flip sign across domains, negative on math and near zero on retrieval. Classifier transfer therefore tracks task similarity rather than benchmark identity. Transfers within the computation cluster (MathBench~$\leftrightarrow$~OmniMath) and within the factual-recall cluster (LegalBench~$\leftrightarrow$~SQuAD) preserve most of the diagonal signal. Transfers across the computation--recall divide cluster near chance (Figure~\ref{fig:si:transfer}).

\subsection{ROC curves for internal and external predictors}
\label{app:si:roc}

\begin{figure}[H]
\centering
\begin{subfigure}[b]{0.48\textwidth}
\includegraphics[width=\textwidth]{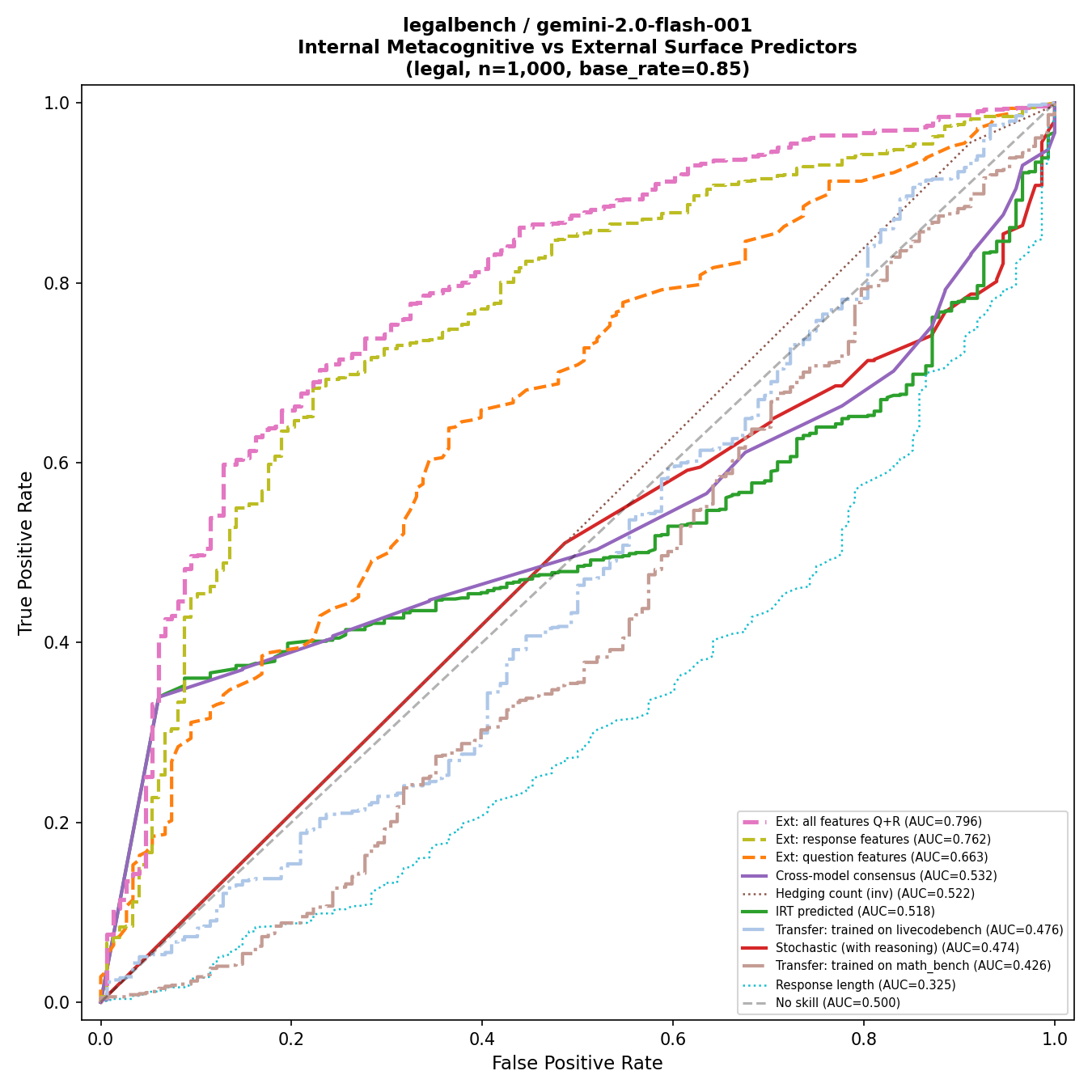}
\caption{LegalBench, Gemini 2.0 Flash (prospective). The QM lexicon classifier reaches AUC $0.79$ against an internal binary commit of $0.47$, an $+0.32$ gap.}
\end{subfigure}
\hfill
\begin{subfigure}[b]{0.48\textwidth}
\includegraphics[width=\textwidth]{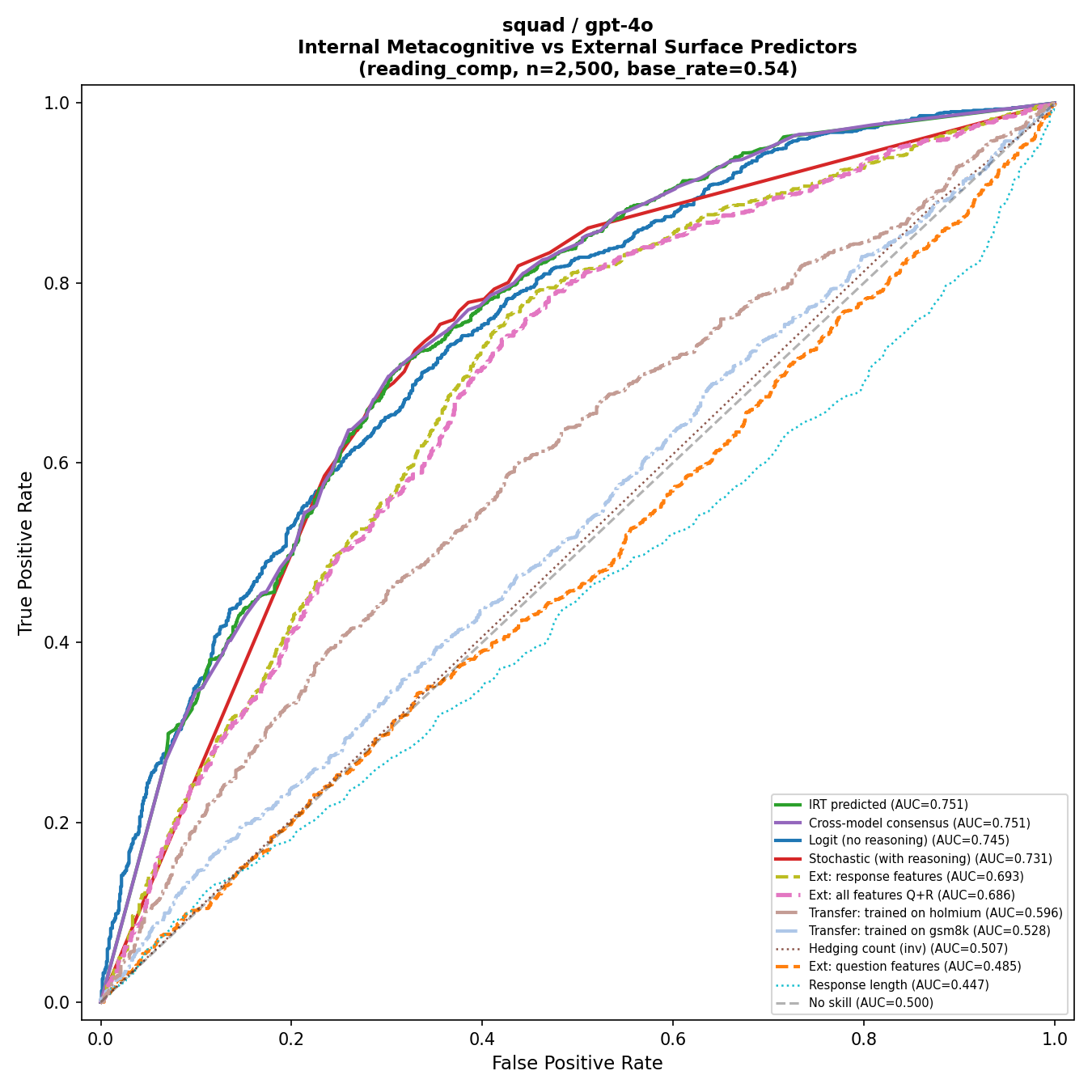}
\caption{SQuAD, GPT-4o (prospective). Internal binary commit ($0.731$) and cross-model consensus ($0.751$) sit at the top, ahead of every external feature subset.}
\end{subfigure}
\caption{ROC curves on the prospective no-help condition for two illustrative (benchmark, model) cells. Each panel overlays multiple correctness predictors: the internal binary commit, the cross-model consensus, ridge logistic regressions on the Q, M, and QM feature subsets defined in \S\ref{app:si:external_classifier}, two transfer classifiers trained on a different benchmark, two univariate surface baselines (hedging count, response length), and the no-skill diagonal. Legend entries are sorted by AUC. The two panels illustrate the two qualitative regimes that recur in Table~\ref{tab:calibration_roc}: a regime in which the QM lexicon classifier substantially exceeds the internal binary commit (panel a), and a regime in which the internal commit matches or beats the best surface predictor (panel b). AUC values for all rows are in main paper Table~\ref{tab:calibration_roc}.}
\label{fig:si:roc_comparison}
\end{figure}

\subsection{Classifier transfer matrix}
\label{app:si:transfer}

\subsection{Performance vs.\ confidence across all benchmarks}
\label{app:si:calibration_all}

\begin{figure}[H]
\centering
\newcommand{\sicalpanel}[3]{%
  \begin{subfigure}[t]{0.235\textwidth}%
    \includegraphics[width=\linewidth]{#1}%
    \caption{#2}\label{fig:si:calibration_all:#3}%
  \end{subfigure}%
}
\sicalpanel{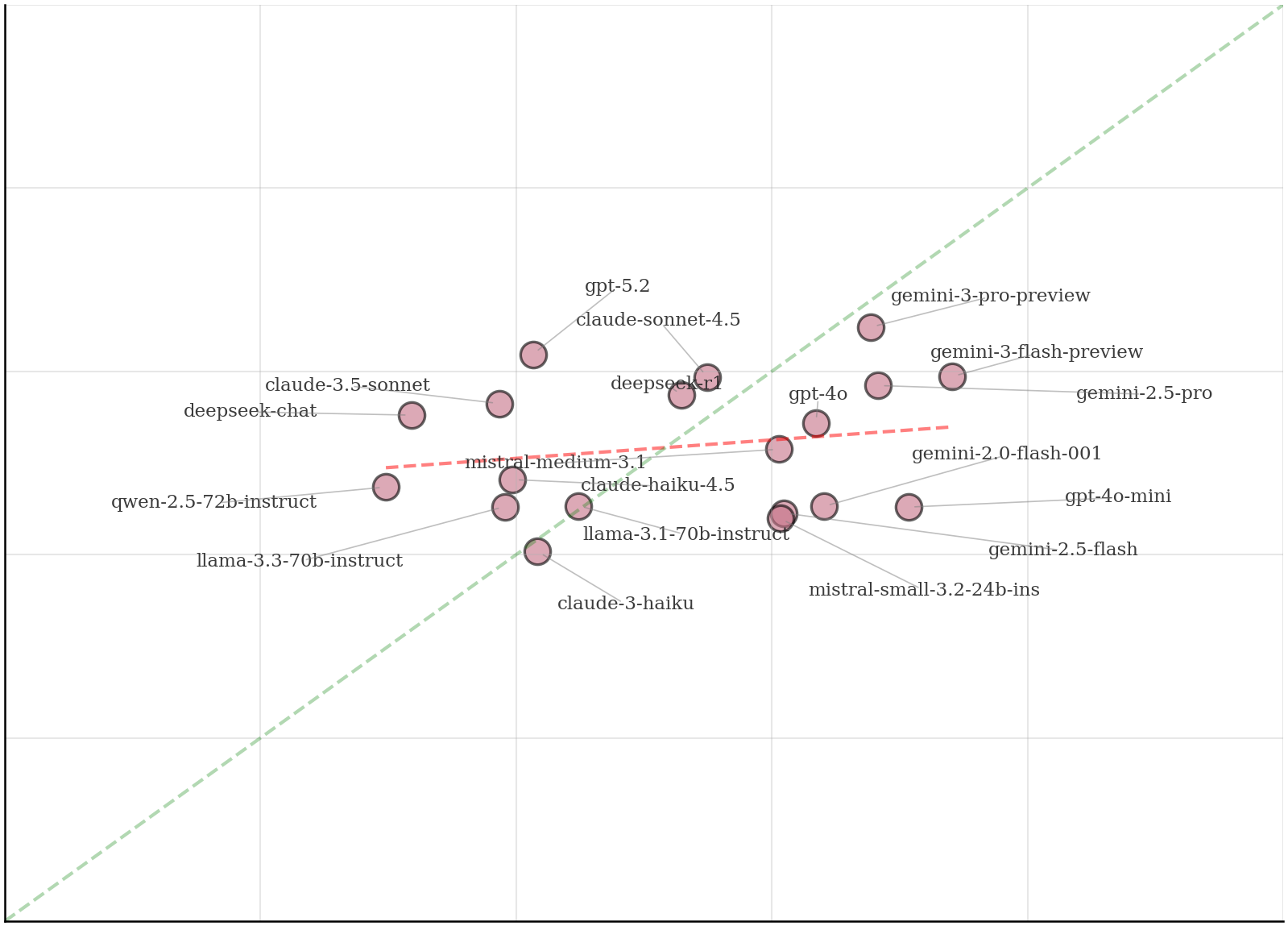}{SQuAD-before}{a}\hfill
\sicalpanel{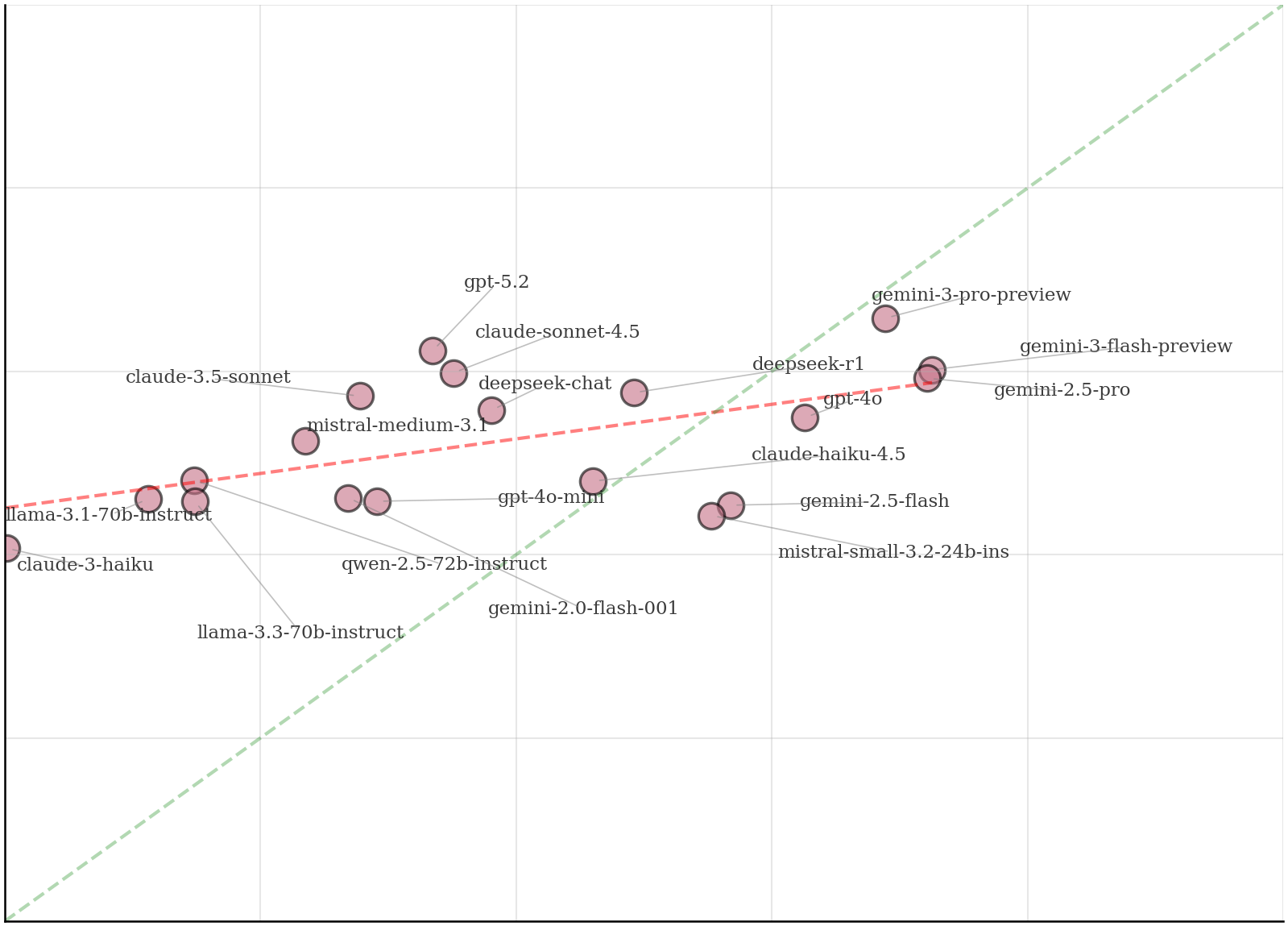}{SQuAD-after}{b}\hfill
\sicalpanel{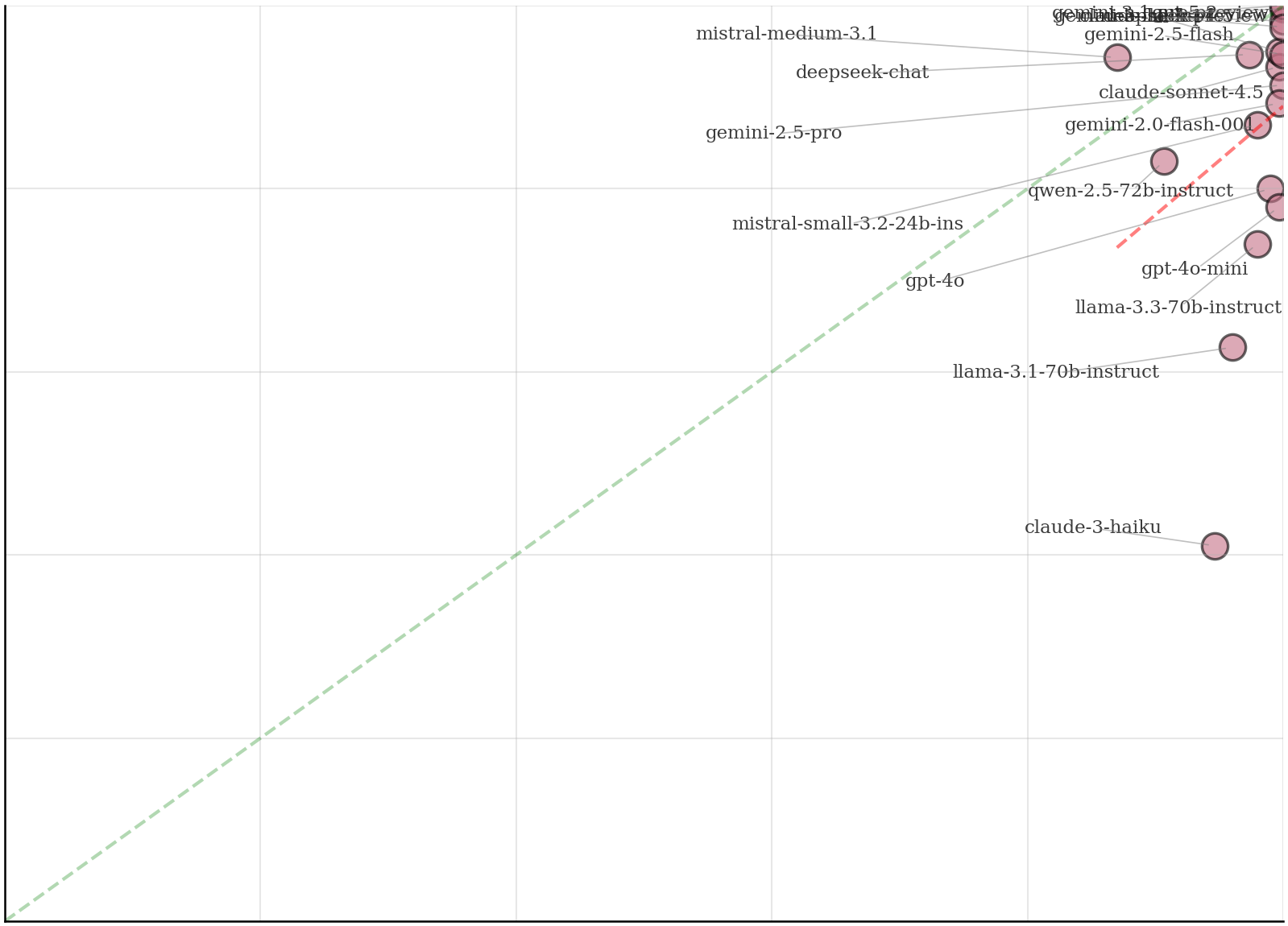}{MathBench-before}{c}\hfill
\sicalpanel{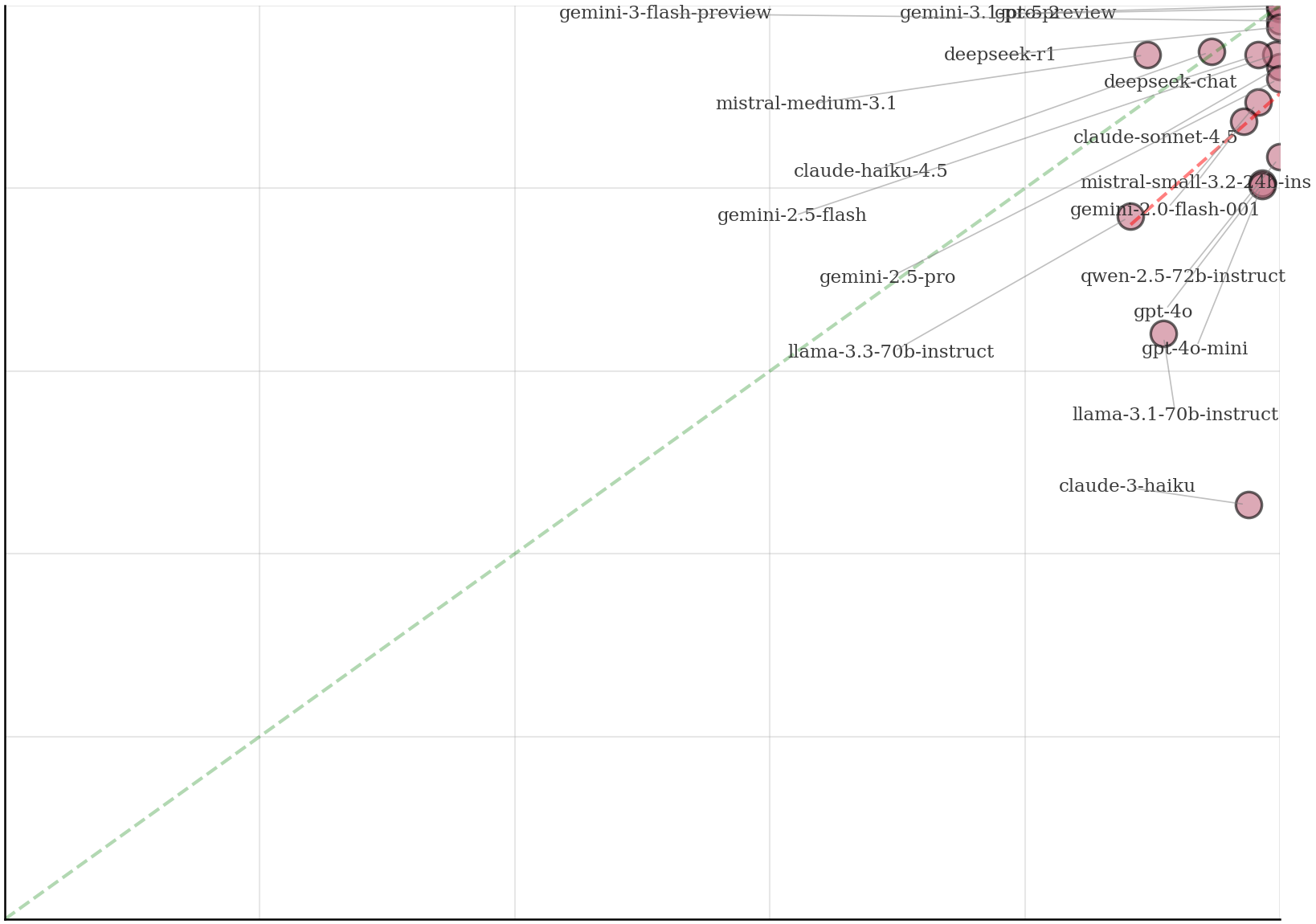}{MathBench-after}{d}

\vspace{0.5em}
\sicalpanel{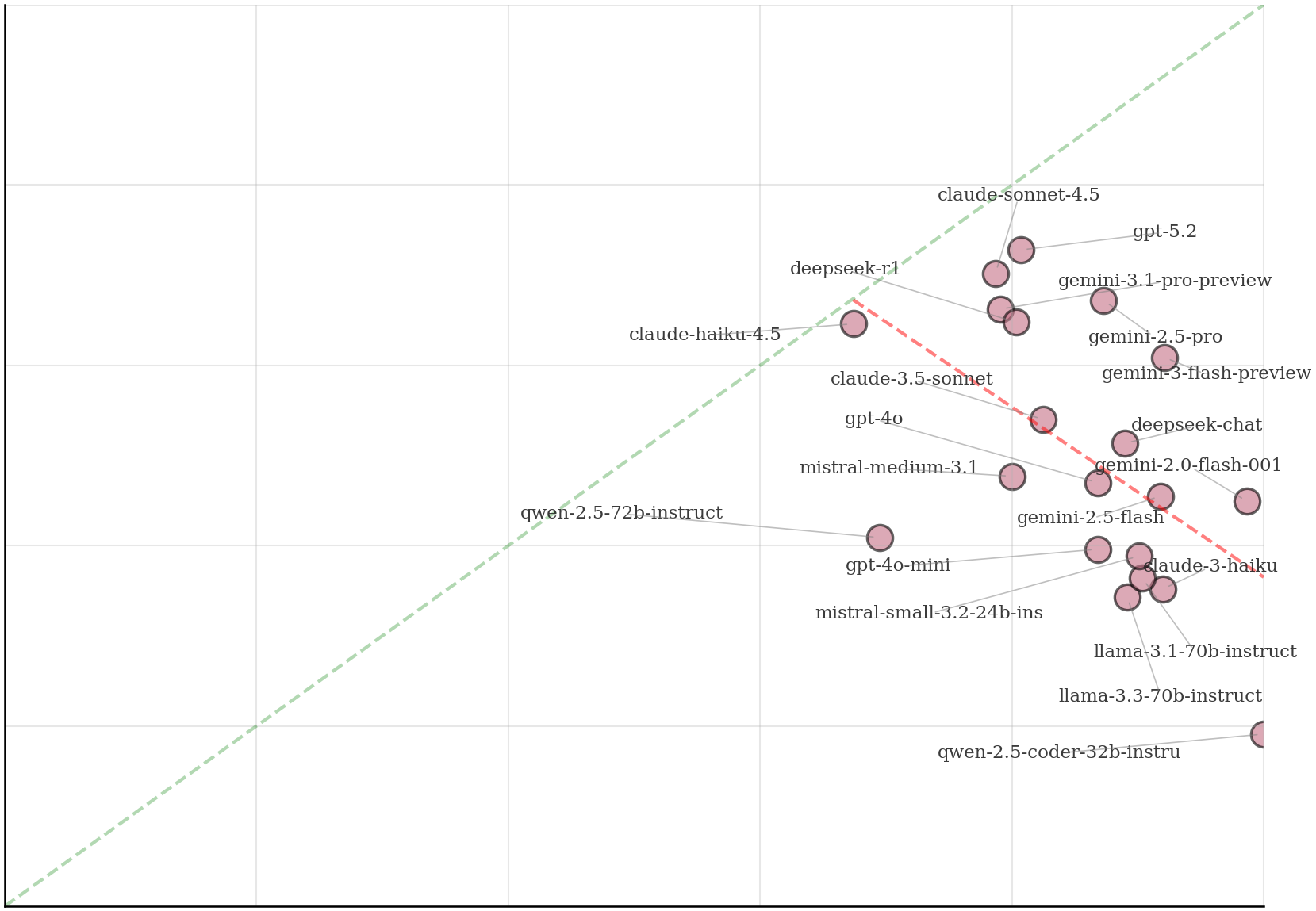}{MMLU-Pro-before}{e}\hfill
\sicalpanel{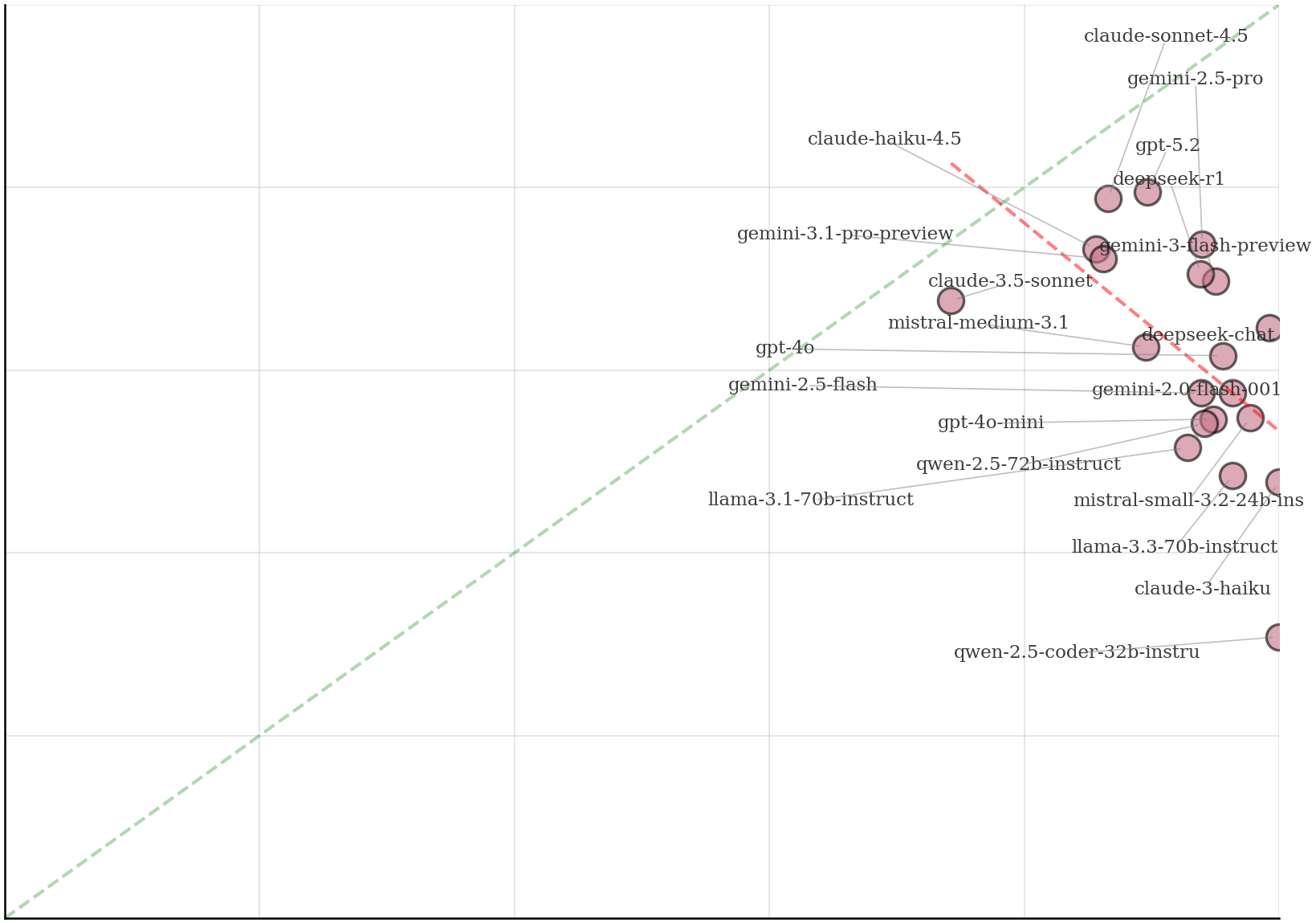}{MMLU-Pro-after}{f}\hfill
\sicalpanel{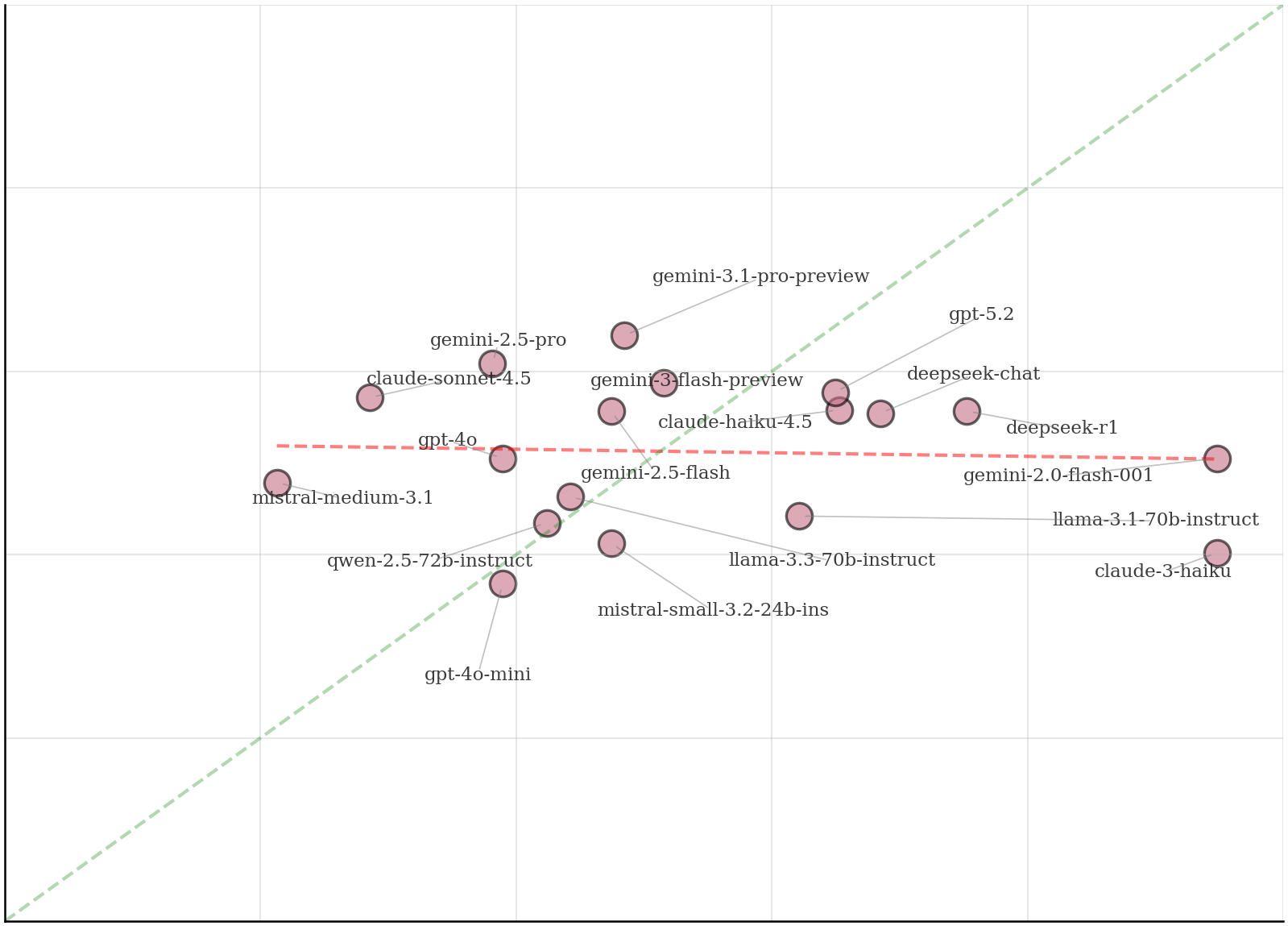}{SciCode-before}{g}\hfill
\sicalpanel{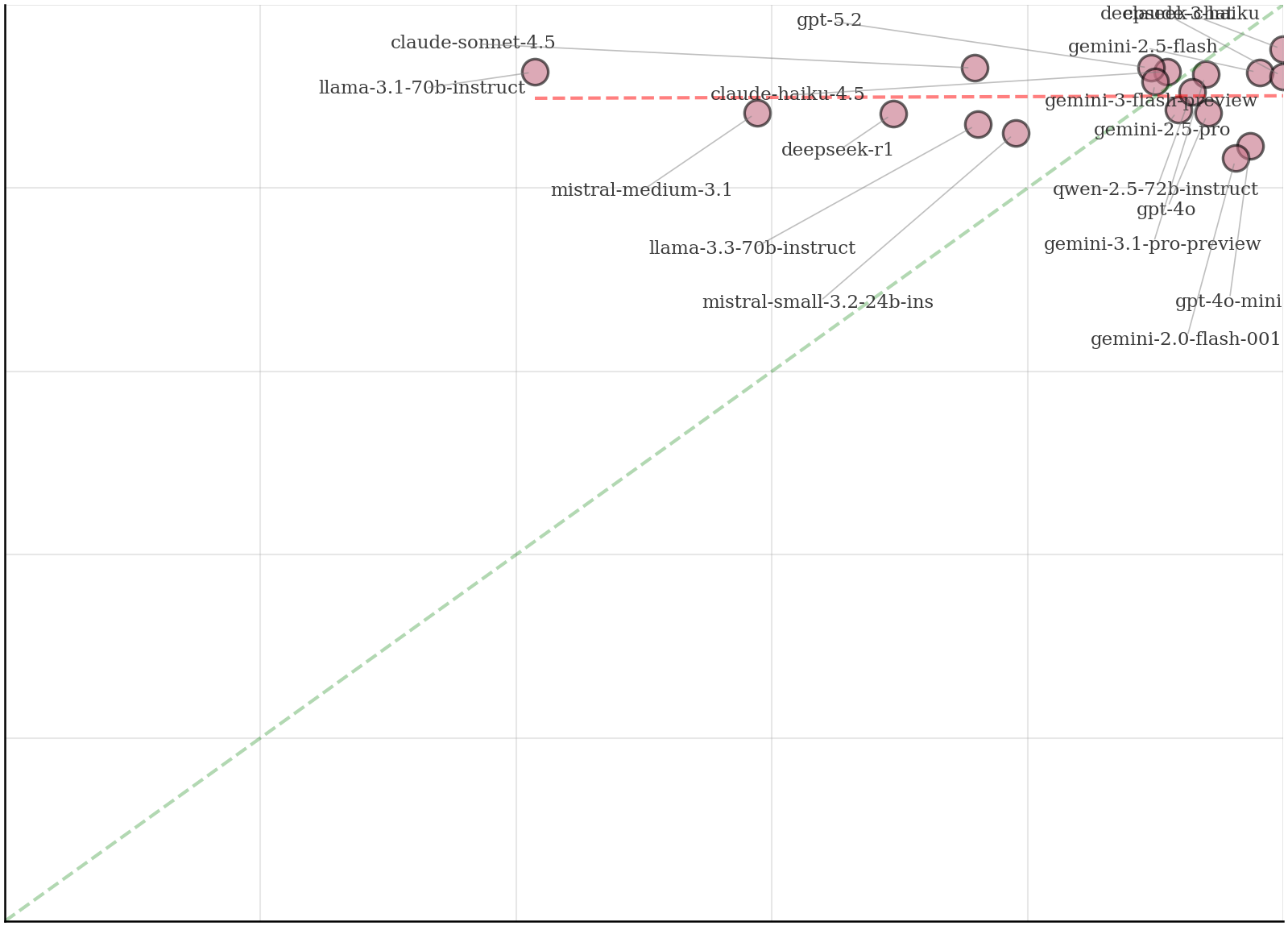}{SciCode-after}{h}

\vspace{0.5em}
\sicalpanel{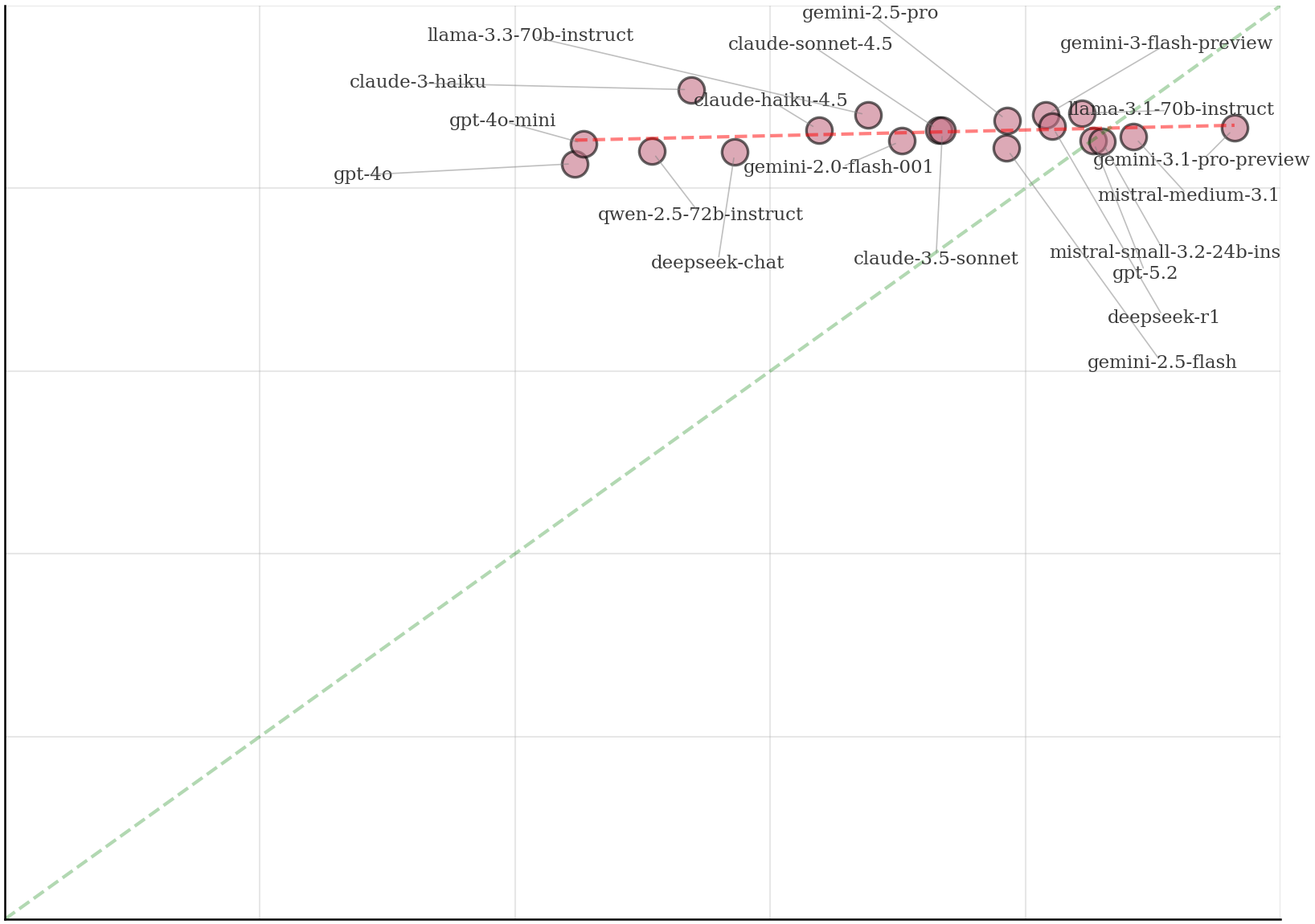}{LegalBench-before}{i}\hfill
\begin{subfigure}[t]{0.235\textwidth}\centering\vspace{2.5em}\small (no counterfactual probe)\end{subfigure}\hfill
\sicalpanel{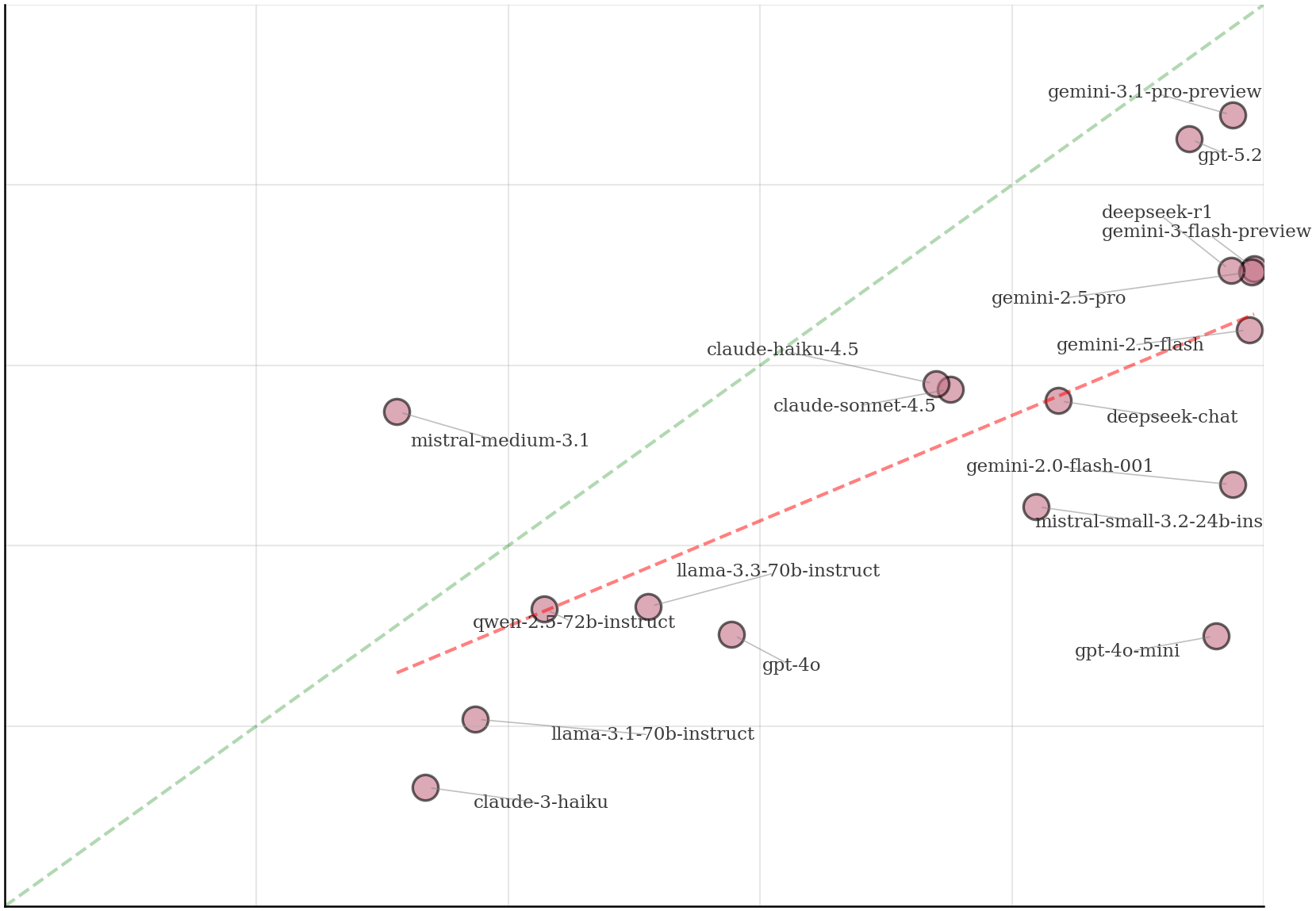}{OmniMath-before}{j}\hfill
\sicalpanel{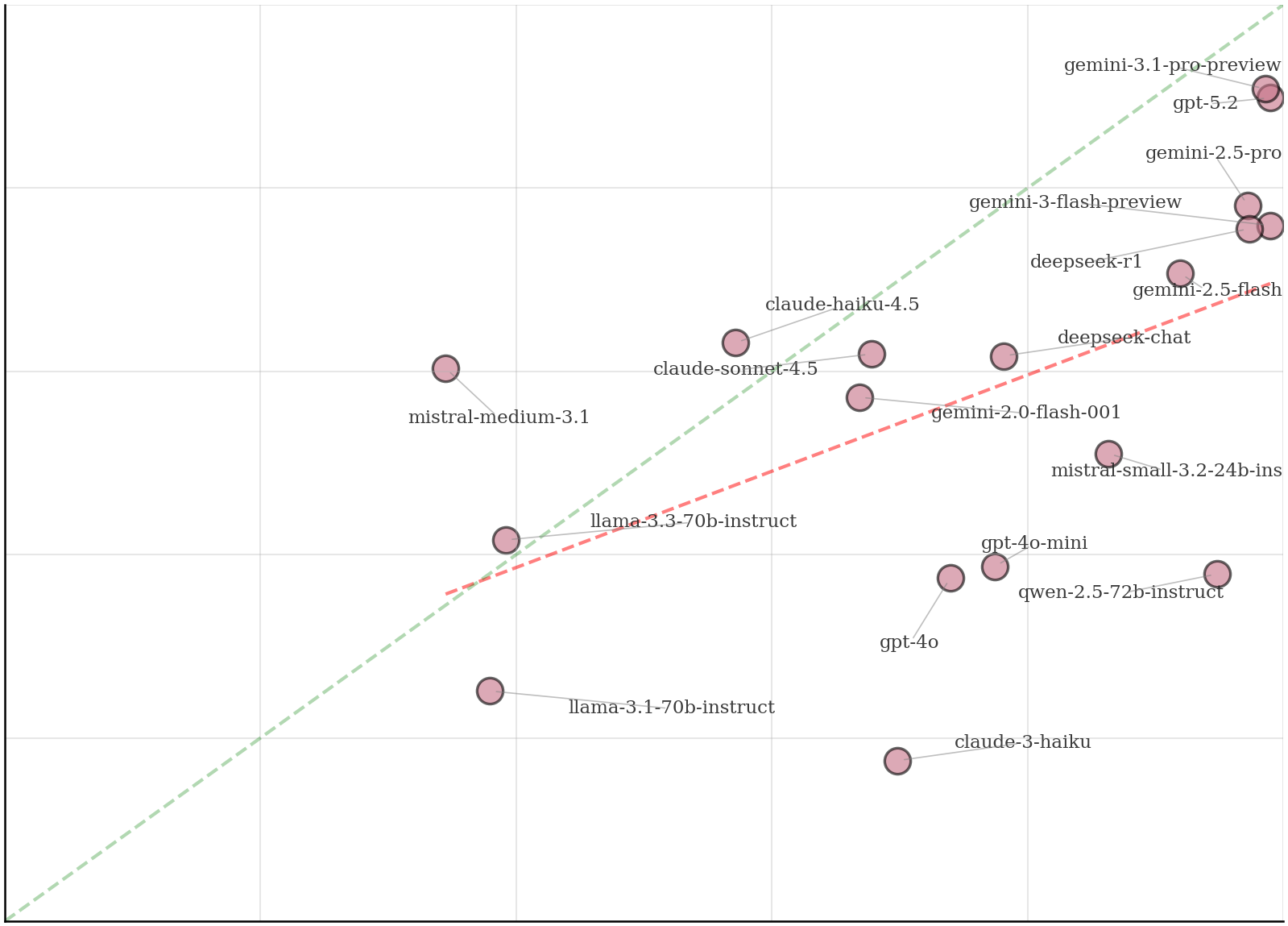}{OmniMath-after}{k}

\caption{Labelled variant of Fig.~\ref{fig:calibration_all}: model-level aggregate confidence against accuracy for all six benchmarks, with each scatter point annotated by the model identifier. In every panel the horizontal axis is the model's mean confidence (fraction of items rated yes) and the vertical axis is its accuracy (fraction of items answered correctly); both run from 0 to 1 and tick labels are suppressed for legibility at this grid size. The left block (columns 1--2) shows retrieval and factual benchmarks (SQuAD, MMLU-Pro, LegalBench), the right block (columns 3--4) shows reasoning benchmarks (MathBench, SciCode, OmniMath). Within each block, the first column is the prospective probe and the second column is the counterfactual probe. The LegalBench counterfactual probe is not in the protocol and is omitted.}
\label{fig:si:calibration_all}
\end{figure}

\subsection{\texorpdfstring{Tetrachoric eigenspectra across all benchmarks}{Tetrachoric eigenspectra across all benchmarks}}
\label{app:si:eigenspectra}

\subsubsection*{Tetrachoric eigenspectrum}
\noindent Eigenvalues of the tetrachoric correlation matrix of binary metacognitive judgements across the 20 models, normalised so the spectrum sums to one. Left panel: per-rank eigenvalue (markers) with the per-rank 95th percentile of a base-rate-preserving empirical null overlaid as a red dashed curve. The null is generated by independently column-shuffling each model's judgements and re-estimating the tetrachoric matrix, $B = 100$ draws. Green markers lie above the null (signal), grey markers below (noise). Right panel: cumulative explained variance for the observed spectrum (green) and the null mean (pink).\par\medskip

\begin{figure}[H]
  \centering
  \begin{subfigure}[b]{0.31\textwidth}
    \centering
    \includegraphics[width=\textwidth]{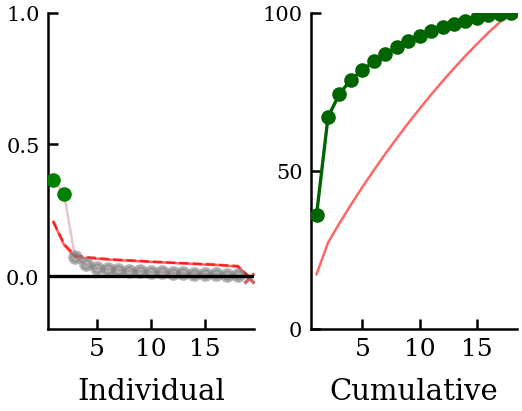}
    \caption{LegalBench (prospective)}
    \label{fig:si:latent_fa-eigenspectrum:prospective:legalbench:confidence_without_definition}
  \end{subfigure}
  \begin{subfigure}[b]{0.31\textwidth}
    \centering
    \includegraphics[width=\textwidth]{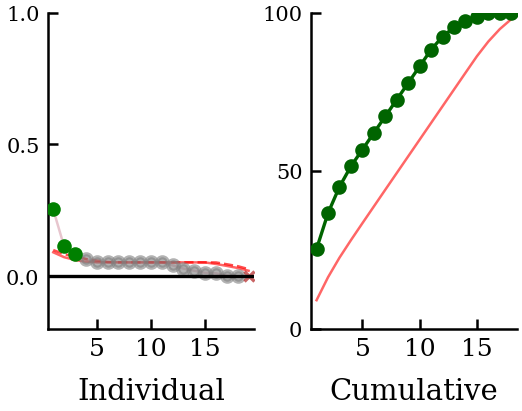}
    \caption{MathBench (prospective)}
    \label{fig:si:latent_fa-eigenspectrum:prospective:math_bench:confidence_before}
  \end{subfigure}
  \begin{subfigure}[b]{0.31\textwidth}
    \centering
    \includegraphics[width=\textwidth]{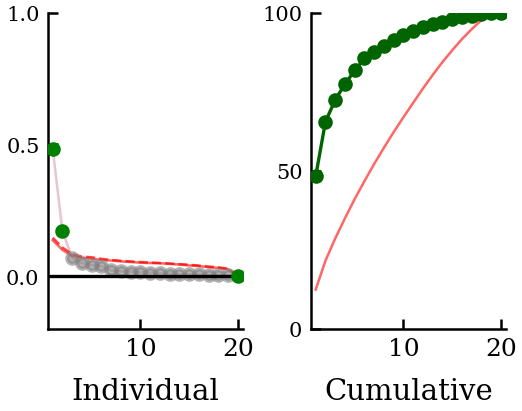}
    \caption{MMLU (prospective)}
    \label{fig:si:latent_fa-eigenspectrum:prospective:mmlu:confidence_without_choices}
  \end{subfigure}
  \\[1ex]
  \begin{subfigure}[b]{0.31\textwidth}
    \centering
    \includegraphics[width=\textwidth]{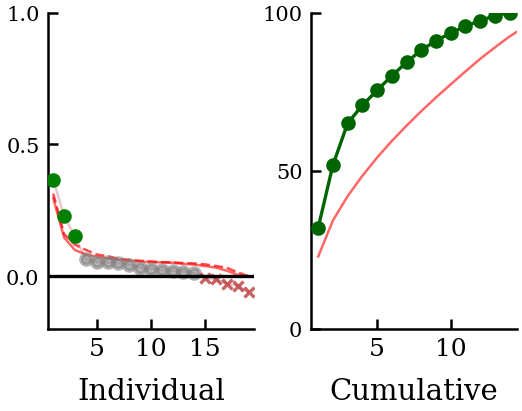}
    \caption{omnimath (prospective)}
    \label{fig:si:latent_fa-eigenspectrum:prospective:omnimath:confidence_before}
  \end{subfigure}
  \begin{subfigure}[b]{0.31\textwidth}
    \centering
    \includegraphics[width=\textwidth]{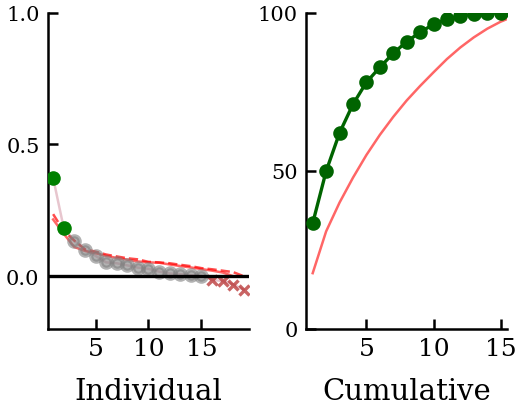}
    \caption{SciCode (prospective)}
    \label{fig:si:latent_fa-eigenspectrum:prospective:scicode:confidence_before}
  \end{subfigure}
  \begin{subfigure}[b]{0.31\textwidth}
    \centering
    \includegraphics[width=\textwidth]{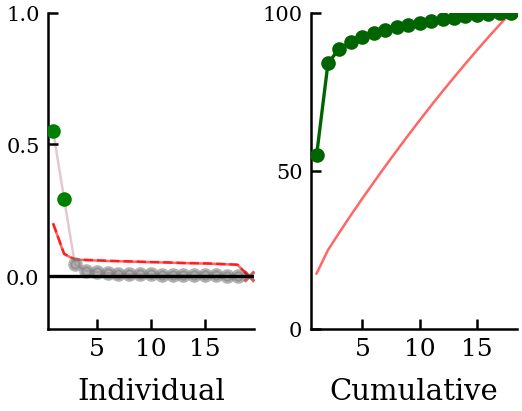}
    \caption{SQuAD (prospective)}
    \label{fig:si:latent_fa-eigenspectrum:prospective:squad:needs_context}
  \end{subfigure}
  \caption{\textbf{Tetrachoric eigenspectrum}, prospective (pre-task) judgments. Eigenvalues of the tetrachoric correlation matrix of binary metacognitive judgements across the 20 models, normalised so the spectrum sums to one. Left panel: per-rank eigenvalue (markers) with the per-rank 95th percentile of a base-rate-preserving empirical null overlaid as a red dashed curve. The null is generated by independently column-shuffling each model's judgements and re-estimating the tetrachoric matrix, $B = 100$ draws. Green markers lie above the null (signal), grey markers below (noise). Right panel: cumulative explained variance for the observed spectrum (green) and the null mean (pink).}
  \label{fig:si:latent_fa-eigenspectrum:prospective}
\end{figure}

\begin{figure}[H]
  \centering
  \begin{subfigure}[b]{0.31\textwidth}
    \centering
    \includegraphics[width=\textwidth]{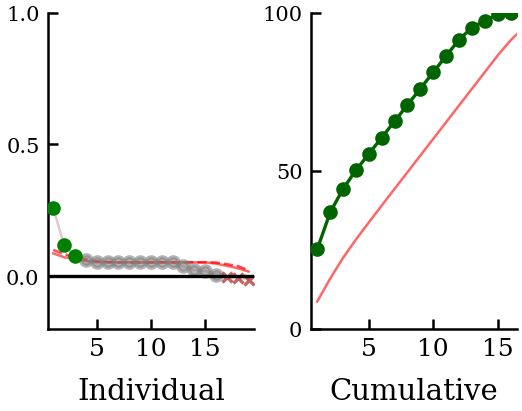}
    \caption{MathBench (counterfactual)}
    \label{fig:si:latent_fa-eigenspectrum:counterfactual:math_bench:confidence_after}
  \end{subfigure}
  \begin{subfigure}[b]{0.31\textwidth}
    \centering
    \includegraphics[width=\textwidth]{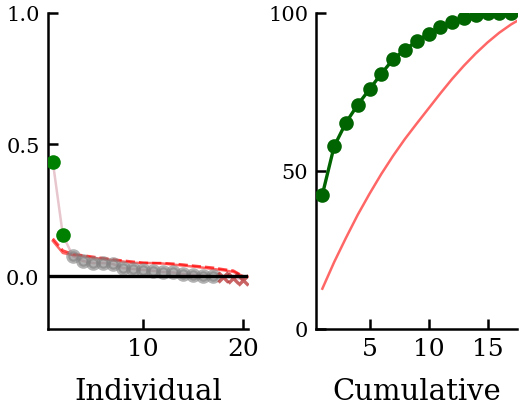}
    \caption{MMLU (counterfactual)}
    \label{fig:si:latent_fa-eigenspectrum:counterfactual:mmlu:choices_necessary}
  \end{subfigure}
  \begin{subfigure}[b]{0.31\textwidth}
    \centering
    \includegraphics[width=\textwidth]{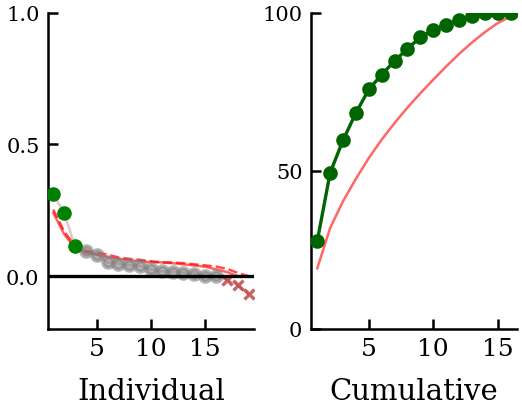}
    \caption{omnimath (counterfactual)}
    \label{fig:si:latent_fa-eigenspectrum:counterfactual:omnimath:confidence_after}
  \end{subfigure}
  \\[1ex]
  \begin{subfigure}[b]{0.31\textwidth}
    \centering
    \includegraphics[width=\textwidth]{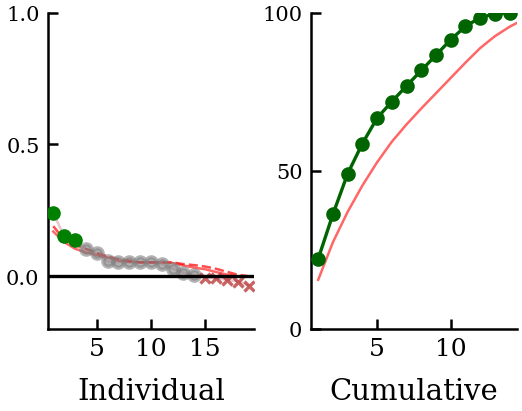}
    \caption{SciCode (counterfactual)}
    \label{fig:si:latent_fa-eigenspectrum:counterfactual:scicode:confidence_after}
  \end{subfigure}
  \begin{subfigure}[b]{0.31\textwidth}
    \centering
    \includegraphics[width=\textwidth]{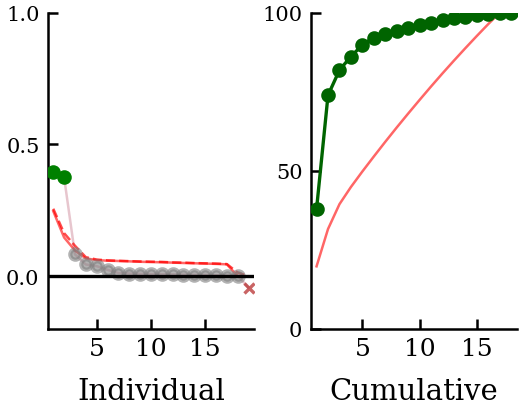}
    \caption{SQuAD (counterfactual)}
    \label{fig:si:latent_fa-eigenspectrum:counterfactual:squad:context_necessary}
  \end{subfigure}
  \caption{\textbf{Tetrachoric eigenspectrum}, counterfactual (post-task) judgments. Eigenvalues of the tetrachoric correlation matrix of binary metacognitive judgements across the 20 models, normalised so the spectrum sums to one. Left panel: per-rank eigenvalue (markers) with the per-rank 95th percentile of a base-rate-preserving empirical null overlaid as a red dashed curve. The null is generated by independently column-shuffling each model's judgements and re-estimating the tetrachoric matrix, $B = 100$ draws. Green markers lie above the null (signal), grey markers below (noise). Right panel: cumulative explained variance for the observed spectrum (green) and the null mean (pink).}
  \label{fig:si:latent_fa-eigenspectrum:counterfactual}
\end{figure}

\subsection{\texorpdfstring{Factor overview across all benchmarks}{Factor overview across all benchmarks}}
\label{app:si:factor_overview}

Each model has a probit response threshold $\tau_i = \Phi^{-1}(\text{yes-rate}_i)$. A 50\% yes-rate model has $\tau_i = 0$, with $|\tau_i| \to \infty$ at floor or ceiling. Eigendecomposing the tetrachoric correlation matrix gives each model a PC1 loading. The plots below show PC1 loading against $\tau_i$, one point per model, with both linear and quadratic fits.

Tetrachoric correlation estimates the latent Pearson correlation between two continuous variables thresholded into binary outcomes. Information per $2 \times 2$ contingency table is maximal at $\tau_i = 0$ and shrinks symmetrically as $|\tau_i|$ grows, because a model whose yes-rate is near 0 or 1 contributes mostly to a single cell of every contingency table it joins. Its row of the tetrachoric matrix is therefore noisier and its PC1 loading is mechanically attenuated. Under a single shared latent factor the PC1 loadings trace a downward-opening parabola in $\tau_i$, peaked at zero. The shape is a property of the estimator on binary data, not a substantive claim about the models. A high quadratic $R^2$ explains apparent dispersion in the loadings as a function of threshold position alone. A low $R^2$ is consistent with either multi-factor structure or with the non-PSD signature flagged in Fig.~\ref{fig:eigenspectra}.

\begin{figure}[H]
  \centering
  \begin{subfigure}[b]{0.31\textwidth}
    \centering
    \includegraphics[width=\textwidth]{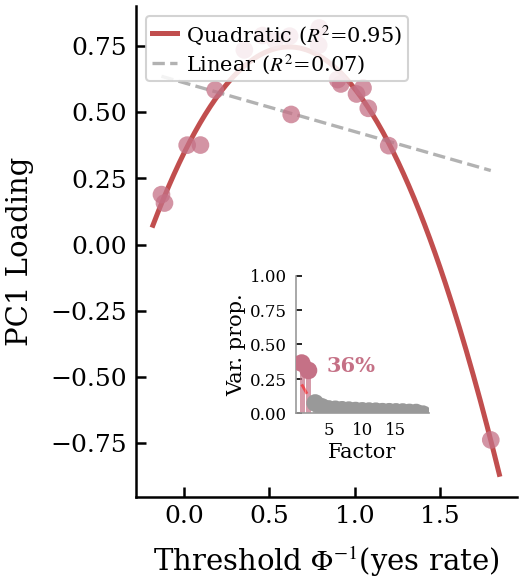}
    \caption{LegalBench (prospective)}
    \label{fig:si:latent_fa-factor_overview:prospective:legalbench:confidence_without_definition}
  \end{subfigure}
  \begin{subfigure}[b]{0.31\textwidth}
    \centering
    \includegraphics[width=\textwidth]{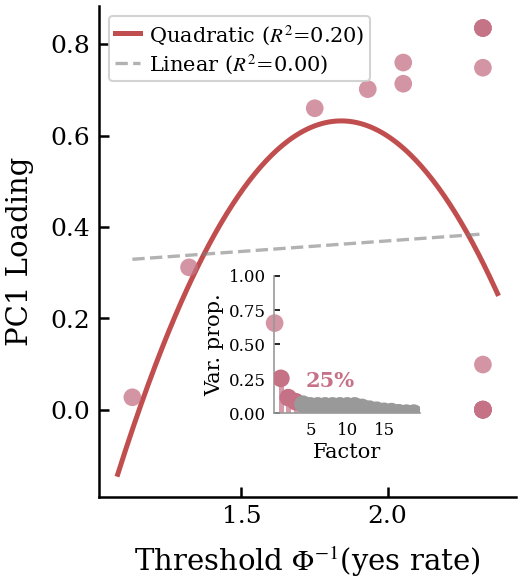}
    \caption{MathBench (prospective)}
    \label{fig:si:latent_fa-factor_overview:prospective:math_bench:confidence_before}
  \end{subfigure}
  \begin{subfigure}[b]{0.31\textwidth}
    \centering
    \includegraphics[width=\textwidth]{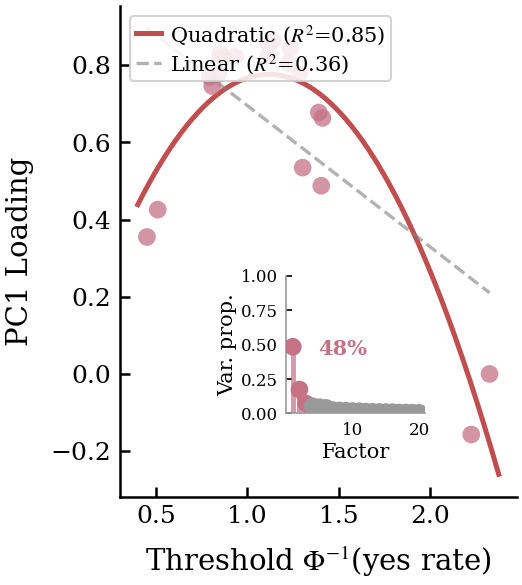}
    \caption{MMLU (prospective)}
    \label{fig:si:latent_fa-factor_overview:prospective:mmlu:confidence_without_choices}
  \end{subfigure}
  \\[1ex]
  \begin{subfigure}[b]{0.31\textwidth}
    \centering
    \includegraphics[width=\textwidth]{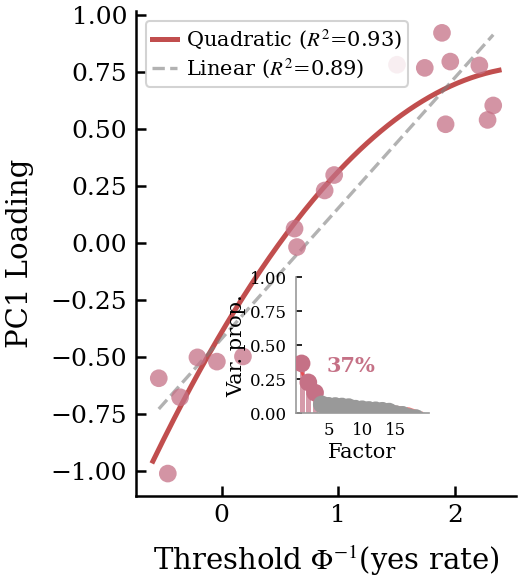}
    \caption{omnimath (prospective)}
    \label{fig:si:latent_fa-factor_overview:prospective:omnimath:confidence_before}
  \end{subfigure}
  \begin{subfigure}[b]{0.31\textwidth}
    \centering
    \includegraphics[width=\textwidth]{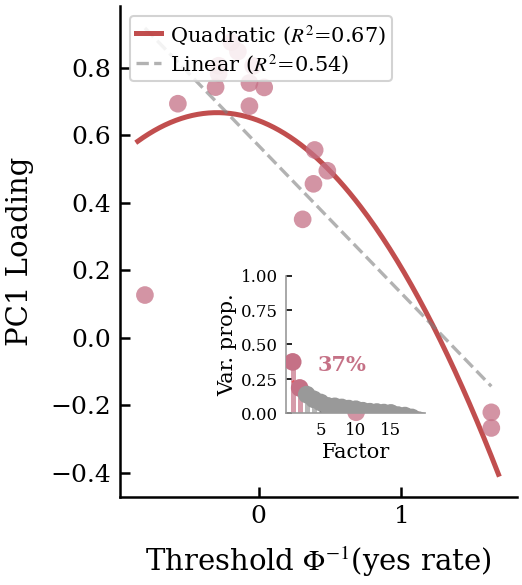}
    \caption{SciCode (prospective)}
    \label{fig:si:latent_fa-factor_overview:prospective:scicode:confidence_before}
  \end{subfigure}
  \begin{subfigure}[b]{0.31\textwidth}
    \centering
    \includegraphics[width=\textwidth]{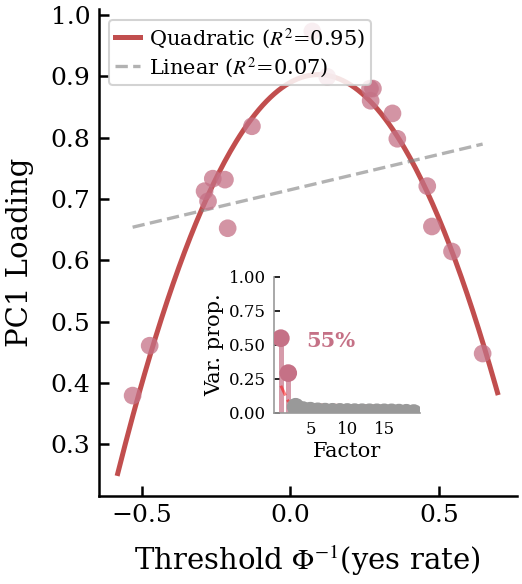}
    \caption{SQuAD (prospective)}
    \label{fig:si:latent_fa-factor_overview:prospective:squad:needs_context}
  \end{subfigure}
  \caption{\textbf{Factor overview}, prospective (pre-task) judgments. Per-model PC1 loading on the metacognitive tetrachoric correlation matrix, against the model's probit response threshold $\tau_i = \Phi^{-1}(\text{yes-rate}_i)$. The red curve is a quadratic fit and the grey dashed curve is a linear fit. A high quadratic $R^2$ is the single-factor signature, because under one shared latent the loadings trace a downward-opening parabola peaked at $\tau_i = 0$ and attenuating mechanically toward floor and ceiling thresholds. The inset reproduces the normalised eigenspectrum from Fig.~\ref{fig:eigenspectra} for the same condition.}
  \label{fig:si:latent_fa-factor_overview:prospective}
\end{figure}

\begin{figure}[H]
  \centering
  \begin{subfigure}[b]{0.31\textwidth}
    \centering
    \includegraphics[width=\textwidth]{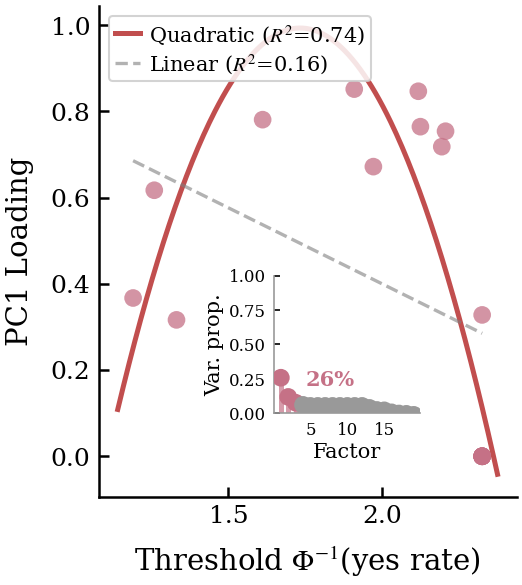}
    \caption{MathBench (counterfactual)}
    \label{fig:si:latent_fa-factor_overview:counterfactual:math_bench:confidence_after}
  \end{subfigure}
  \begin{subfigure}[b]{0.31\textwidth}
    \centering
    \includegraphics[width=\textwidth]{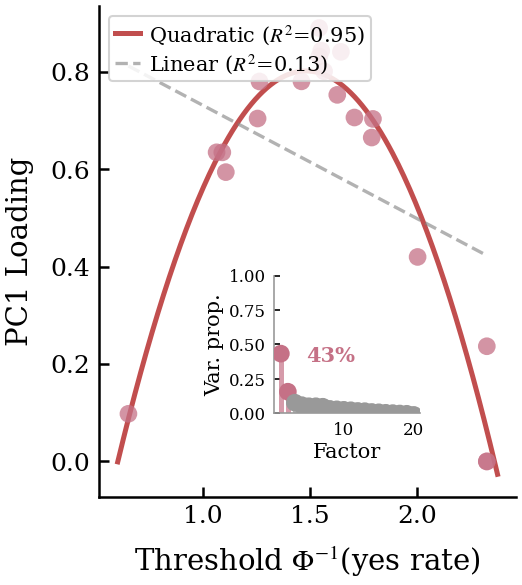}
    \caption{MMLU (counterfactual)}
    \label{fig:si:latent_fa-factor_overview:counterfactual:mmlu:choices_necessary}
  \end{subfigure}
  \begin{subfigure}[b]{0.31\textwidth}
    \centering
    \includegraphics[width=\textwidth]{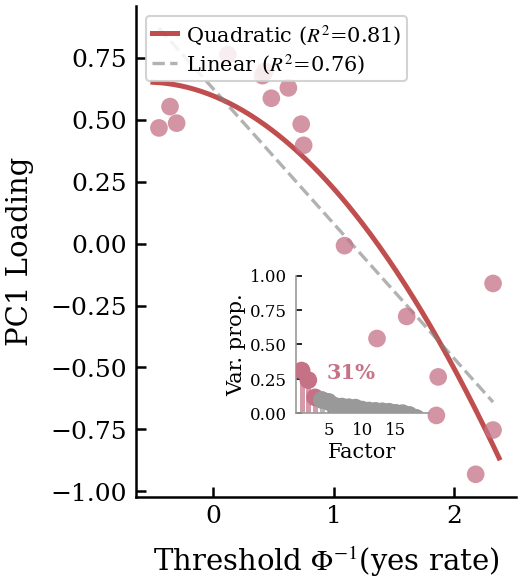}
    \caption{omnimath (counterfactual)}
    \label{fig:si:latent_fa-factor_overview:counterfactual:omnimath:confidence_after}
  \end{subfigure}
  \\[1ex]
  \begin{subfigure}[b]{0.31\textwidth}
    \centering
    \includegraphics[width=\textwidth]{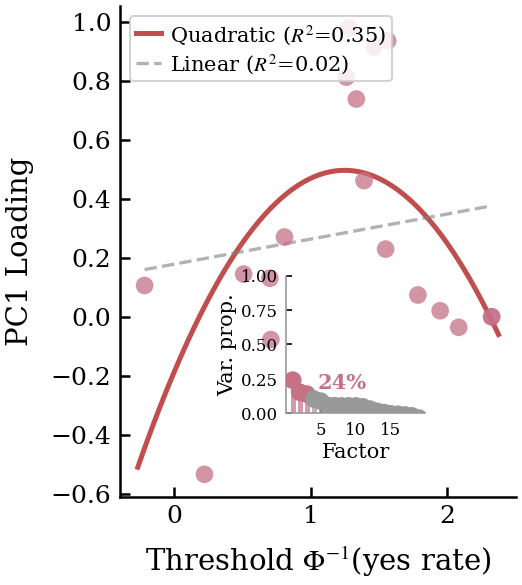}
    \caption{SciCode (counterfactual)}
    \label{fig:si:latent_fa-factor_overview:counterfactual:scicode:confidence_after}
  \end{subfigure}
  \begin{subfigure}[b]{0.31\textwidth}
    \centering
    \includegraphics[width=\textwidth]{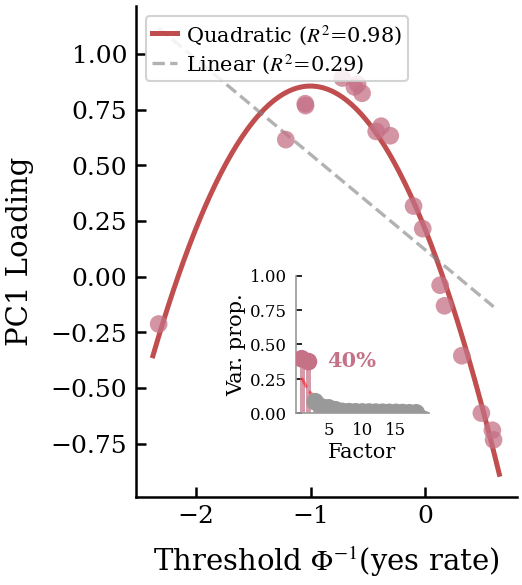}
    \caption{SQuAD (counterfactual)}
    \label{fig:si:latent_fa-factor_overview:counterfactual:squad:context_necessary}
  \end{subfigure}
  \caption{\textbf{Factor overview}, counterfactual (post-task) judgments. Per-model PC1 loading on the metacognitive tetrachoric correlation matrix, against the model's probit response threshold $\tau_i = \Phi^{-1}(\text{yes-rate}_i)$. The red curve is a quadratic fit and the grey dashed curve is a linear fit. A high quadratic $R^2$ is the single-factor signature, because under one shared latent the loadings trace a downward-opening parabola peaked at $\tau_i = 0$ and attenuating mechanically toward floor and ceiling thresholds. The inset reproduces the normalised eigenspectrum from Fig.~\ref{fig:eigenspectra} for the same condition.}
  \label{fig:si:latent_fa-factor_overview:counterfactual}
\end{figure}

Per-benchmark plots are in Fig.~\ref{fig:si:latent_fa-factor_overview:prospective} (prospective) and Fig.~\ref{fig:si:latent_fa-factor_overview:counterfactual} (counterfactual). SQuAD, MMLU-Pro, and LegalBench show $R^2_{\text{quad}} \in [0.63, 0.95]$ on prospective probes. MathBench, SciCode, and Omni-MATH sit lower, but their tetrachoric matrices are also non-Gramian, so the low fit reflects estimator breakdown rather than evidence for multi-factor structure.

\subsubsection*{Factor overview}
\noindent Per-model PC1 loading on the metacognitive tetrachoric correlation matrix, against the model's probit response threshold $\tau_i = \Phi^{-1}(\text{yes-rate}_i)$. The red curve is a quadratic fit and the grey dashed curve is a linear fit. A high quadratic $R^2$ is the single-factor signature, because under one shared latent the loadings trace a downward-opening parabola peaked at $\tau_i = 0$ and attenuating mechanically toward floor and ceiling thresholds. The inset reproduces the normalised eigenspectrum from Fig.~\ref{fig:eigenspectra} for the same condition.\par\medskip

\subsection{\texorpdfstring{Performance tetrachoric eigenspectra across all benchmarks}{Performance tetrachoric eigenspectra across all benchmarks}}
\label{app:si:eigenanalysis_perf}

Each panel shows the tetrachoric correlation matrix of per-item correctness across the 20 models, eigendecomposed with the same base-rate-preserving empirical null used on the confidence side. The red dashed curve is the per-rank 95th percentile of an independent column-shuffle bootstrap with $B = 100$ draws.

SQuAD, MMLU-Pro, and LegalBench all have a dominant first performance factor. SQuAD PC1 captures ${\sim}68\%$ of the variance and the other two retrieval benchmarks are similar. MathBench, Omni-MATH, and SciCode show the same long-tail and non-Gramian pattern as on the confidence side. Both signals are therefore approximately rank-one on the retrieval benchmarks, which is what makes the F1-vs-F1 alignment test in Fig.~\ref{fig:filtered}c,d the load-bearing question. A rank-one confidence signal is naively the right thing to emit for a rank-one performance signal, but the two rank-one axes coincide only on items everyone agrees on.

\subsubsection*{Performance tetrachoric eigenspectrum}
\noindent Eigenvalues of the tetrachoric correlation matrix of binary correctness judgements (right/wrong) across the 20 models, normalised so the spectrum sums to one. Left panel: per-rank eigenvalue with the red dashed per-rank 95th percentile of a base-rate-preserving empirical null. Green markers lie above the null, grey below. Right panel: cumulative explained variance for the observed spectrum (green) and the null mean (pink).\par\medskip

\begin{figure}[H]
  \centering
  \begin{subfigure}[b]{0.31\textwidth}
    \centering
    \includegraphics[width=\textwidth]{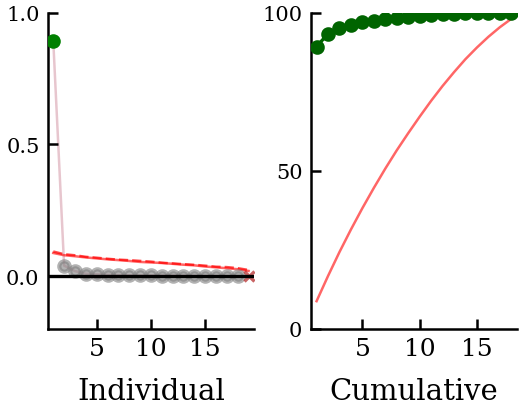}
    \caption{LegalBench (prospective)}
    \label{fig:si:eigenanalysis_perf-eigenspectrum:prospective:legalbench:confidence_without_definition}
  \end{subfigure}
  \begin{subfigure}[b]{0.31\textwidth}
    \centering
    \includegraphics[width=\textwidth]{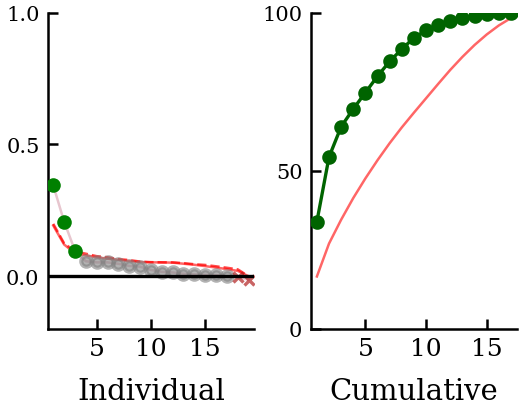}
    \caption{MathBench (prospective)}
    \label{fig:si:eigenanalysis_perf-eigenspectrum:prospective:math_bench:confidence_before}
  \end{subfigure}
  \begin{subfigure}[b]{0.31\textwidth}
    \centering
    \includegraphics[width=\textwidth]{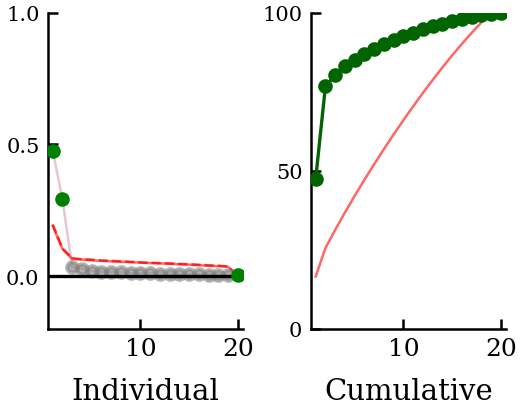}
    \caption{MMLU (prospective)}
    \label{fig:si:eigenanalysis_perf-eigenspectrum:prospective:mmlu:confidence_without_choices}
  \end{subfigure}
  \\[1ex]
  \begin{subfigure}[b]{0.31\textwidth}
    \centering
    \includegraphics[width=\textwidth]{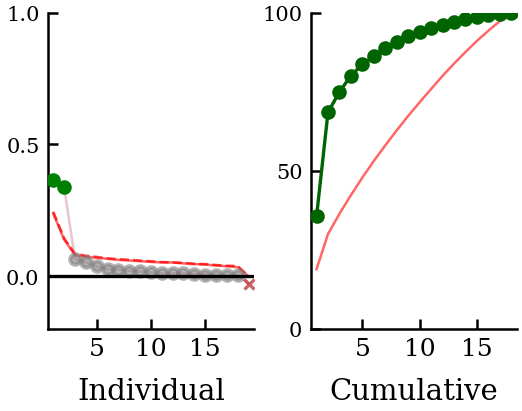}
    \caption{omnimath (prospective)}
    \label{fig:si:eigenanalysis_perf-eigenspectrum:prospective:omnimath:confidence_before}
  \end{subfigure}
  \begin{subfigure}[b]{0.31\textwidth}
    \centering
    \includegraphics[width=\textwidth]{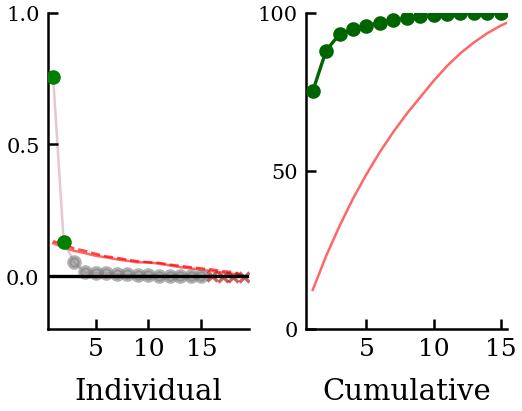}
    \caption{SciCode (prospective)}
    \label{fig:si:eigenanalysis_perf-eigenspectrum:prospective:scicode:confidence_before}
  \end{subfigure}
  \begin{subfigure}[b]{0.31\textwidth}
    \centering
    \includegraphics[width=\textwidth]{figures/squad/needs_context/eigenanalysis_perf/eigenspectrum_clean.png}
    \caption{SQuAD (prospective)}
    \label{fig:si:eigenanalysis_perf-eigenspectrum:prospective:squad:needs_context}
  \end{subfigure}
  \caption{\textbf{Performance tetrachoric eigenspectrum}, prospective (pre-task) judgments. Eigenvalues of the tetrachoric correlation matrix of binary correctness judgements (right/wrong) across the 20 models, normalised so the spectrum sums to one. Left panel: per-rank eigenvalue with the red dashed per-rank 95th percentile of a base-rate-preserving empirical null. Green markers lie above the null, grey below. Right panel: cumulative explained variance for the observed spectrum (green) and the null mean (pink).}
  \label{fig:si:eigenanalysis_perf-eigenspectrum:prospective}
\end{figure}

\begin{figure}[H]
  \centering
  \begin{subfigure}[b]{0.31\textwidth}
    \centering
    \includegraphics[width=\textwidth]{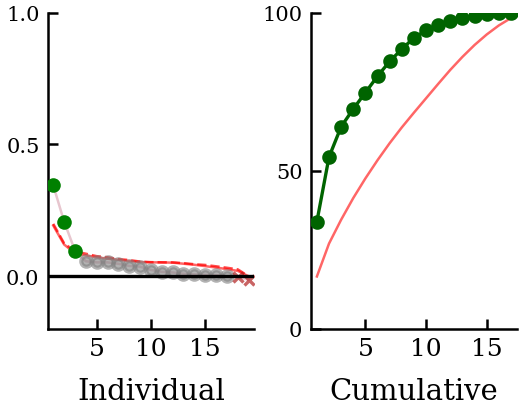}
    \caption{MathBench (counterfactual)}
    \label{fig:si:eigenanalysis_perf-eigenspectrum:counterfactual:math_bench:confidence_after}
  \end{subfigure}
  \begin{subfigure}[b]{0.31\textwidth}
    \centering
    \includegraphics[width=\textwidth]{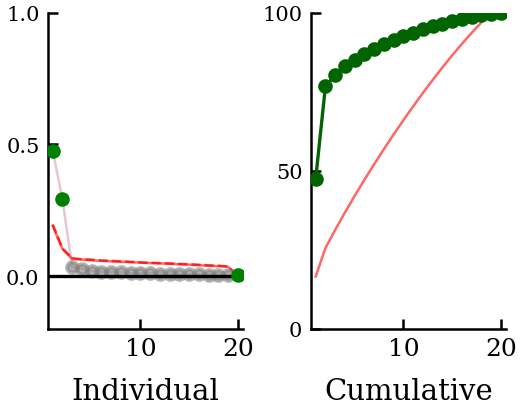}
    \caption{MMLU (counterfactual)}
    \label{fig:si:eigenanalysis_perf-eigenspectrum:counterfactual:mmlu:choices_necessary}
  \end{subfigure}
  \begin{subfigure}[b]{0.31\textwidth}
    \centering
    \includegraphics[width=\textwidth]{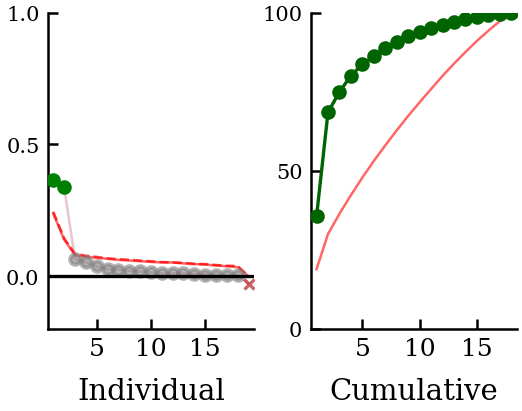}
    \caption{omnimath (counterfactual)}
    \label{fig:si:eigenanalysis_perf-eigenspectrum:counterfactual:omnimath:confidence_after}
  \end{subfigure}
  \\[1ex]
  \begin{subfigure}[b]{0.31\textwidth}
    \centering
    \includegraphics[width=\textwidth]{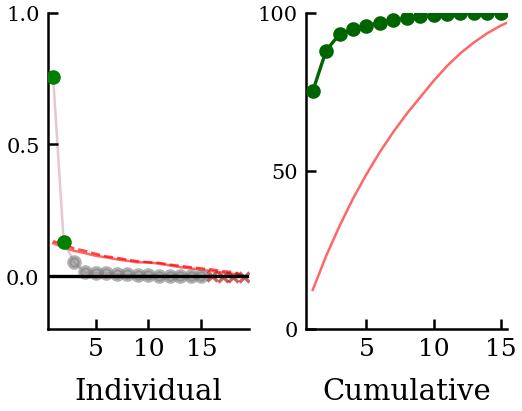}
    \caption{SciCode (counterfactual)}
    \label{fig:si:eigenanalysis_perf-eigenspectrum:counterfactual:scicode:confidence_after}
  \end{subfigure}
  \begin{subfigure}[b]{0.31\textwidth}
    \centering
    \includegraphics[width=\textwidth]{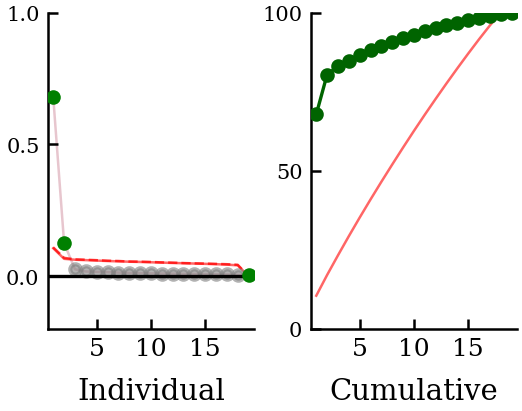}
    \caption{SQuAD (counterfactual)}
    \label{fig:si:eigenanalysis_perf-eigenspectrum:counterfactual:squad:context_necessary}
  \end{subfigure}
  \caption{\textbf{Performance tetrachoric eigenspectrum}, counterfactual (post-task) judgments. Eigenvalues of the tetrachoric correlation matrix of binary correctness judgements (right/wrong) across the 20 models, normalised so the spectrum sums to one. Left panel: per-rank eigenvalue with the red dashed per-rank 95th percentile of a base-rate-preserving empirical null. Green markers lie above the null, grey below. Right panel: cumulative explained variance for the observed spectrum (green) and the null mean (pink).}
  \label{fig:si:eigenanalysis_perf-eigenspectrum:counterfactual}
\end{figure}

\subsection{\texorpdfstring{Pairwise Kendall $\tau$ distributions across all benchmarks}{Pairwise Kendall distributions across all benchmarks}}
\label{app:si:pairwise_calibration}

\noindent Pairwise $\tau$-b between $\Delta$performance and $\Delta$metacognition over $\binom{n_\text{models}}{2}$ pairs. Left: orange is observed; blue is a base-rate-matched null (perf$_A$, perf$_B$, conf$_A$, conf$_B$ each shuffled independently); green is a calibration-preserving null (each model's perf and conf permuted together, calibration profile fixed). Solid verticals mark the three means; dashed grey marks $\tau=0$. Right: per-pair permutation $p$-values against the dashed grey uniform-null reference. Effect size is small but statistically significant with more highly significant pairs than expected if sampling from the null distribution. The observed mean sits significantly below the calibration-preserving null, implying that almost all calibration is shared rather than individuated.\par\medskip

\begin{figure}[H]
  \centering
  \begin{subfigure}[b]{0.31\textwidth}
    \centering
    \includegraphics[width=\textwidth]{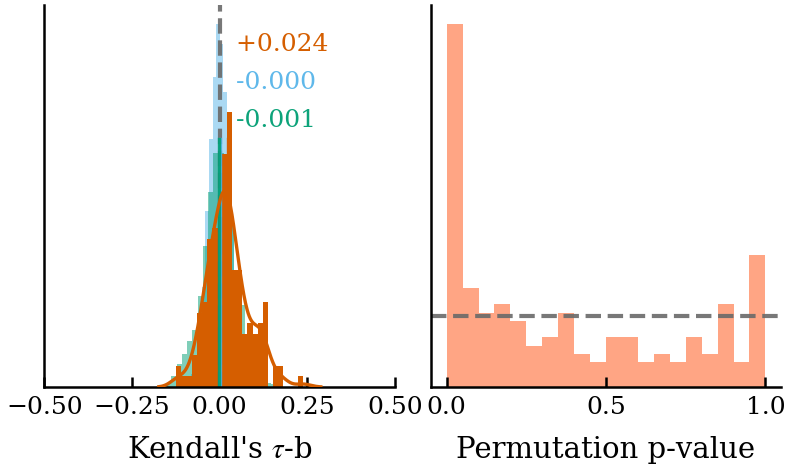}
    \caption{LegalBench (prospective)}
    \label{fig:si:tau_histogram:prospective:legalbench:confidence_without_definition}
  \end{subfigure}
  \begin{subfigure}[b]{0.31\textwidth}
    \centering
    \includegraphics[width=\textwidth]{figures/math_bench/confidence_before/tau_histogram_clean.png}
    \caption{MathBench (prospective)}
    \label{fig:si:tau_histogram:prospective:math_bench:confidence_before}
  \end{subfigure}
  \begin{subfigure}[b]{0.31\textwidth}
    \centering
    \includegraphics[width=\textwidth]{figures/mmlu/confidence_without_choices/tau_histogram_clean.png}
    \caption{MMLU (prospective)}
    \label{fig:si:tau_histogram:prospective:mmlu:confidence_without_choices}
  \end{subfigure}
  \\[1ex]
  \begin{subfigure}[b]{0.31\textwidth}
    \centering
    \includegraphics[width=\textwidth]{figures/omnimath/confidence_before/tau_histogram_clean.png}
    \caption{omnimath (prospective)}
    \label{fig:si:tau_histogram:prospective:omnimath:confidence_before}
  \end{subfigure}
  \begin{subfigure}[b]{0.31\textwidth}
    \centering
    \includegraphics[width=\textwidth]{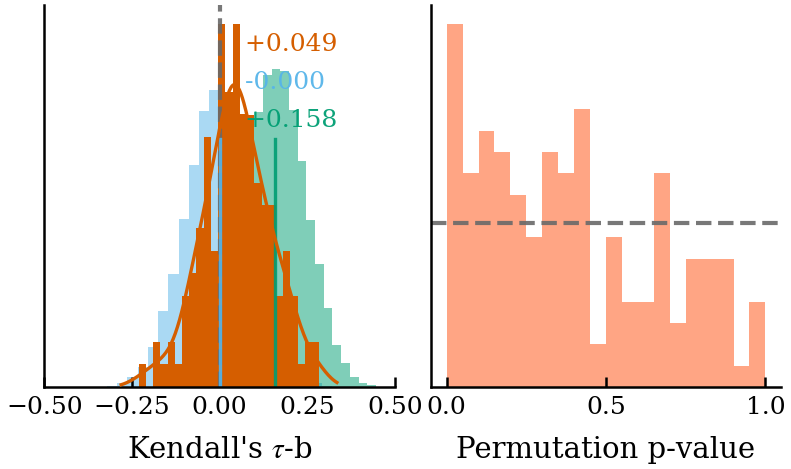}
    \caption{SciCode (prospective)}
    \label{fig:si:tau_histogram:prospective:scicode:confidence_before}
  \end{subfigure}
  \begin{subfigure}[b]{0.31\textwidth}
    \centering
    \includegraphics[width=\textwidth]{figures/squad/needs_context/tau_histogram_clean.png}
    \caption{SQuAD (prospective)}
    \label{fig:si:tau_histogram:prospective:squad:needs_context}
  \end{subfigure}
  \caption{\textbf{Pairwise Kendall $\tau$}, prospective (pre-task) judgments. Pairwise $\tau$-b between $\Delta$performance and $\Delta$metacognition over $\binom{n_\text{models}}{2}$ pairs. Left: orange is observed; blue is a base-rate-matched null (perf$_A$, perf$_B$, conf$_A$, conf$_B$ each shuffled independently); green is a calibration-preserving null (each model's perf and conf permuted together, calibration profile fixed). Solid verticals mark the three means; dashed grey marks $\tau=0$. Right: per-pair permutation $p$-values against the dashed grey uniform-null reference. Effect size is small but statistically significant with more highly significant pairs than expected if sampling from the null distribution. The observed mean sits significantly below the calibration-preserving null, implying that almost all calibration is shared rather than individuated.}
  \label{fig:si:tau_histogram:prospective}
\end{figure}

\begin{figure}[H]
  \centering
  \begin{subfigure}[b]{0.31\textwidth}
    \centering
    \includegraphics[width=\textwidth]{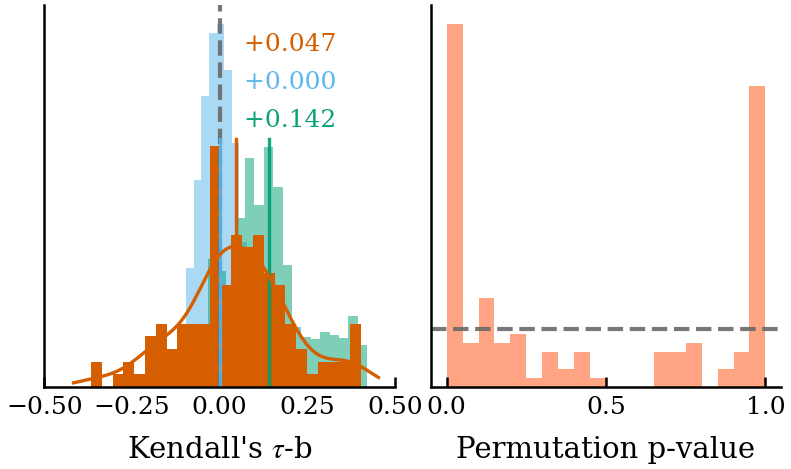}
    \caption{MathBench (counterfactual)}
    \label{fig:si:tau_histogram:counterfactual:math_bench:confidence_after}
  \end{subfigure}
  \begin{subfigure}[b]{0.31\textwidth}
    \centering
    \includegraphics[width=\textwidth]{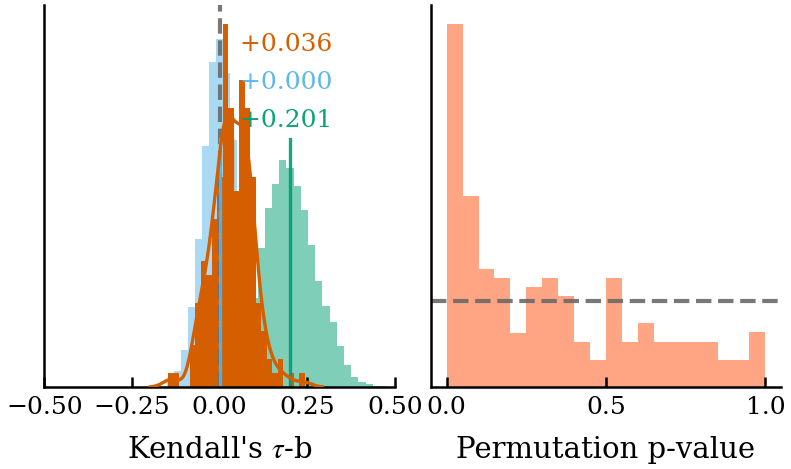}
    \caption{MMLU (counterfactual)}
    \label{fig:si:tau_histogram:counterfactual:mmlu:choices_necessary}
  \end{subfigure}
  \begin{subfigure}[b]{0.31\textwidth}
    \centering
    \includegraphics[width=\textwidth]{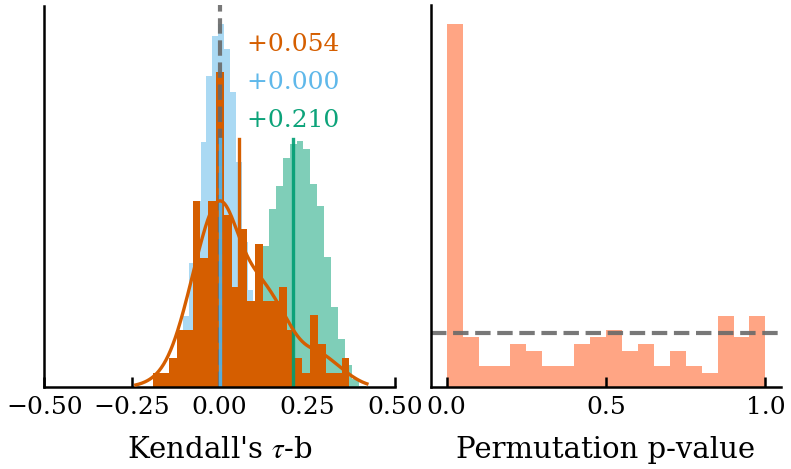}
    \caption{omnimath (counterfactual)}
    \label{fig:si:tau_histogram:counterfactual:omnimath:confidence_after}
  \end{subfigure}
  \\[1ex]
  \begin{subfigure}[b]{0.31\textwidth}
    \centering
    \includegraphics[width=\textwidth]{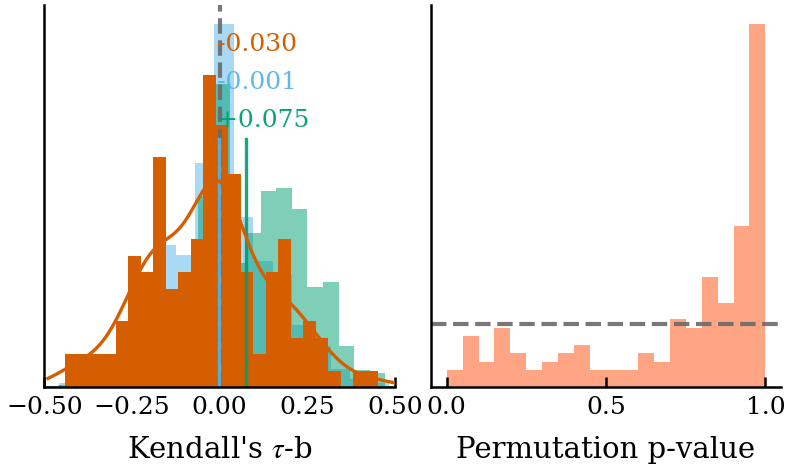}
    \caption{SciCode (counterfactual)}
    \label{fig:si:tau_histogram:counterfactual:scicode:confidence_after}
  \end{subfigure}
  \begin{subfigure}[b]{0.31\textwidth}
    \centering
    \includegraphics[width=\textwidth]{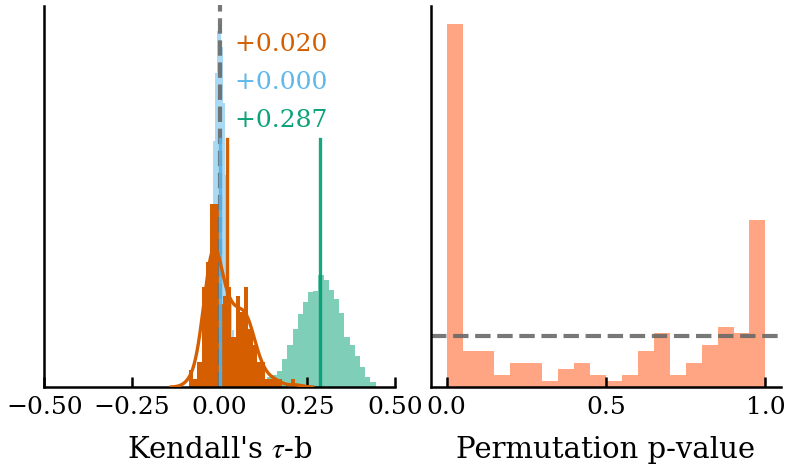}
    \caption{SQuAD (counterfactual)}
    \label{fig:si:tau_histogram:counterfactual:squad:context_necessary}
  \end{subfigure}
  \caption{\textbf{Pairwise Kendall $\tau$}, counterfactual (post-task) judgments. Pairwise $\tau$-b between $\Delta$performance and $\Delta$metacognition over $\binom{n_\text{models}}{2}$ pairs. Left: orange is observed; blue is a base-rate-matched null (perf$_A$, perf$_B$, conf$_A$, conf$_B$ each shuffled independently); green is a calibration-preserving null (each model's perf and conf permuted together, calibration profile fixed). Solid verticals mark the three means; dashed grey marks $\tau=0$. Right: per-pair permutation $p$-values against the dashed grey uniform-null reference. Effect size is small but statistically significant with more highly significant pairs than expected if sampling from the null distribution. The observed mean sits significantly below the calibration-preserving null, implying that almost all calibration is shared rather than individuated.}
  \label{fig:si:tau_histogram:counterfactual}
\end{figure}

\subsection{Factor--performance alignment for SQuAD}
\label{app:si:filtered_alignment}

The alignment numbers reported in \S\ref{sec:alignment} are computed as follows. For each benchmark we have two binary matrices $X \in \{0,1\}^{n_\text{items} \times n_\text{models}}$, one for performance and one for confidence. We form the $n_\text{models} \times n_\text{models}$ tetrachoric correlation matrix $R$ between models and take its eigendecomposition $R = V\,\mathrm{diag}(\lambda)\,V^\top$, giving factor-analytic loadings $\Lambda = V \cdot \mathrm{diag}(\sqrt{\lambda})$. Per-item Factor 1 scores are the mean-centred projection
\begin{equation}
f_1[i] = \sum_{m=1}^{n_\text{models}} \tilde X_{im} \cdot \Lambda_{m,1},
\qquad
\tilde X_{im} = X_{im} - \bar X_m,
\end{equation}
where $\bar X_m$ is model $m$'s marginal yes-rate or correct-rate. Each item then has one performance score and one confidence score, and the alignment statistic is the Pearson $r$ between the two score vectors across items. Panels (c) and (d) plot the z-scored version of these vectors, which leaves $r$ unchanged but places both axes on a comparable scale. The filter in panel (d) restricts to the contentious-item subset, items where both cross-model rates lie in $[0.1, 0.9]$, isolating items with substantial inter-model disagreement on both axes.

SQuAD prospective shows $r = 0.591$ unfiltered, reducing to $r = 0.230$ on the $N = 1015$ items where both cross-model rates fall within $[0.1, 0.9]$. Per-condition alignment numbers for every benchmark are tabulated in main paper Table~\ref{tab:alignment_full}.

\subsection{F1 alignment scatter plots across all conditions}
\label{app:si:f1_alignment_collection}

Per-item F1 alignment scatters for every benchmark and probe condition. Each row of subfigures pairs the unfiltered scatter (left) with the $[0.1, 0.9]$-filtered scatter (right). Pearson $r$ values for every panel are tabulated in main paper Table~\ref{tab:alignment_full}.

\begin{figure}[H]
\centering
\begin{subfigure}[b]{0.48\textwidth}
\includegraphics[width=\textwidth]{figures/squad/needs_context/factor_correlation/f1_perf_vs_metacog_clean.png}
\caption{SQuAD prospective, unfiltered}
\end{subfigure}
\hfill
\begin{subfigure}[b]{0.48\textwidth}
\includegraphics[width=\textwidth]{figures/squad/needs_context/factor_correlation/f1_perf_vs_metacog_filtered_clean.png}
\caption{SQuAD prospective, filtered}
\end{subfigure}
\\[1ex]
\begin{subfigure}[b]{0.48\textwidth}
\includegraphics[width=\textwidth]{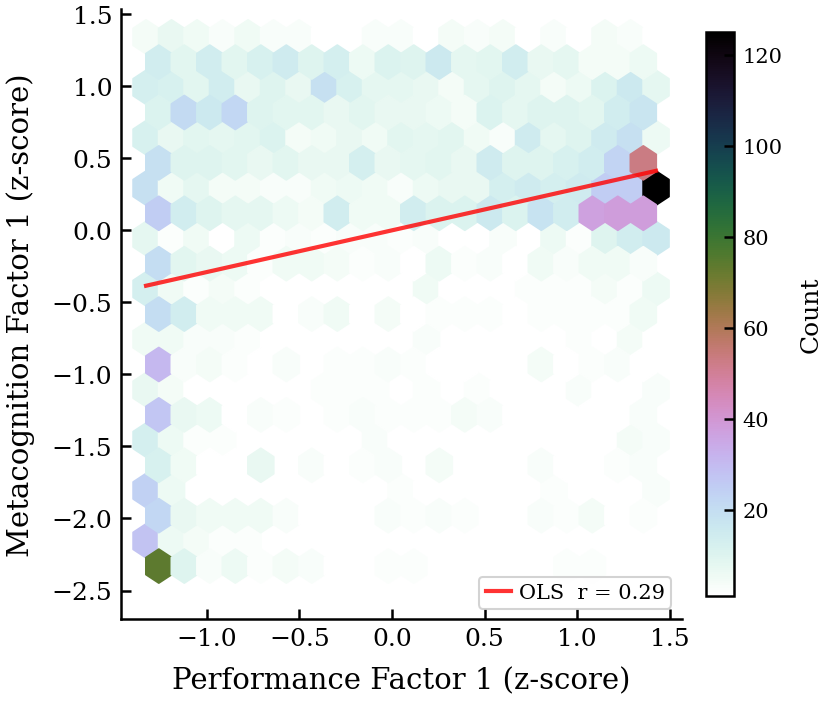}
\caption{SQuAD counterfactual, unfiltered}
\end{subfigure}
\hfill
\begin{subfigure}[b]{0.48\textwidth}
\includegraphics[width=\textwidth]{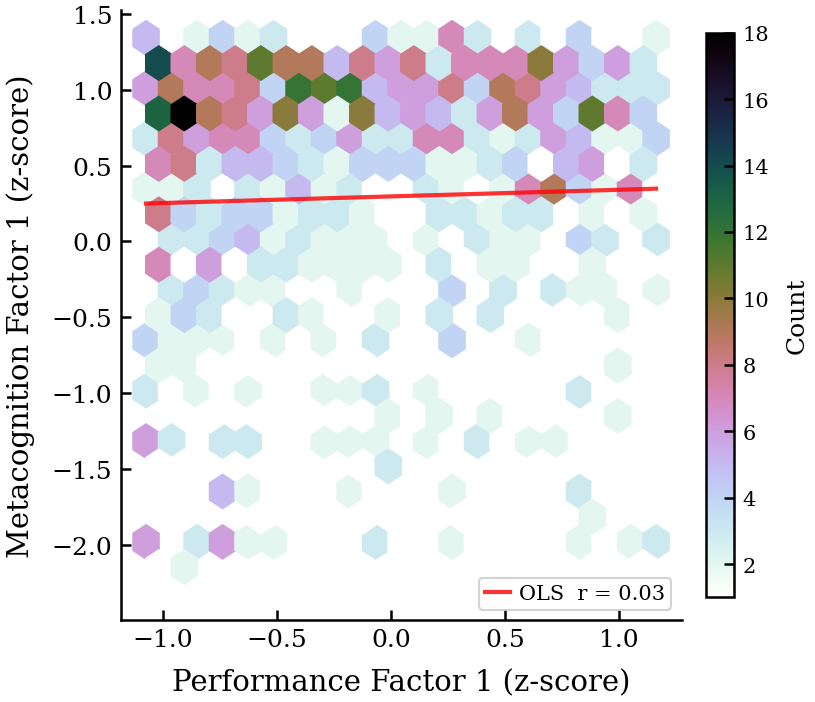}
\caption{SQuAD counterfactual, filtered}
\end{subfigure}
\caption{SQuAD F1 alignment scatter plots across both probe conditions.}
\label{fig:si:f1_alignment_squad}
\end{figure}

\begin{figure}[H]
\centering
\begin{subfigure}[b]{0.48\textwidth}
\includegraphics[width=\textwidth]{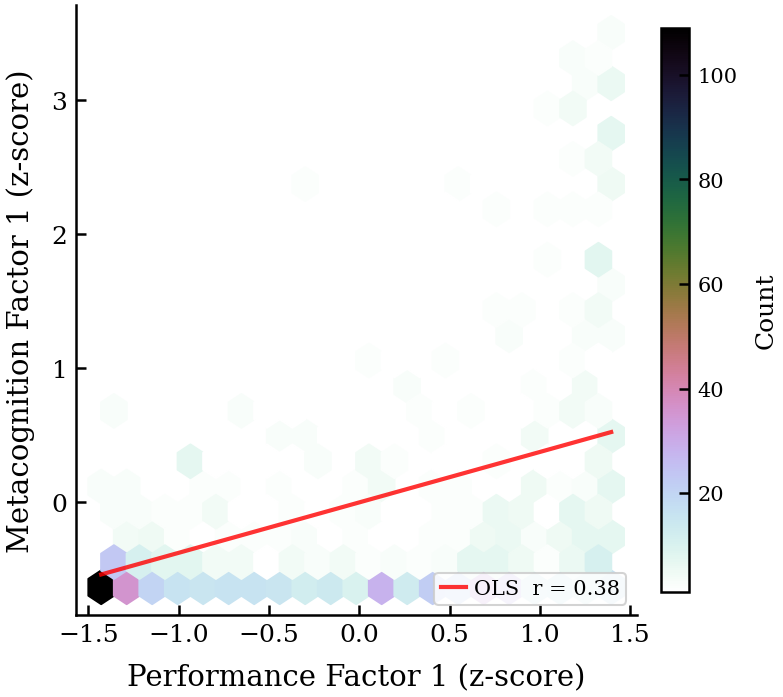}
\caption{MMLU-Pro prospective, unfiltered}
\end{subfigure}
\hfill
\begin{subfigure}[b]{0.48\textwidth}
\includegraphics[width=\textwidth]{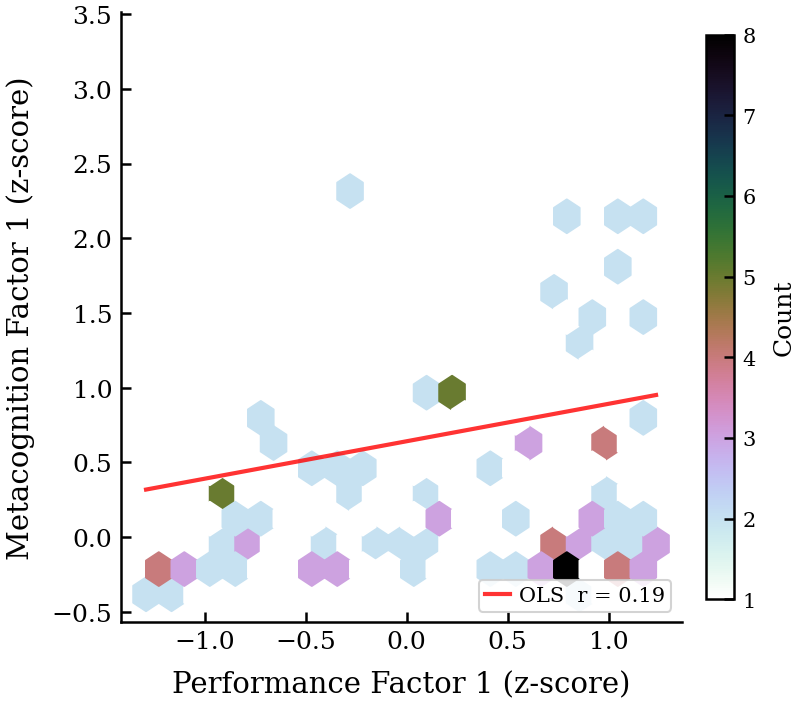}
\caption{MMLU-Pro prospective, filtered}
\end{subfigure}
\\[1ex]
\begin{subfigure}[b]{0.48\textwidth}
\includegraphics[width=\textwidth]{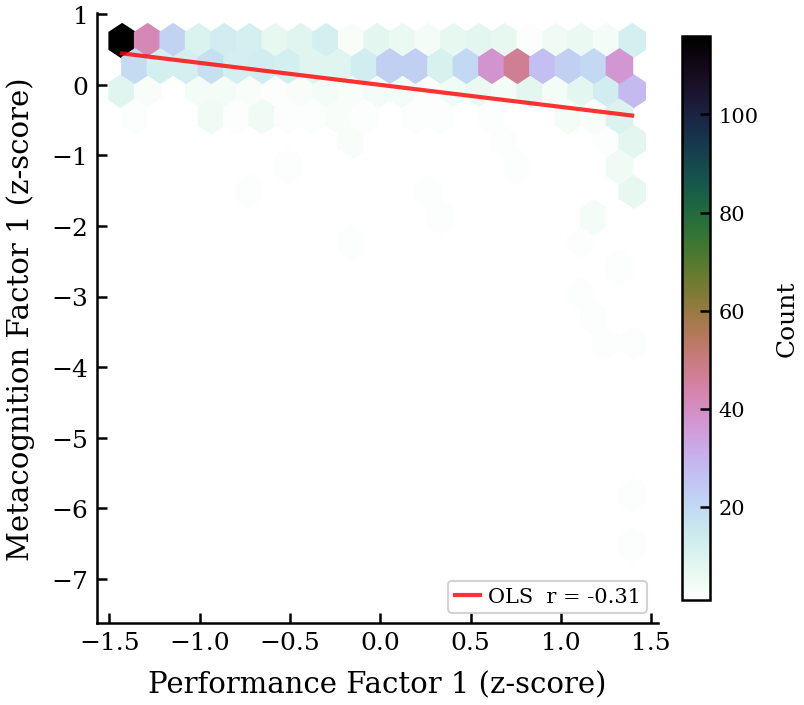}
\caption{MMLU-Pro counterfactual, unfiltered}
\end{subfigure}
\hfill
\begin{subfigure}[b]{0.48\textwidth}
\includegraphics[width=\textwidth]{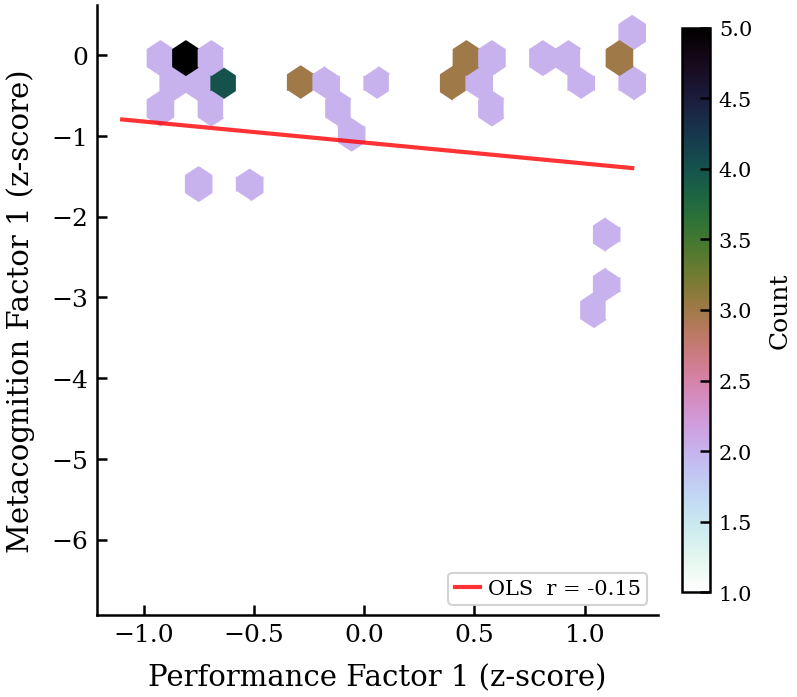}
\caption{MMLU-Pro counterfactual, filtered}
\end{subfigure}
\caption{MMLU-Pro F1 alignment scatter plots across all probe conditions.}
\label{fig:si:f1_alignment_mmlu}
\end{figure}

\begin{figure}[H]
\centering
\begin{subfigure}[b]{0.48\textwidth}
\includegraphics[width=\textwidth]{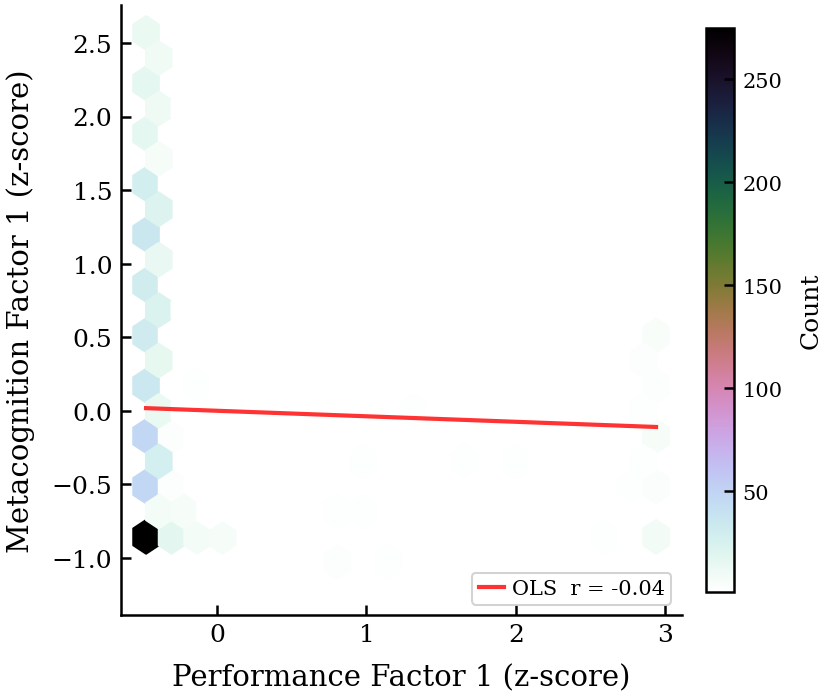}
\caption{LegalBench prospective, unfiltered}
\end{subfigure}
\hfill
\begin{subfigure}[b]{0.48\textwidth}
\includegraphics[width=\textwidth]{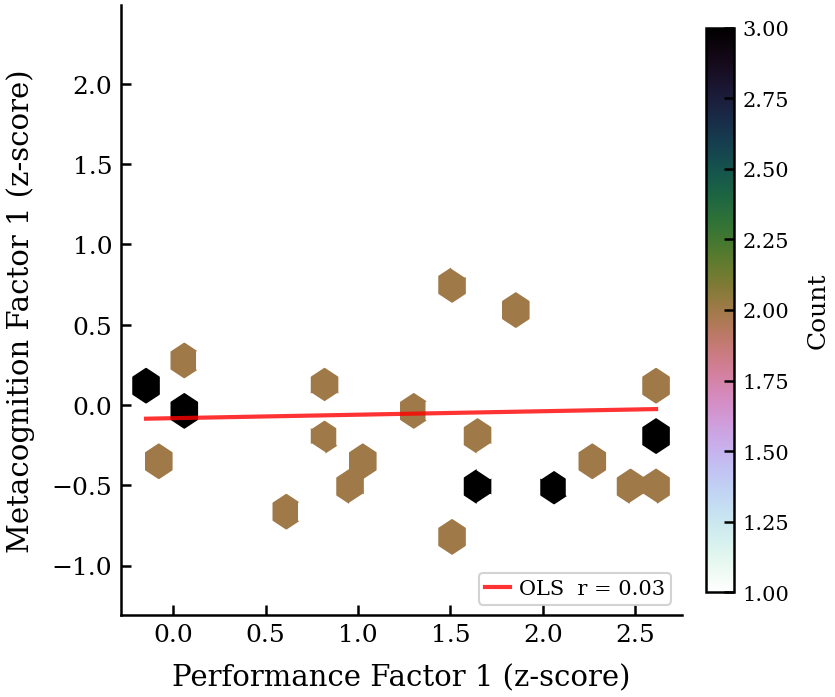}
\caption{LegalBench prospective, filtered}
\end{subfigure}
\caption{LegalBench F1 alignment scatter plots.}
\label{fig:si:f1_alignment_legalbench}
\end{figure}

\begin{figure}[H]
\centering
\begin{subfigure}[b]{0.48\textwidth}
\includegraphics[width=\textwidth]{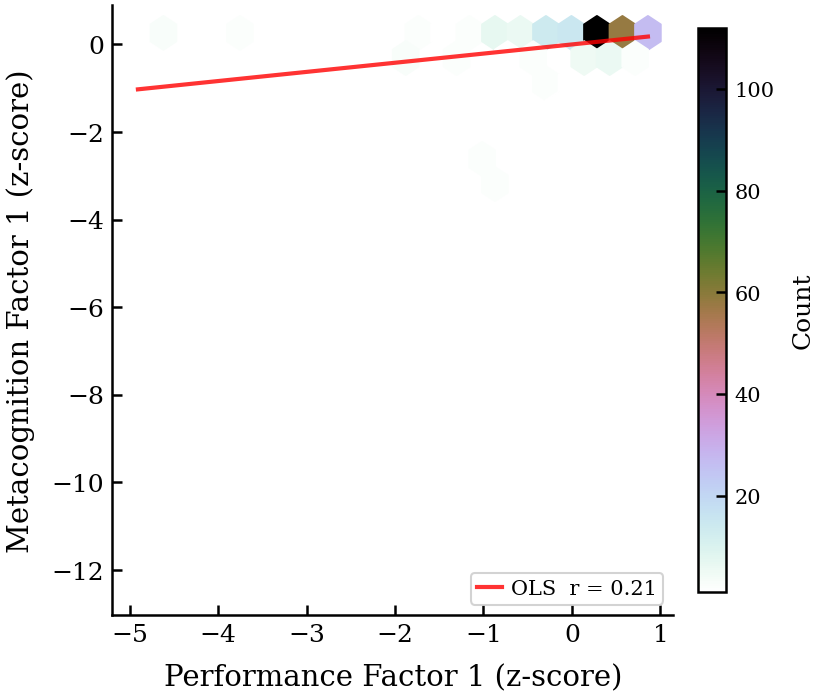}
\caption{MathBench prospective, unfiltered}
\end{subfigure}
\hfill
\begin{subfigure}[b]{0.48\textwidth}
\includegraphics[width=\textwidth]{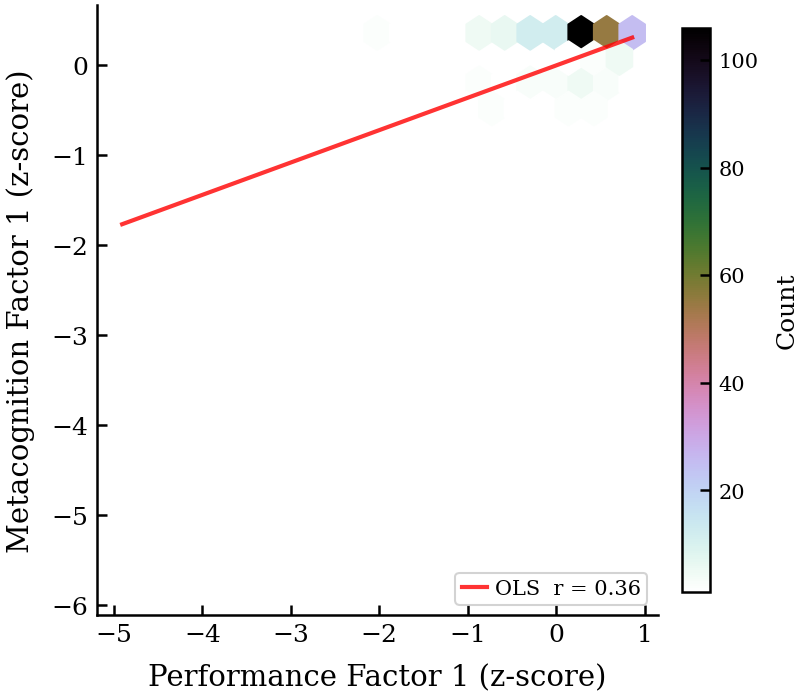}
\caption{MathBench counterfactual, unfiltered}
\end{subfigure}
\\[1ex]
\begin{subfigure}[b]{0.48\textwidth}
\includegraphics[width=\textwidth]{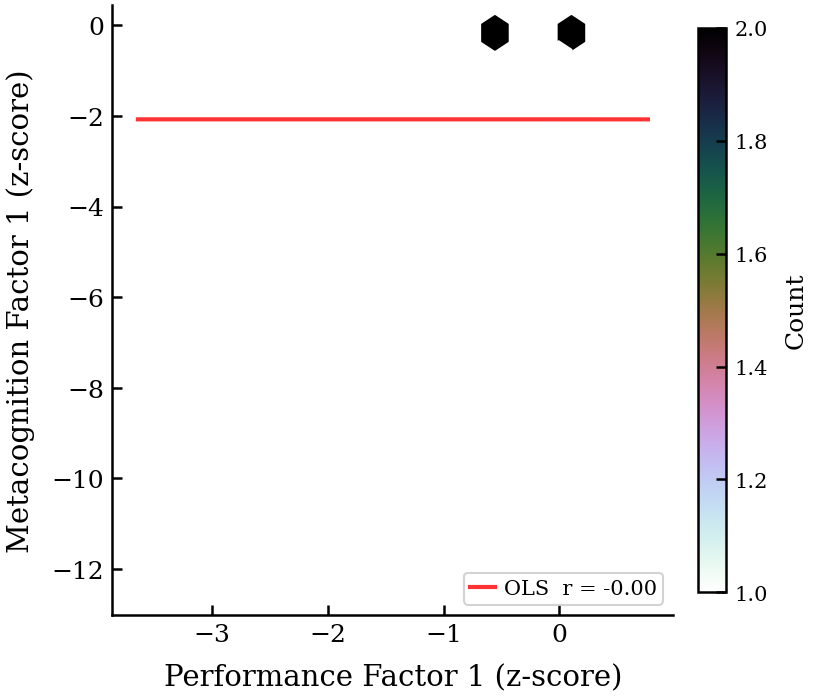}
\caption{MathBench prospective, filtered}
\end{subfigure}
\hfill
\begin{subfigure}[b]{0.48\textwidth}
\includegraphics[width=\textwidth]{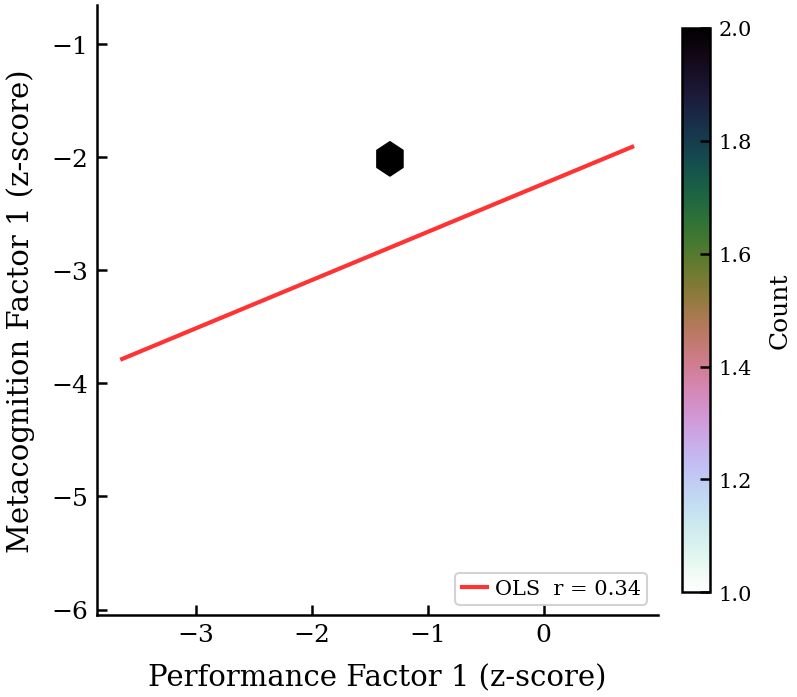}
\caption{MathBench counterfactual, filtered}
\end{subfigure}
\caption{MathBench F1 alignment scatter plots across both probe conditions.}
\label{fig:si:f1_alignment_math_bench}
\end{figure}

\begin{figure}[H]
\centering
\begin{subfigure}[b]{0.48\textwidth}
\includegraphics[width=\textwidth]{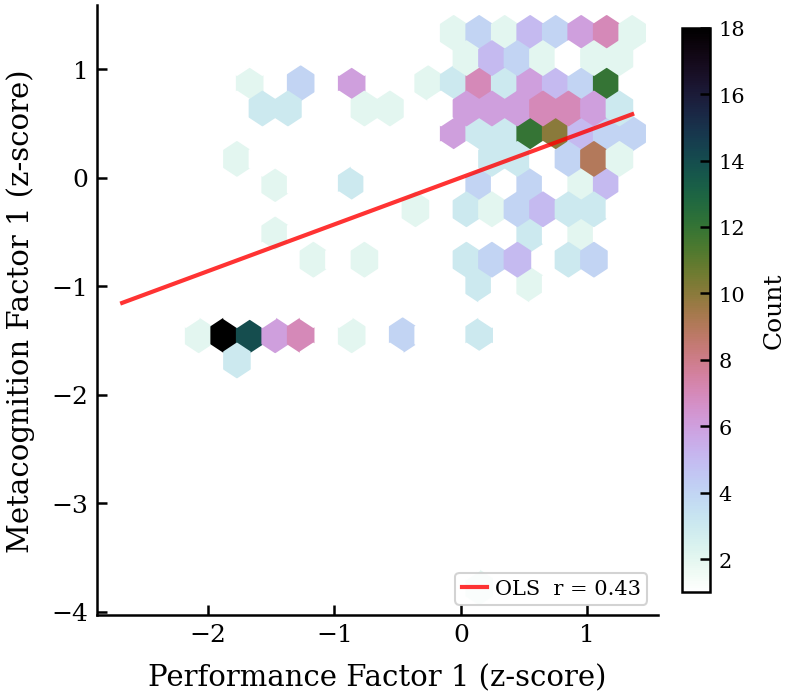}
\caption{Omni-MATH prospective, unfiltered}
\end{subfigure}
\hfill
\begin{subfigure}[b]{0.48\textwidth}
\includegraphics[width=\textwidth]{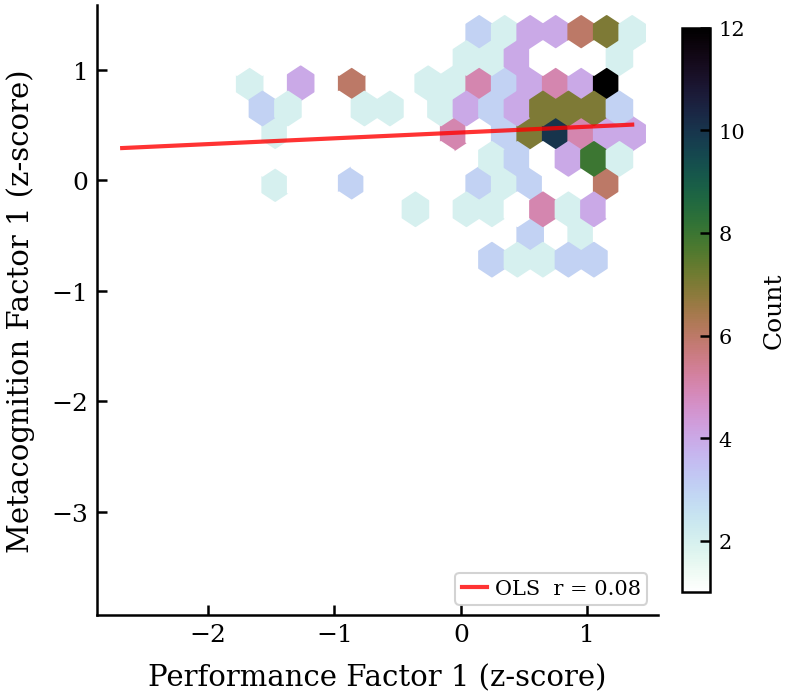}
\caption{Omni-MATH prospective, filtered}
\end{subfigure}
\\[1ex]
\begin{subfigure}[b]{0.48\textwidth}
\includegraphics[width=\textwidth]{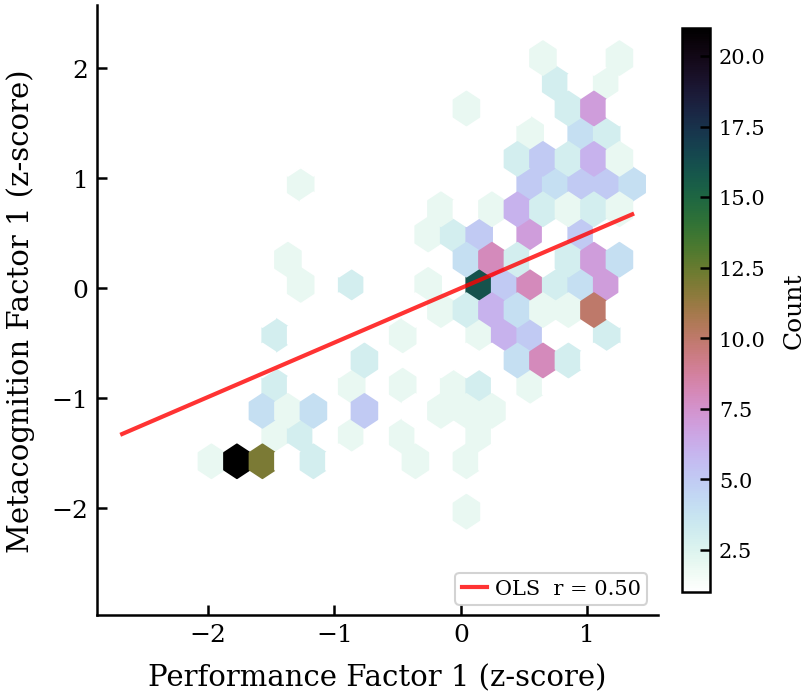}
\caption{Omni-MATH counterfactual, unfiltered}
\end{subfigure}
\hfill
\begin{subfigure}[b]{0.48\textwidth}
\includegraphics[width=\textwidth]{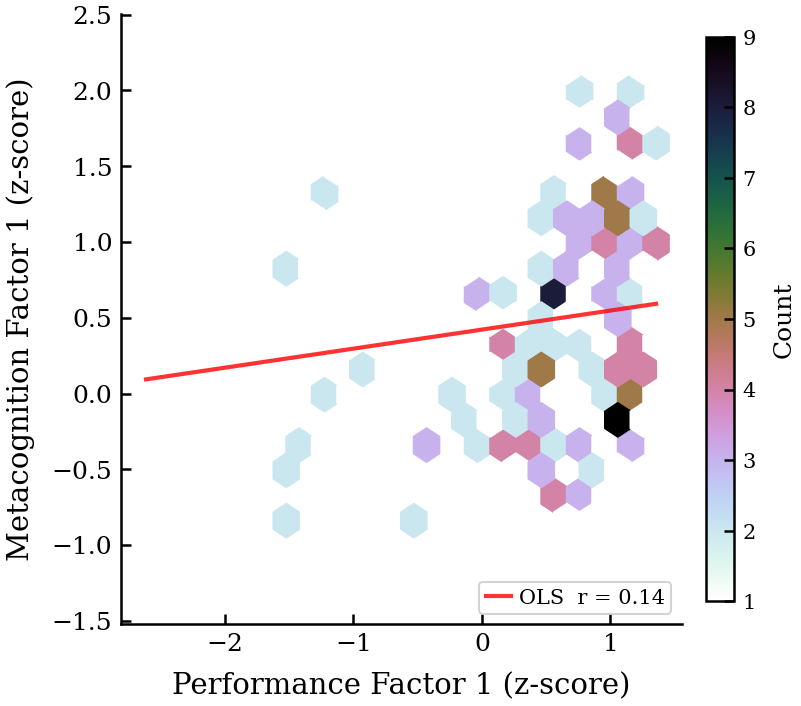}
\caption{Omni-MATH counterfactual, filtered}
\end{subfigure}
\caption{Omni-MATH F1 alignment scatter plots across both probe conditions.}
\label{fig:si:f1_alignment_omnimath}
\end{figure}

\begin{figure}[H]
\centering
\begin{subfigure}[b]{0.48\textwidth}
\includegraphics[width=\textwidth]{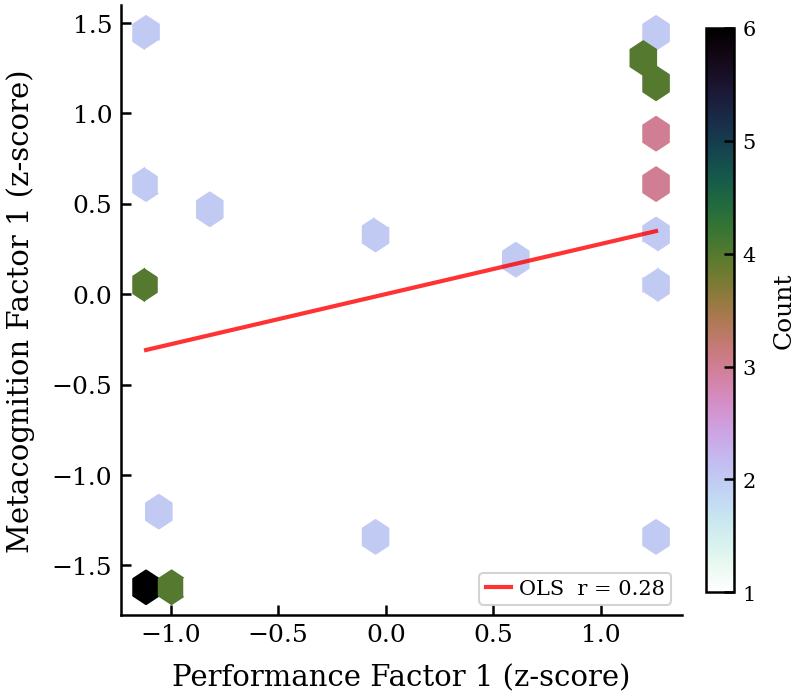}
\caption{SciCode prospective, unfiltered}
\end{subfigure}
\hfill
\begin{subfigure}[b]{0.48\textwidth}
\includegraphics[width=\textwidth]{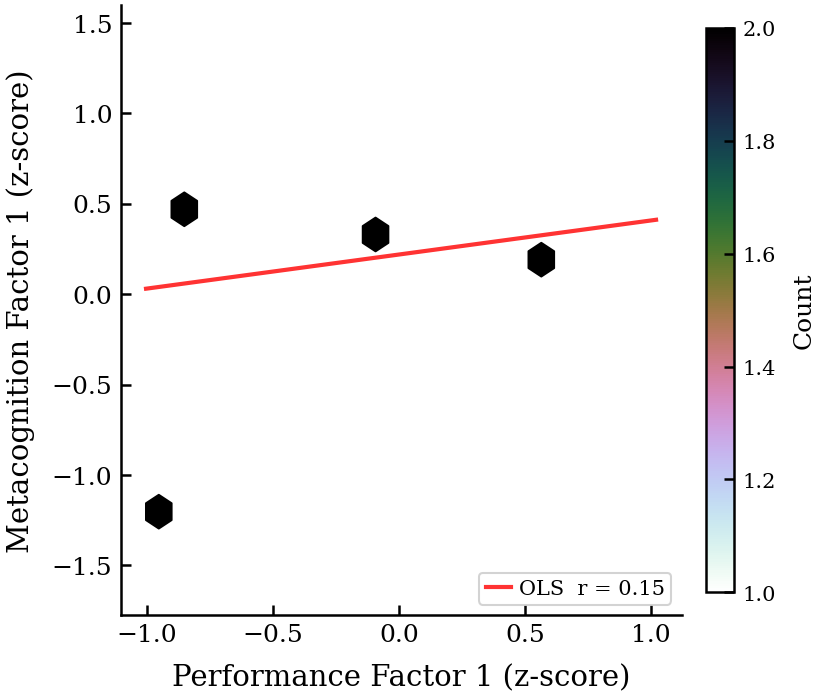}
\caption{SciCode prospective, filtered}
\end{subfigure}
\\[1ex]
\begin{subfigure}[b]{0.48\textwidth}
\includegraphics[width=\textwidth]{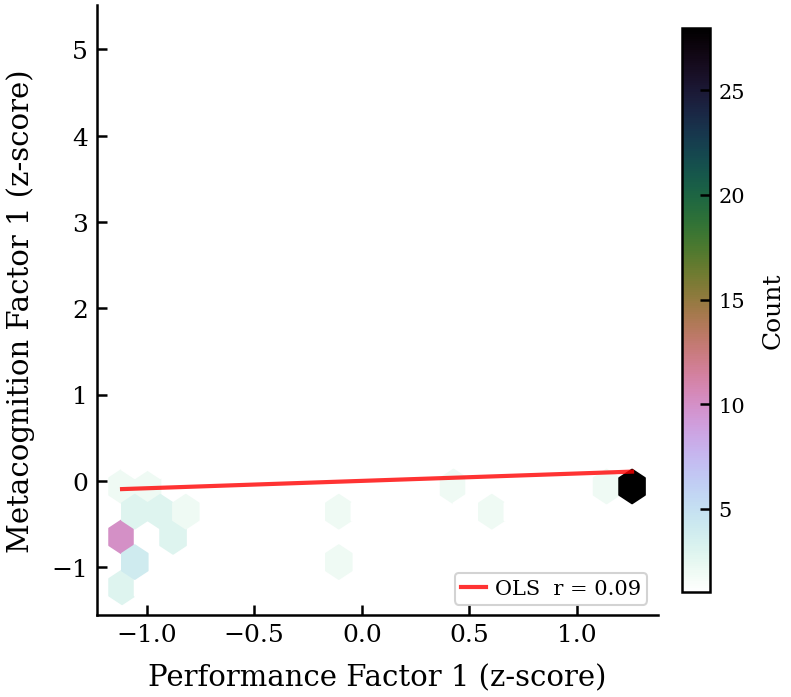}
\caption{SciCode counterfactual, unfiltered}
\end{subfigure}
\caption{SciCode F1 alignment scatter plots.}
\label{fig:si:f1_alignment_scicode}
\end{figure}

\subsection{Pairwise Kendall-$\tau$ versus performance gap across all conditions}
\label{app:si:tau_vs_performance}

Each panel plots mean pairwise Kendall's $\tau$ (left $y$-axis, pink curve) over the model pairs whose absolute performance gap is at most the threshold on the $x$-axis. The pink shaded band is the confidence interval on the mean. The grey curve and right $y$-axis show the number of pairs admitted at each threshold; tightening the gap restriction (moving leftward) reduces the admitted-pair count. The load-bearing test for \S\ref{sec:alignment} is whether mean $\tau$ falls toward zero as the gap restriction tightens. A fall indicates that the unfiltered $\tau$ was carried by base-rate differences along the shared difficulty axis. A flat curve indicates $\tau$ that survives at matched performance. Items in this analysis are restricted to those with cross-model rates in $[0.25, 0.75]$, a tighter disagreement window than the $[0.1, 0.9]$ filter applied in the F1-alignment table (main paper Table~\ref{tab:alignment_full}); the tighter window stabilises pair-level $\tau$ estimates against floor- and ceiling-dominated items.

\begin{figure}[H]
\centering
\begin{subfigure}[b]{0.48\textwidth}
\includegraphics[width=\textwidth]{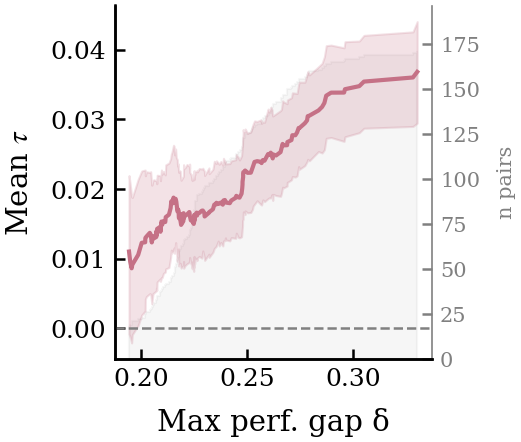}
\caption{SQuAD prospective}
\end{subfigure}
\hfill
\begin{subfigure}[b]{0.48\textwidth}
\includegraphics[width=\textwidth]{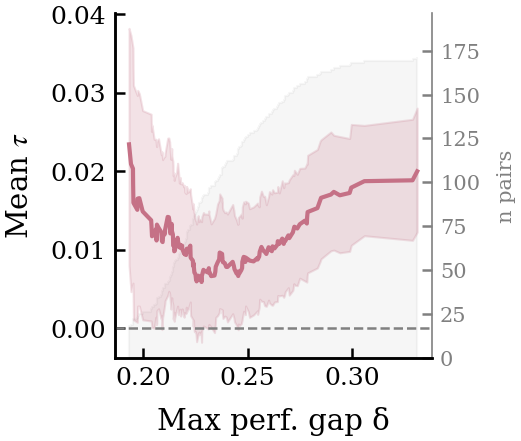}
\caption{SQuAD counterfactual}
\end{subfigure}
\caption{SQuAD pairwise Kendall-$\tau$ vs.\ performance gap across both probe conditions.}
\label{fig:si:tau_vs_perf_squad}
\end{figure}

\begin{figure}[H]
\centering
\begin{subfigure}[b]{0.48\textwidth}
\includegraphics[width=\textwidth]{figures/mmlu/confidence_without_choices/tau_vs_performance_clean.png}
\caption{MMLU-Pro prospective}
\end{subfigure}
\hfill
\begin{subfigure}[b]{0.48\textwidth}
\includegraphics[width=\textwidth]{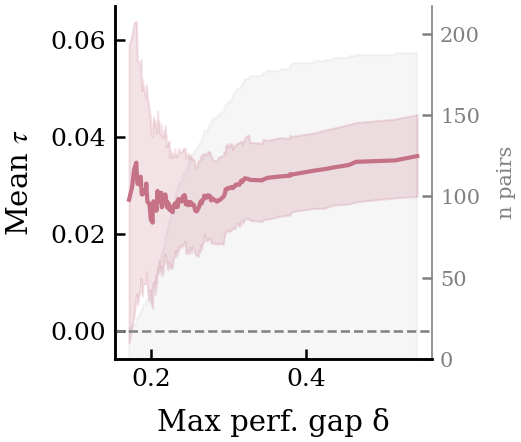}
\caption{MMLU-Pro counterfactual}
\end{subfigure}
\caption{MMLU-Pro pairwise Kendall-$\tau$ vs.\ performance gap across both probe conditions.}
\label{fig:si:tau_vs_perf_mmlu}
\end{figure}

\begin{figure}[H]
\centering
\includegraphics[width=0.48\textwidth]{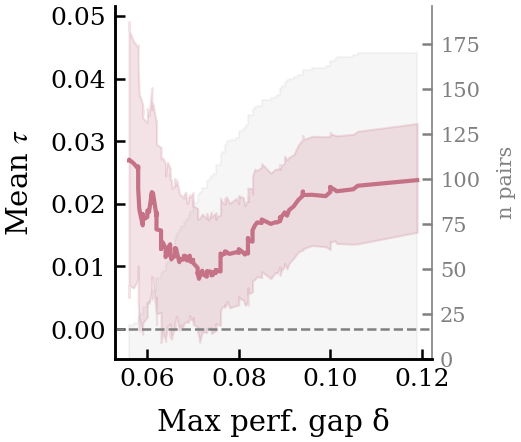}
\caption{LegalBench pairwise Kendall-$\tau$ vs.\ performance gap.}
\label{fig:si:tau_vs_perf_legalbench}
\end{figure}

\begin{figure}[H]
\centering
\begin{subfigure}[b]{0.48\textwidth}
\includegraphics[width=\textwidth]{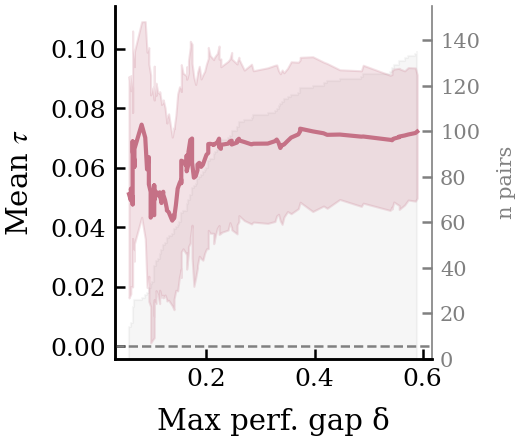}
\caption{MathBench prospective}
\end{subfigure}
\hfill
\begin{subfigure}[b]{0.48\textwidth}
\includegraphics[width=\textwidth]{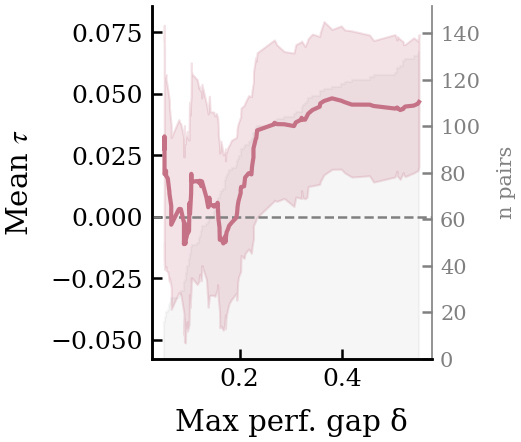}
\caption{MathBench counterfactual}
\end{subfigure}
\caption{MathBench pairwise Kendall-$\tau$ vs.\ performance gap across both probe conditions.}
\label{fig:si:tau_vs_perf_math_bench}
\end{figure}

\begin{figure}[H]
\centering
\begin{subfigure}[b]{0.48\textwidth}
\includegraphics[width=\textwidth]{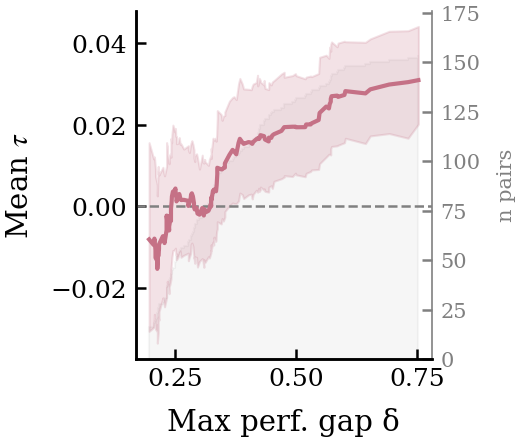}
\caption{Omni-MATH prospective}
\end{subfigure}
\hfill
\begin{subfigure}[b]{0.48\textwidth}
\includegraphics[width=\textwidth]{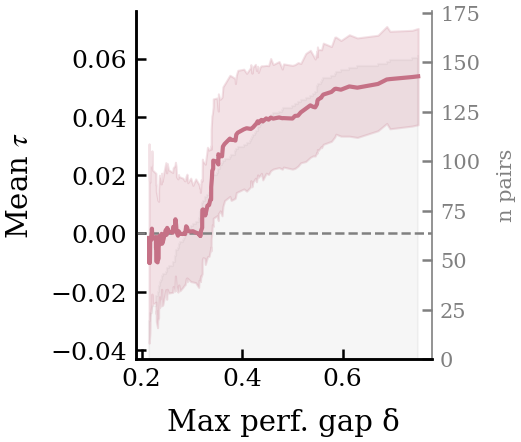}
\caption{Omni-MATH counterfactual}
\end{subfigure}
\caption{Omni-MATH pairwise Kendall-$\tau$ vs.\ performance gap across both probe conditions.}
\label{fig:si:tau_vs_perf_omnimath}
\end{figure}

\begin{figure}[H]
\centering
\begin{subfigure}[b]{0.48\textwidth}
\includegraphics[width=\textwidth]{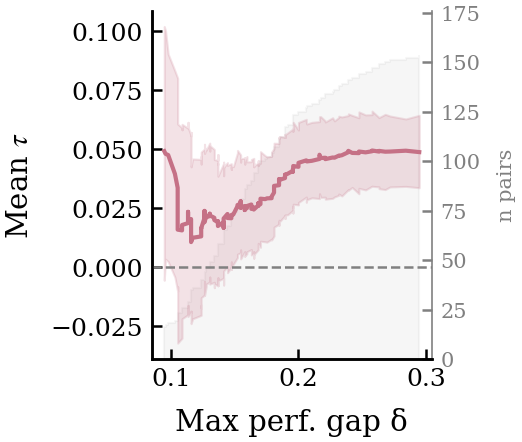}
\caption{SciCode prospective}
\end{subfigure}
\hfill
\begin{subfigure}[b]{0.48\textwidth}
\includegraphics[width=\textwidth]{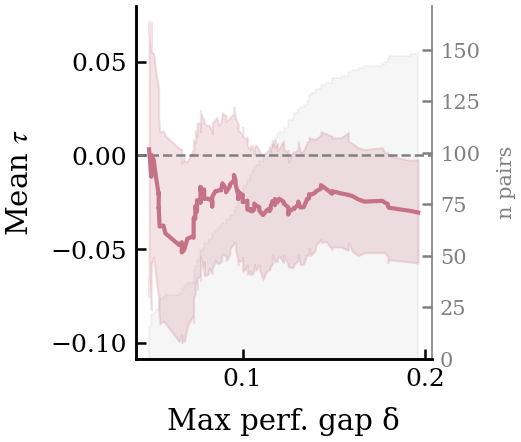}
\caption{SciCode counterfactual}
\end{subfigure}
\caption{SciCode pairwise Kendall-$\tau$ vs.\ performance gap across both probe conditions.}
\label{fig:si:tau_vs_perf_scicode}
\end{figure}

\subsection{Null distribution methodologies}
\label{app:si:nulls}

Three empirical null procedures are used in the paper, all bootstrap-based. They differ in what they preserve about the data and what they shuffle. Prospective and counterfactual probes are processed independently throughout, since marginal yes-rates differ between probe types and the nulls are re-fit accordingly.

\noindent\textbf{Eigenspectrum null (column-shuffle on the judgement matrix).} For each draw, each model's column of binary judgements in the $n_\text{items} \times n_\text{models}$ matrix is independently permuted across items, with the missingness pattern preserved. The tetrachoric correlation matrix is re-estimated on the permuted matrix and eigendecomposed. The per-rank 95th percentile of the resulting eigenvalue distribution is overlaid as the red dashed curve on each observed spectrum (Fig.~\ref{fig:eigenspectra}). Observed eigenvalues above their rank's 95th percentile are classified as signal, below as noise. The null preserves each model's marginal yes-rate exactly while destroying all inter-model item-level correlation. Implementation: \texttt{parallel\_analysis} with \texttt{data\_matrix} set, invoked from each benchmark's \texttt{analyze\_correlations.py}, $B = 100$ draws per (benchmark, probe).

\noindent\textbf{Pairwise $\tau$, base-rate-matched null.} For each model pair $(A, B)$, the four binary vectors $\text{perf}_A, \text{perf}_B, \text{conf}_A, \text{conf}_B$ are each \emph{independently} permuted (four independent permutations per draw), then Kendall $\tau$-b is computed between $(\text{perf}_A' - \text{perf}_B')$ and $(\text{conf}_A' - \text{conf}_B')$. Each vector's marginal yes-rate is preserved; within-model perf--conf calibration, cross-model perf--perf and conf--conf alignment, and any shared item-level difficulty axis are all destroyed. Under this null $\mathbb{E}[\tau] \approx 0$ because all four covariances in the $\tau$ decomposition vanish in expectation. This is the blue distribution in the histograms of Fig.~\ref{fig:tau_hist}. Implementation: \texttt{sample\_base\_rate\_matched\_null\_taus}, 100 draws per pair, pooled across the $\binom{n_\text{models}}{2}$ pairs.

\noindent\textbf{Pairwise $\tau$, calibration-preserving null.} For each pair, the same random permutation $\sigma_A$ is applied to both $\text{perf}_A$ and $\text{conf}_A$ as a unit, and an independent $\sigma_B$ to $\text{perf}_B$ and $\text{conf}_B$ as a unit, then Kendall $\tau$-b is computed as before. Each model's marginal yes-rate \emph{and} its within-model perf--conf covariance (its individual calibration profile) are preserved; only item identity is randomised between models. Under this null the mean $\tau$ sits at the sum of within-model perf--conf covariances, the level of pairwise calibration that would obtain if each model had correct self-calibration but the signals did not align with each other on which items are gotten right. The observed pairwise $\tau$ mean sitting significantly below this null mean (green distribution in Fig.~\ref{fig:tau_hist}, with the exception of MathBench) is the central paper result: almost all observed pairwise calibration is shared rather than individuated. Implementation: \texttt{sample\_calibration\_preserving\_null\_taus}, 100 draws per pair.

\noindent\textbf{Per-pair significance test (Fisher's exact on the $3 \times 3$).} For per-pair p-values rather than the pooled bootstrap distribution, we use Fisher's exact test~\citep{5d8fa06c-0d2a-391a-b97f-85223f1f3b0f} on the $3 \times 3$ table of (sign of $\Delta\text{perf}$, sign of $\Delta\text{conf}$) cells over items, taking the probability of the observed table under fixed row and column marginals. This is the analytic complement to a per-pair permutation test and yields the right-panel p-value distributions in Fig.~\ref{fig:tau_hist}.

\subsection{Inline-solve detector}
\label{app:si:inline_solve_detector}

The detector classifies a model's free-text confidence reasoning as an inline solve attempt when it scores at least three out of five features. Each feature corresponds to a regular expression run over the response text, with one length predicate.

\begin{enumerate}
\item \textbf{Solution-convergence phrases}: ``therefore'', ``which gives'', ``so $\langle$var$\rangle\,=\,\ldots$'', ``thus $\ldots\,=\,\ldots$'', or ``we get''.
\item \textbf{Mathematical notation beyond the question stem}: any LaTeX command (\texttt{\textbackslash frac}, \texttt{\textbackslash sqrt}, or \verb|\command{|), inline math delimiters (\verb|$\ldots$|), or superscripts.
\item \textbf{Computation markers}: stems of \emph{substitut-}, \emph{integrat-}, \emph{differentiat-}, \emph{expand-}, \emph{simplif-}, \emph{calculat-}, or \emph{complet-}.
\item \textbf{Numbered solution steps}: lines beginning with a digit followed by ``)'' or ``.'' (e.g.\ ``1) Compute $\ldots$'').
\item \textbf{Length}: response longer than 120 words.
\end{enumerate}

The threshold of three was chosen to require multiple independent signatures rather than any single cue. With one or two features, hits are common in confidence judgements that merely reference the question domain (``this is an integration problem and I think I can do it''). With three or more, the response typically contains both a partial computation and the surface markers of formal solution writing.

We have no ground truth labels for inline solving, and the boundary between metacognitive reasoning and partial computation is fuzzy. To validate the detector we manually inspected a sample of classifications across both flagged and unflagged responses for each reasoning-trained model until the divide seemed reasonable. The difference in rates between the model categories is not subtle, so we did not optimize this classification process any further.

\subsection{Eigenstructure of metacognitive correlation matrices (prospective conditions)}
\label{app:si:eigenstructure}

\begin{table}[h]
\centering
\caption{Eigenstructure of metacognitive tetrachoric correlation matrices for the prospective probe of each benchmark. $\lambda_1$ and $\lambda_2$ are the first and second eigenvalues, normalised by $n_\text{models}$ as in \S\ref{sec:setup} so each value is the fraction of total variance carried by that component and the full spectrum sums to one. $\lambda_1/\lambda_2$ is the dominance ratio, invariant to the normalisation.}
\label{tab:si:eigenstructure}
\small
\begin{tabular}{lccc}
\toprule
Benchmark & $\lambda_1$ & $\lambda_2$ & $\lambda_1/\lambda_2$ \\
\midrule
SciCode & $0.37$ & $0.18$ & $2.03$ \\
LegalBench & $0.36$ & $0.31$ & $1.17$ \\
MathBench & $0.25$ & $0.12$ & $2.15$ \\
MMLU & $0.48$ & $0.17$ & $2.81$ \\
OmniMath & $0.36$ & $0.21$ & $1.70$ \\
SQuAD & $0.55$ & $0.29$ & $1.88$ \\
\bottomrule
\end{tabular}
\end{table}

\section{Marginal yes-rate (base rate) handling}
\label{app:si:yesrate}

The marginal yes-rate plays a central role throughout the paper. Several distinct terms refer to the same quantity:

\begin{itemize}
\item $p_i = \mathbb{P}(X_i = 1)$, the empirical share of yes answers from model $i$ on a benchmark.
\item $\tau_i$, the latent threshold in the tetrachoric model (\S\ref{app:si:tetrachoric}), with $p_i = \Phi(\tau_i)$.
\item ``Response threshold'' and ``confidence base rate'' are used interchangeably with $p_i$ in figure captions and prose.
\end{itemize}

\noindent The yes-rate is the only one-dimensional summary of a model's marginal behaviour on a benchmark, and is the structure preserved by every null distribution used in the paper (\S\ref{app:si:nulls}). Three observations make it the central nuisance variable to control for.

\noindent\textbf{Yes-rates differ substantially across models.} Within every benchmark, the highest- and lowest-base-rate models are separated by several standard deviations in $\Phi^{-1}(p_i)$. Confidence biases are a property of model identity rather than question content (Fig.~\ref{fig:confidence_consistency}).

\noindent\textbf{Yes-rates correlate across benchmarks.} A model that is over-confident on one benchmark and probe condition is typically over-confident on others. The per-model bias travels: Fig.~\ref{fig:confidence_consistency} z-scored boxes are tight relative to the between-model spread along the model ordering.

\noindent\textbf{Two models with disjoint yes-rates can show high naive pairwise calibration with no shared item-level signal.} Pearson and Kendall correlations on binary vectors are bounded above by their marginals; the bound is reached when, e.g., one model says ``yes'' to nearly every item the other says ``no'' to, regardless of whether either is right. Pairwise $\tau$-b on $(\Delta\text{perf}, \Delta\text{conf})$ inherits this marginal bound. All inter-model analyses in the paper therefore either (a) condition on the matched-base-rate subset (Fig.~\ref{fig:climax}(a--c), restriction to model pairs with absolute base-rate gap $\leq \delta$) or (b) compare to a base-rate-preserving null (\S\ref{app:si:nulls}). The tetrachoric correlation (\S\ref{app:si:tetrachoric}) circumvents the marginal bound on the factor side by inferring a latent continuous variable beneath the threshold, but the tetrachoric estimate itself depends on the marginal yes-rates entering as fixed thresholds.

\noindent\textbf{Factor-loading shrinkage at extreme base rates.} The first-factor loading $\Lambda_{i,1}$ is empirically parabolic in $\Phi^{-1}(p_i)$: extreme-base-rate models have shrunken loadings because near-floor and near-ceiling responders contribute little discriminating information per pair relative to noise. This shrinkage inflates the apparent intrinsic dimensionality of the data: the second eigenvalue absorbs variance that the first eigenvalue would carry if loadings were uniform across models, so the spectral signature of a true one-factor structure is partially masked. The factor-overview plots in \S\ref{app:si:factor_overview} display this directly on a per-(benchmark, probe) basis.

\section{Tetrachoric estimator}
\label{app:si:tetrachoric}

For binary confidence judgments $X, Y \in \{0,1\}$ from two models with empirical yes-rates $p_X = P(X{=}1)$ and $p_Y = P(Y{=}1)$, the tetrachoric correlation~\citep{pearson1900} $\rho_{\text{tet}}$ is the value satisfying
\begin{equation}
P(X{=}1, Y{=}1) = \int_{h}^{\infty}\!\int_{k}^{\infty} \phi_2(z_1, z_2;\, \rho_{\text{tet}})\, dz_1\, dz_2
\label{eq:si:tetrachoric}
\end{equation}
where $\phi_2$ is the bivariate standard normal density with correlation $\rho_{\text{tet}}$ and the thresholds $h = \Phi^{-1}(p_X)$, $k = \Phi^{-1}(p_Y)$ are fixed by the empirical yes-rates. Concretely, we estimate $\rho_{\text{tet}}$ pairwise. Thresholds are fixed first from each model's marginal yes-rate, then $\rho_{\text{tet}}$ is found by one-dimensional root-finding so that the right-hand side reproduces the empirical ``both yes'' cell of the $2 \times 2$ contingency table over items.

\section{Per-Benchmark Question/Context Decomposition}
\label{app:decomposition}

Table~\ref{tab:decomposition_summary} summarises the question specification / context split for each benchmark.

\begin{table}[h]
\centering
\caption{Per-benchmark decomposition. ``Source'' indicates whether the split uses existing dataset fields or restructured data.}
\label{tab:decomposition_summary}
\small
\begin{tabular}{llll}
\toprule
Benchmark  & Question specification                      & Context held out                   & Source \\
\midrule
SQuAD      & Question text                               & Wikipedia passage                  & Dataset field \\
MMLU-Pro   & Question stem                               & Answer choices (A--J)              & Dataset field \\
LegalBench & Question + contract excerpt                 & ---                                & Dataset field \\
MathBench  & Problem statement                           & Worked solution (answer redacted)  & Dataset field + redaction \\
OmniMath   & Problem statement                           & Worked solution (answer redacted)  & Dataset field + redaction \\
SciCode    & Task spec + function header                 & Disambiguation + background        & Restructured \\
\bottomrule
\end{tabular}
\end{table}

\subsection{SQuAD}
\label{app:decomposition:squad}

We use SQuAD v1.1 dev split~\citep{rajpurkar2016}, randomly sampled to 1{,}000. The question text is the question specification, and the Wikipedia passage is the optional context. We use an LLM judge to grade semantic equivalence to the ground truth answer.

\subsection{MMLU-Pro}
\label{app:decomposition:mmlu}

We use the MMLU-Pro test split~\citep{wang2024mmlupro}, which contains 12{,}032 items across 14 subject categories, randomly sampled to 1{,}000. The question stem is the question specification, and the ten answer options (A--J) are the optional context. We use an LLM judge to grade semantic equivalence to the ground truth answer.

\subsection{LegalBench}
\label{app:decomposition:legalbench}

We used the LegalBench~\citep{guha2023legalbench} CUAD subset containing 38 binary clause-presence task types, 17{,}980 items, randomly sampled to 1{,}000. The question specification is the clause-type question together with the contract excerpt. There was no scope for a counterfactual probe with LegalBench, since the question is meaningless without the contract. We use an LLM judge to grade semantic equivalence to the ground truth answer.

\subsection{MathBench}
\label{app:decomposition:mathbench}

The question specification is the problem statement. The optional additional context is the worked solution with the final \verb|\boxed{}| redacted to \verb|\boxed{?}| via brace-matching on the last top-level box. Intermediate boxes are preserved. Both prospective and counterfactual probes are run. Grading is performed through SymPy symbolic equivalence, followed by string normalisation, with an LLM judge testing for semantic equivalence as a fallback.

\subsection{OmniMath}
\label{app:decomposition:omnimath}

The question specification is the problem statement. The optional additional context is the complete solution with the final answer assertion replaced by \verb|[ANSWER REDACTED]|. Unlike MathBench, olympiad solutions typically end with a direct answer statement, making the with-solution condition structurally easier. Accuracy ranges from 74\% to 97\% across models. No frontier model exceeds 91\% accuracy without solutions, providing the performance spread required for factor analysis. Grading is performed through SymPy symbolic equivalence, followed by string normalization, with an LLM judge testing for semantic equivalence as a fallback.

\subsection{SciCode}
\label{app:decomposition:scicode}

SciCode~\citep{tian2024scicode} distributes as JSONL with four fields per sub-problem:

\begin{itemize}
  \item \texttt{step\_description\_prompt}, task objective
  \item \texttt{step\_background}, scientific background
  \item \texttt{function\_header}, target function signature and docstring
  \item \texttt{test\_cases}, unit tests
\end{itemize}

The original \texttt{step\_background} field often contained information about units or conventions that the model needed to solve the problem, conflating the question specification and the optional context. We restructure each sub-problem into six named sections using an LLM with the original data as input.

\begin{itemize}
  \item \textsc{Task Specification}, minimal statement of the objective
  \item \textsc{Task Disambiguation}, conventions, units, tolerances, and edge cases
  \item \textsc{Relevant Facts}, background definitions and formulae sufficient to solve without external reference
  \item \textsc{Function Header}, signature and docstring (shown to model)
  \item \textsc{Test Mocks}, stubs for external dependencies
  \item \textsc{Test Cases}, unit tests used for grading
\end{itemize}

These map onto the experimental conditions as follows:

\begin{itemize}
  \item \textbf{Question specification:} \textsc{Task Specification}, \textsc{Task Disambiguation}, \textsc{Function Header}.
  \item \textbf{Context (held out):} \textsc{Relevant Facts}.
\end{itemize}

Transpilation was performed by an LLM with literature access and reviewed by a human. Ambiguities were resolved in favour of the reading implied by the test cases, and unit-test coverage was expanded relative to the original release. The restructured dataset yields 421 trials: 341 sub-problem trials plus 80 general-solution trials (one per problem). For general-solution trials, ground-truth sub-problem implementations are injected as helpers.

\subsection{Quality control}
SQuAD, MMLU-Pro, MathBench, and OmniMath decompositions are programmatic operations on existing dataset fields. The SciCode restructuring introduces LLM-assisted segmentation noise, bounded by the human review pass. The base-rate-aware null bootstrap used throughout the analysis is robust to moderate i.i.d.\ label noise, and the inter-model analyses are robust to errors that affect all models uniformly.

\section{Confidence elicitation prompts}
\label{app:si:prompts}

The verbatim prompt templates used for binary confidence elicitation are listed below. The same template is used for every model, so any phrasing bias that affects every model identically cancels in the inter-model comparisons that drive our results. Logit-based variants (used where providers expose token-level logprobs) share the prompt text and only differ in API call parameters.

\subsection{SQuAD}
\label{app:si:prompts:squad}

\noindent\textbf{Prospective probe.}

\begin{Verbatim}[breaklines=true,breakanywhere=true,fontsize=\small,baselinestretch=1.0]
You will be asked to answer the following question:

"{question}"

Do you think you can answer this question correctly without any additional information?

Respond in JSON format:
{{
    "reasoning": "Brief explanation",
    "can_answer": true or false
}}
\end{Verbatim}

\noindent\textbf{Counterfactual probe.}

\begin{Verbatim}[breaklines=true,breakanywhere=true,fontsize=\small,baselinestretch=1.0]
You were asked the following question with this context passage:

Context passage: {context}

Question: "{question}"

You answered correctly using the context passage above.

Do you think you could have answered this question correctly WITHOUT the context passage?

Respond in JSON format:
{{
    "reasoning": "Brief explanation of your assessment",
    "can_answer": true or false
}}
\end{Verbatim}

\subsection{MMLU-Pro}
\label{app:si:prompts:mmlu}

\noindent\textbf{Prospective probe.}

\begin{Verbatim}[breaklines=true,breakanywhere=true,fontsize=\small,baselinestretch=1.0]
You will be asked the following question:

"{question}"

Do you think you can answer this question correctly without seeing the multiple-choice options?

Respond in JSON format:
{{
    "reasoning": "Brief explanation",
    "can_answer": true or false
}}
\end{Verbatim}

\noindent\textbf{Counterfactual probe.}

\begin{Verbatim}[breaklines=true,breakanywhere=true,fontsize=\small,baselinestretch=1.0]
You were asked the following question with these multiple-choice options:

Question: "{question}"

Choices:
{choices}

You answered correctly using the multiple-choice options above.

Do you think you could have answered this question correctly WITHOUT seeing the multiple-choice options?

Respond in JSON format:
{{
    "reasoning": "Brief explanation of your assessment",
    "can_answer": true or false
}}
\end{Verbatim}

\subsection{LegalBench}
\label{app:si:prompts:legalbench}

\noindent\textbf{Prospective probe.}

\begin{Verbatim}[breaklines=true,breakanywhere=true,fontsize=\small,baselinestretch=1.0]
You will be asked the following question about a legal contract:

"{question}"

Here is the contract text:
{legal_text}

Can you answer this correctly without seeing a formal definition of the clause type?

Respond in JSON format:
{{
    "reasoning": "Brief explanation",
    "can_answer": true or false
}}
\end{Verbatim}

\subsection{MathBench}
\label{app:si:prompts:math_bench}

\noindent\textbf{Prospective probe.}

\begin{Verbatim}[breaklines=true,breakanywhere=true,fontsize=\small,baselinestretch=1.0]
You will be asked to solve the following competition math problem:

"{question}"

Do you think you can solve this problem correctly without seeing any solution steps or hints?

Respond in JSON format:
{{
    "reasoning": "Brief explanation of your confidence assessment",
    "can_solve": true or false
}}
\end{Verbatim}

\noindent\textbf{Counterfactual probe.}

\begin{Verbatim}[breaklines=true,breakanywhere=true,fontsize=\small,baselinestretch=1.0]
You were asked to solve the following competition math problem:

"{question}"

You were shown the following solution steps (with the final answer redacted):
{solution_redacted}

You answered correctly.

Do you think you could have solved this problem correctly without seeing those solution steps?

Respond in JSON format:
{{
    "reasoning": "Brief explanation of your assessment",
    "can_solve": true or false
}}
\end{Verbatim}

\subsection{OmniMath}
\label{app:si:prompts:omnimath}

\noindent\textbf{Prospective probe.}

\begin{Verbatim}[breaklines=true,breakanywhere=true,fontsize=\small,baselinestretch=1.0]
You will be asked to solve the following olympiad-level competition math problem:

"{question}"

Do you think you can solve this problem correctly without seeing any solution steps or hints?

Respond in JSON format:
{{
    "reasoning": "Brief explanation of your confidence assessment",
    "can_solve": true or false
}}
\end{Verbatim}

\noindent\textbf{Counterfactual probe.}

\begin{Verbatim}[breaklines=true,breakanywhere=true,fontsize=\small,baselinestretch=1.0]
You were asked to solve the following olympiad-level competition math problem:

"{question}"

You were shown the following solution steps (with the final answer redacted):
{solution_redacted}

You answered correctly.

Do you think you could have solved this problem correctly without seeing those solution steps?

Respond in JSON format:
{{
    "reasoning": "Brief explanation of your assessment",
    "can_solve": true or false
}}
\end{Verbatim}

\subsection{SciCode}
\label{app:si:prompts:scicode}

\noindent\textbf{Prospective probe.}

\begin{Verbatim}[breaklines=true,breakanywhere=true,fontsize=\small,baselinestretch=1.0]
You will be asked to implement the following Python function:

{signature}

Task description:
{task_spec}

Disambiguation (clarifies expected behaviour, array shapes, conventions, edge cases):
{disambiguation}

Test cases it must pass:
{test_cases}

Given the task description, function signature, and disambiguation above, do you think you can implement this function correctly WITHOUT any additional scientific background or mathematical facts?

Respond in JSON format:
{{
    "reasoning": "Brief explanation",
    "can_solve": true or false
}}
\end{Verbatim}

\noindent\textbf{Counterfactual probe.}

\begin{Verbatim}[breaklines=true,breakanywhere=true,fontsize=\small,baselinestretch=1.0]
You were asked to implement the following Python function:

{signature}

Task description:
{task_spec}

Clarifications you were given:
{disambiguation}

Scientific background you were given:
{facts}

Test cases passed:
{test_cases}

You successfully solved this problem with the clarifications and scientific background provided. Could you have solved it with ONLY the task description and clarifications (WITHOUT the background facts)?

Respond in JSON format:
{{
    "reasoning": "Brief explanation of your assessment",
    "can_solve": true or false
}}
\end{Verbatim}

\section{Compute and access details}
\label{app:si:compute}

All experiments were API-driven and ran between February and April 2026. Models were accessed via OpenAI, Anthropic, Google (Vertex / Gemini API), and OpenRouter (covering DeepSeek, Meta, Mistral, and Qwen). Total spend was approximately \$4{,}000 in API credits across all providers.

\noindent\textbf{Total queries.} The wide cross-model decomposition issues approximately $4{,}921 \text{ items} \times 20 \text{ models} \times 1.8 \text{ probes per item} \approx 1.8 \times 10^{5}$ API calls. The tall stochastic-confidence analysis (\S\ref{sec:residual}) issues an additional batch of calls, one representative model per benchmark with 20 samples per item at high temperature.

\noindent\textbf{Per-provider model identifiers.} The 20 evaluated models are listed by family and identifier in Section~\ref{sec:setup}. Per-call access timestamps and exact provider-snapshot strings are preserved in the framework's local cache. We do not reproduce them here as a per-row table because OpenRouter and provider-side model versioning shifts within the elicitation window are absorbed into the cached responses, and the analysis operates on those responses rather than on live API calls.

\noindent\textbf{Caching and reproducibility.} Every query was cached locally with its prompt, model id, sampling parameters, response, and parsed outcome, and is preserved with the project's experimental release. The paper's results can be regenerated from the cache without re-issuing API calls.

\end{document}